\UseRawInputEncoding 
\documentclass[acmlarge,nonacm]{acmart}
\usepackage{stfloats}

\makeatletter
\def\algbackskip{\hskip\dimexpr-\algorithmicindent-\labelsep}
\def\LState{\Statex \algbackskip}
\makeatother

\makeatletter
\def\subsubsection{\@startsection{subsubsection}{3}%
  \z@{.5\linespacing\@plus.7\linespacing}{.1\linespacing}%
  {\normalfont\itshape}}
\makeatother

\renewcommand\footnotetextcopyrightpermission[1]{} 
\usepackage{url}
\renewcommand\footnotetextcopyrightpermission[1]{} 
\usepackage{listings}

\usepackage{xspace}

\usepackage{placeins}

\usepackage{algorithm}
\usepackage[noend]{algpseudocode}
\algnewcommand{\LineComment}[1]{\State \emph{\textcolor{blue}{\(\triangleright\) #1}}}
\algrenewcommand\algorithmicindent{1em}%
\newcommand{\pluseq}{\mathrel{+}=}
\usepackage{graphicx}
\usepackage{caption}
\usepackage{subcaption}
\usepackage{tabularx}
\usepackage{enumitem}   

\usepackage{color, soul} 
\soulregister\emph7

\newcommand{\newtext}[1]{#1}

\begin{abstract}

During the past decade, novel Deep Learning (DL) algorithms, workloads and hardware have been developed to tackle a wide range of problems. Despite the advances in workload and hardware ecosystems, the programming methodology of DL systems is stagnant. DL workloads leverage either highly-optimized, yet platform-specific and inflexible kernels from DL libraries, or in the case of novel operators, reference implementations are built via DL framework primitives with underwhelming performance. This work introduces the Tensor Processing Primitives (TPP), a programming abstraction striving for efficient, portable implementation of DL workloads with high-productivity. TPPs define a compact, yet versatile set of 2D-tensor operators (or a virtual Tensor ISA), which subsequently can be utilized as building-blocks to construct complex operators on high-dimensional tensors. The TPP specification is platform-agnostic, thus code expressed via TPPs is portable, whereas the TPP implementation is highly-optimized and platform-specific. We demonstrate the efficacy and viability of our approach using standalone kernels and end-to-end DL \& HPC workloads expressed entirely via TPPs that outperform state-of-the-art implementations on multiple platforms.

\end{abstract}

\begin{document}
\newcolumntype{R}{>{\centering\arraybackslash}m{3.5cm}}
\newcolumntype{L}{>{\centering\arraybackslash}m{1.5cm}}
\newcolumntype{M}{>{\centering\arraybackslash}m{3.5cm}}


\title{Tensor Processing Primitives: A Programming Abstraction for Efficiency and Portability in Deep Learning \& HPC Workloads}




\makeatletter
\let\@authorsaddresses\@empty
\makeatother

 
 \author{Evangelos Georganas$^*$, Dhiraj Kalamkar$^*$, Sasikanth Avancha$^*$, Menachem Adelman$^*$, Deepti Aggarwal$^*$, Cristina Anderson$^*$, Alexander Breuer$^\#$, Jeremy Bruestle$^*$, Narendra Chaudhary$^*$,  Abhisek Kundu$^*$, Denise Kutnick$^*$, Frank Laub$^*$, Vasimuddin Md$^*$, Sanchit Misra$^*$, Ramanarayan Mohanty$^*$, Hans Pabst$^*$,  Brian Retford$^*$, Barukh Ziv$^*$, Alexander Heinecke$^*$ \newline \newline
 $^*$I\lowercase{ntel} C\lowercase{orporation}\newline
 $^\#$F\lowercase{riedrich}-S\lowercase{chiller}--U\lowercase{niversität} J\lowercase{ena}
 }

\renewcommand{\shortauthors}{E. Georganas et al.}

\settopmatter{printacmref=false}
\settopmatter{printacmref=false} 
\renewcommand\footnotetextcopyrightpermission[1]{} 

\maketitle

%

\section{Introduction}
\label{sec:introduction}
Since the advent of Deep Learning (DL) as one of the most promising machine learning paradigms almost 10 years ago, deep neural networks have advanced the fields of computer vision, natural language processing, recommender systems, and gradually pervade an increasing number of scientific domains~\citep{origalexnet,szegedy2015going,simonyan2014very,yu2013feature,wu2016google,cheng2016wide,wolf2020transformers,gawehn2016deep,goh2017deep,raghu2020survey}.
Due to the diverse nature of the problems under consideration, these DL workloads exhibit a wide range of computational characteristics and demands. Furthermore, due to the immense computational cost of such workloads, industry and academia have developed specialized hardware features on commodity processors, and even specialized accelerators in order to harness these computational needs~\citep{alom2019state}.

In contrary to the fast-evolving ecosystems of DL workloads and DL-oriented hardware/accelerators, the programming paradigm of DL systems has reached a plateau~\citep{barham2019machine}. More specifically, the development of novel DL workloads involves two types of components: i) Well-established operators within DL libraries (e.g.\ 2D convolutions, inner-product, batch-norm layers in oneDNN~\citep{onednn} and cuDNN~\citep{chetlur2014cudnn}), and ii) Unprecedented, custom primitives which typically instantiate new algorithmic concepts/computational motifs. Unfortunately both of these components come with their shortcomings.

On one hand, the operators within DL libraries are heavily optimized and tuned (usually by vendors) in a platform-specific fashion, leading to monolithic, non-portable and inflexible kernels. Additionally, such opaque and high-level operators prohibit modular design choices since the user/frameworks have to adhere to particular interfaces that may not be adapted to fit the operation under consideration. On the other hand, the custom/unprecedented primitives are typically implemented by the user via the available generic/reference primitives of an ML framework which are not optimized and as such yield underwhelming performance. It is up to the user to create optimized implementations for the custom primitives, leading again to code which is non-portable and potentially requires hardware expertise in order to achieve peak performance. Unfortunately, most of the times such expertise is not available to the data/ML scientist who is developing the custom DL primitive. Therefore, the deployment (or even the evaluation) of a new operator typically requires yet another stage in the development cycle where low-level optimization experts are working on the re-write/fine-tuning of the operator. Later on, in case an operator proves to be important for the community, systems researchers and vendors standardize it, and potentially create yet another monolithic kernel within a DL library for further re-use within DL frameworks. This entire development cycle potentially takes a considerable amount of time (up to years in some cases) and inadvertently impedes the efficient exploration of innovative machine learning ideas~\citep{barham2019machine}. An alternative approach to optimize both types of operators is to leverage contemporary Tensor Compilers (e.g.\ ~\cite{plaidml,chen2018tvm,vasilache2018tensor,zheng2020ansor}), however the state-of-the-art tools are only suitable for compiling small code-blocks whereas large-scale operators require prohibitive compilation times, and often the resulting code performs far from the achievable peak~\citep{barham2019machine}.

We identify that the common source of the problems mentioned in the previous paragraph is the extreme levels of abstraction offered by the DL libraries and the Tensor Compilers. The DL libraries offer coarse-grain, monolithic and inflexible operators whereas the Tensor Compilers usually go to the other extreme, allowing the user to express arbitrary low-level operators without any minimal restrictions that would readily enable efficient lifting and code-generation in their back-ends (e.g.\ they offer no minimal/compact set of allowed operations on tensors). To exacerbate the challenge of optimal code generation, Tensor Compilers usually undertake the cumbersome tasks of efficient parallelization, loop re-ordering, automatic tiling and layout transformations, which, to date, remain unsolved in the general setup. Also, there is not a well-established way to share state-of-the-art optimizations among the plethora of Tensor Compilers and as a result each one has its own advantages and disadvantages, which translates eventually to sub-optimal performance on real-world scenarios~\citep{li2020deep}. We note here the recent, promising effort of MLIR~\citep{mlir} towards unifying the optimization efforts in the Tensor Compiler IR infrastructure.

In this work we introduce the Tensor Processing Primitives (TPP), a programming abstraction striving for efficient and portable implementation of Tensor operations, with a special focus on DL workloads. TPPs define a set of relatively low-level primitive operators on 2D Tensors, which in turn can be used as basic building blocks to construct more complex operators on high-dimensional tensors. TPPs comprise a minimal and compact, yet expressive set of precision-aware, 2D tensor level operators to which high-level DL operators can be reduced. TPPs's specification is agnostic to targeted platform, DL framework, and compiler back-end. As such the code which is expressed in terms of TPPs is portable. Since the level of abstraction that TPPs adopt is at the sub-tensor granularity, TPPs can be directly employed by DL workload developers within the frameworks, or could be alternatively used to back up an IR within a Tensor Compiler stack, i.e.\ TPPs could form the basis of an MLIR dialect.

While the TPP specification is agnostic of the targeted framework/platform/compiler stack, its implementation is platform specific, and is optimized for the target architectures. This subtle detail offers a clear separation of concerns: the user-entity of TPPs, either a developer or a compiler framework, can focus on expressing the desired algorithm and its execution schedule (e.g.\ parallelization, loop orders) using the TPP tensor abstraction, whereas the efficient, platform-specific code generation pertaining to the TPP operations belongs to the TPP back-end. To this extent, TPPs could be also viewed as a ``virtual Tensor ISA" that abstracts the actual physical ISA of the target (e.g.\ SSE, AVX2, AVX512, AMX for x86, AArch64 and ARMv8 SVE , xPU).

\begin{figure}[t!]
\centering
\includegraphics[width=0.8\columnwidth]{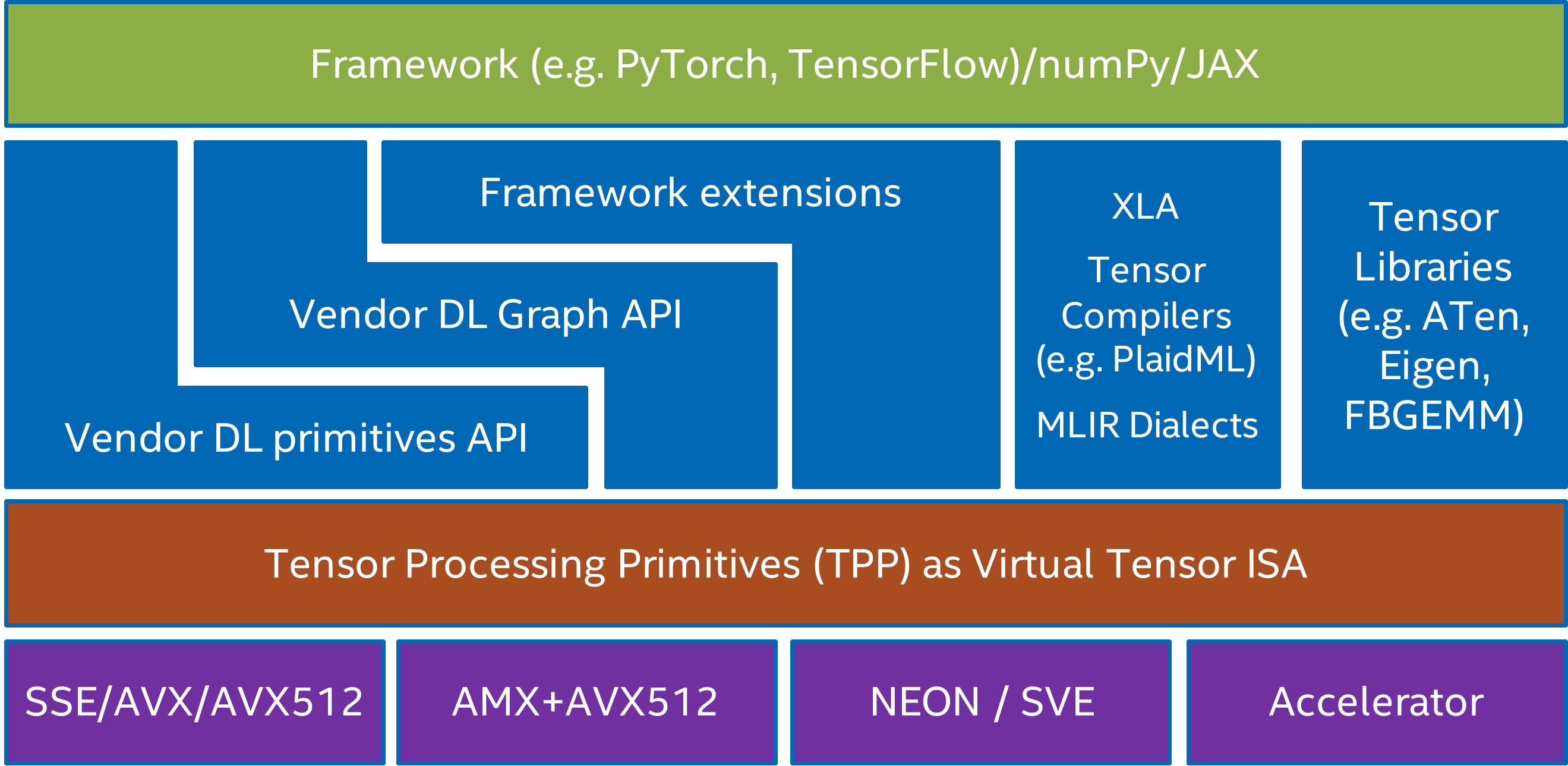}
\caption{Use-cases of TPPs in various software stacks.}
\label{fig:stack}
\end{figure}

Figure~\ref{fig:stack} shows various use-cases of TPPs within multiple software stacks. TPPs can be viewed as a layer abstraction of the actual physical target ISA, and the user-entities can rely on the TPP layer for the code generation pertaining to the tensor operations. Also, Figure~\ref{fig:stack} illustrates the various user-entities that might leverage TPPs. First, the vendor-optimized DL libraries (e.g.\ oneDNN or oneDNN Graph) can use TPPs for optimized code generation in their back-end. Second, the user/developer of the DL operators can directly leverage TPPs within a DL framework extension to express the underlying tensor computations (e.g.\ the user may develop a framework extension for a novel DL operator by employing the TPPs as building blocks). Third, Tensor Compilers can leverage TPPs (e.g.\ as part of an MLIR dialect) to generate high-quality code for the corresponding tensor operators. As such, the TPP layer abstraction offers a clear separation of concerns where the Tensor Compiler may focus on higher-level optimizations (loop tiling and re-ordering, parallelization etc) whereas the platform-specific code generation of the tensor operations is undertaken by the TPP layer. Such a synergistic Tensor Compiler - TPP paradigm is illustrated in Section~\ref{subsec:plaidml}. Last but not least, TPPs could be leveraged by more general Tensor Libraries (e.g. ATen, Eigen) where tensor computations constitute the primary focus and they can be naturally mapped to TPPs.

In our Proof-Of-Concept (POC) implementation of TPPs we leverage JIT technology to emit performant and platform-specific code during runtime. Furthermore, in our POC we define a mini embedded Domain Specific Language (mini-eDSL) where the TPPs can be combined via matrix equations in order to build high-level operators without sacrificing performance.

We demonstrate the efficiency of our approach on multiple platforms using standalone kernels written entirely with TPPs and compare the performance to vectorized-by-expert code and compiler generated code. Finally, we showcase the expressiveness and viability of our methodology by implementing contemporary end-to-end DL workloads using solely the TPP abstractions and show how we can outperform the state-of-the-art implementations on multiple platforms. The main contributions of this work are:
\begin{itemize}
\item A TPP specification/foundation for primitive tensor operations.
\item A Proof-Of-Concept implementation of the TPP specification along with a mini-eDSL (called TPP Matrix Equations), enabling efficient fusion of TPPs that lead to portable, high-level tensor operations. \newtext{We describe in detail various standalone TPP implementations, and also we provide a detailed analysis of our TPP Matrix Equation mini-eDSL framework.}
\item A demonstration of how contemporary and novel DL algorithmic motifs/workloads can be expressed in their entirety via TPPs.
\item An experimental evaluation of the TPP-based DL workloads from all relevant fields (image processing, recommendation systems, natural language processing, graph processing and applications in science) on multiple platforms (different instruction set architectures (ISAs) x86\_64 and aarch64, and micro-architectures for each ISA), including distributed-memory scaling. We show performance that matches/exceeds the state-of-the-art implementations, while maintaining flexibility, portability and obviating the need for low-level platform-specific optimizations.
\item \newtext{We show how TPPs can be leveraged as a virtual Tensor ISA within a Tensor compiler software stack, yielding high-performance DL primitives.}
\item \newtext{We illustrate examples of how TPPs are used outside of Deep Learning, in High Performance Computing (HPC) applications in order to accelerate tensor computations.}
\end{itemize}
 Section~\ref{sec:specification} details the specification of the TPPs. Then, Section~\ref{sec:implementation} illustrates a POC implementation of the TPP specification. \newtext{Section~\ref{sec:tpp_eqs} presents an infrastructure that enables efficient TPP fusion.} In Section~\ref{sec:workloads} we exhibit how contemporary DL motifs/algorithmic paradigms can be expressed via TPPs. Section~\ref{sec:results} presents an experimental evaluation of TPP-based DL kernels and workloads on multiple platforms. \newtext{Section~\ref{subsec:plaidml} outlines our POC implementation of a TPP backend within a tensor compiler (PlaidML), and also presents some results highlighting the viability of the TPP abstraction as a virtual Tensor ISA within tensor compiler stacks. Section~\ref{sec:hpc} presents exemplary usage of TPPs within HPC applications in order to efficiently implement tensor computations.}
 Sections~\ref{sec:related} and~\ref{sec:conclusions} summarize the related work and conclude this paper.

\section{The TPP Specification}
\label{sec:specification}

\subsection{TPP Design Principles}
\label{subsec:principles}
The TPP specification is driven by a few design principles:

1) \emph{Each TPP corresponds to a mathematical operator that takes a number of input(s) and produces an output}. We opt to specify TPPs that correspond to basic, well-defined mathematical tensor operations. In this way we keep the set of TPPs \emph{minimal} albeit \emph{expressive}; basic TPPs can be combined to formulate more complex operators.

2) \emph{The inputs/outputs of the TPPs are abstract 2D tensors that can be fully specified by their shape/size, leading dimensions, and precision}. Additionally, the 2D tensors hold the following complementary \emph{runtime} information: (i) a \emph{primary} field which corresponds to the memory address where the 2D (sub)tensor data resides, (ii) a \emph{secondary} field holding optional data for the tensor (e.g.\ a mask for the tensor), and (iii) a \emph{tertiary} field holding optional, auxiliary information of the tensor (e.g.\ scaling factors for a quantized tensor.)

3) \emph{TPPs are specified as ``memory-to-memory" operations, or equivalently the input/output tensors are residing in memory locations specified by the user}. This design decision is critical in order to abstract the TPPs from all physical ISAs, and enables true platform-agnostic specification. {\color{black}For example, if the TPPs were accepting vector registers as inputs/outputs, then the number of physical registers, the vector length and dimensionality would be exposed in the API of TPPs, making the specification platform-specific.}

4) \emph{TPPs have declarative semantics}. As such, the TPP specification does not preclude potential parallelism (e.g.\ SIMD, SIMT) in the back-end implementation which is target-specific.

5) \emph{TPPs are composable in a producer-consumer fashion}. Since the output of a TPP is a well-defined tensor $O$, it can be fed as input to a subsequent TPP. In such a scenario, this ``intermediate" tensor $O$ is not necessarily exposed to the user, unless the user explicitly requires it (e.g. by combining the TPPs in a manual fashion via an explicit temporary $O$ buffer/tensor which lives in the user space/application). This flexibility allows the TPP implementation (which is platform-specific) to combine TPPs in the most efficient way for the target architecture (e.g.\ the $O$ tensor can live at the physical register file in the composite TPP in order to avoid redundant memory movement).

6) \emph{The TPP input/output tensors as well as the computation itself are precision aware}. This feature makes mixed precision computations (that are prominent in DL workloads) easy to express from the user point of view, and provides information to the TPP back-end that may enable efficient implementation.

\subsection{TPP Arguments}
\label{subsec:args}
As mentioned in the previous subsection, the input to TPPs are 2D tensors. Each 2D tensor can be specified by the number of rows $M$, columns $N$, its leading dimension $ld$ and its datatype $dtype$. Additionally, during runtime each tensor gets fully characterized by specifying its location/address as \emph{primary} info, optional companion tensor info as \emph{secondary} (e.g.\ sparsity bitmask), and optionally \emph{tertiary}  info (e.g.\ in case the tensor shape is dynamically determined at runtime, this info may contain variables specifying $M$/$N$). Each TPP also specifies the shape/precision of the produced/output 2D tensor.

Each TPP also supports input tensors with \emph{broadcast} semantics. More specifically, TPPs accept optional flags dictating that the input 2D tensor should be formed by broadcasting a column/row/scalar $N$/$M$/$M\times N$ times respectively. Finally, the TPPs accept optional flags which further specify the TPP operation. For example, in case a TPP is computing a transcendental function, the flags may be specifying various approximation algorithms used for the computation. In the next subsection we present the TPPs in three groups: \emph{unary}, \emph{binary}, and \emph{ternary} TPPs given the number of input tensors they accept.

\subsection{The TPP collection}
\label{subsec:tpppresentation}
\begin{table}[!tb]
\fontsize{9}{9 }\selectfont
\centering
\begin{tabular}{|p{3cm}|p{12cm}|}
    \hline
    {\textbf{Unary TPP}} & \textbf{Description/Comments}\\
    \hline
    \hline
    \textbf{Identity} &  Copies input to output. Given input/output datatype, it performs datatype conversions \\
    \hline
    \textbf{Zero}  &  Fills output with zeros \\
    \hline
    \textbf{Square}  &  Squares input and stores to output \\
    \hline
    \textbf{Increment / decrement}  &  Increments / Decrements input by 1 and stores to output  \\
    \hline
    \textbf{Square root}  &  Computes the square root of input and stores to output   \\
    \hline
    \textbf{Reciprocal}  &  Computes the reciprocal of input and stores to output   \\
    \hline
    \textbf{Rcp. Sqrt.}  &  Computes the rcp. sqrt. of input and stores to output   \\
    \hline
    \textbf{Exp}  &  Computes the exponential value of the input tensor entries and stores them to output  \\
    \hline
    \textbf{PRNG}  &  Generates an output tensor with pseudo-random entries \\
    \hline
    \textbf{(De)Quantize}  &  Quantizes / Dequantizes the input  \\
    \hline
    \textbf{Reduce}  &  Reduces the rows/columns of the input and stores to output. The reduction function can be SUM/MUL/MIN/MAX; (optionally) reduces the \emph{squared} input  \\
    \hline
    \textbf{Transform}  &  Transforms input and stores to output. Transformations are:  Transpose, VNNI formatting, and VNNI to VNNI-transpose \\
    \hline
    \textbf{Unpack}  &  Takes each entry $x_{i,j}$ of the input tensor, splits it in two parts $x_{i,j}^{lo}$ and $x_{i,j}^{hi}$  with same bit-width, and stores them in two tensors $X^{lo}$, $X^{hi}$ \\
    \hline
    \textbf{Replicate columns}  & Takes an input column/vector, replicates it a variable number of times and forms the output \\
    \hline
    \textbf{Gather / Scatter}  &  Gathers/Scatters rows/columns from input and forms the tensor  \\
    \hline
    \textbf{2D Gather / 2D Scatter}  &  Gathers/scatters elements from input using 2D offsets   \\
    \hline
    \textbf{2D-strided loads / stores}  &  Loads/stores elements from/to a tensor using primary and secondary strides \\ 
    \hline
    \textbf{Tanh  \&Tanh\_inv}  &  Computes the hyperbolic tangent function (or its inv used for back-propagation) on input  \\
    \hline
    \textbf{RELU \& RELU\_inv}  & Apply a Rectified Linear Unit  function (or its inv used for back-propagation) on input   \\
    \hline
    \textbf{Sigmoid \& Sigmoid\_inv}  &  Computes the logistic sigmoid (or its inv used for back-propagation) on input    \\
    \hline
    \textbf{GELU \& GELU\_inv}  &  Apply a Gaussian Error Linear Unit function (or its inv used for back-propagation) on input   \\
    \hline
    \textbf{Dropout \& Dropout\_inv}  &  Drops out values from the input tensor with probability $p$. For the inv/back-propagation pass, the same dropped units are zeroed out  \\
    \hline
  \end{tabular}
\caption{\label{tab:unary}Unary TPPs}
\end{table}

\begin{table}[!tb]
\fontsize{9}{9 }\selectfont
\centering
\begin{tabular}{|p{3cm}|p{12cm}|}
    \hline
    {\textbf{Binary TPP}} & \textbf{Description/Comments}\\
    \hline
    \hline
    \textbf{Add} &  Add two inputs\\
    \hline
    \textbf{Sub}  & Subtracts two inputs \\
    \hline
    \textbf{Mul}  &  Multiples (elementwise) two inputs \\
    \hline
    \textbf{Div}  &  Divides two inputs  \\
    \hline
    \textbf{Max/Min}  &  Finds element-wise max/min of two inputs  \\
    \hline
    \textbf{MatMul}  &  Performs matrix multiplication of two input   \\
    \hline
    \textbf{Pack}  & Concatenates pairs of entries  $x_{i,j}^{lo}$ and $x_{i,j}^{hi}$ from the inputs $X^{lo}$, $X^{hi}$ into $x_{i,j}$ and stores it to the output $X$ \\
    \hline
    \textbf{Compare}  &  Compares element-wise two inputs and stores a bitmask of the comparison   \\
    \hline
  \end{tabular}
\caption{\label{tab:binary}Binary TPPs}
\end{table}

\begin{table}[!tb]
\fontsize{9}{9 }\selectfont
\centering
\begin{tabular}{|p{3cm}|p{12cm}|}
    \hline
    {\textbf{Ternary TPP}} & \textbf{Description/Comments}\\
    \hline
    \hline
    \textbf{GEMM} &  Performs on 2D inputs $A$, $B$, $C$, scalar $\beta$: $C = \beta C +  A\times B$ \\
    \hline
    \textbf{Batch-Reduce GEMM}  &  Performs on 2D inputs $A_i$, $B_i$  (with $i=0, 1,$\dots$, n-1$), $C$, scalar $\beta$: $C = \beta C +  \sum_{i=0}^{i=n-1}A_i\times B_i$ \\
    \hline
    \textbf{(N)MulAdd}  &  Performs on 2D inputs $A$, $B$, $C$: $C = C +  A\odot B$ (or $C = C - A\odot B$ ); $\odot$ denotes element-wise multiplication\\
    \hline
    \textbf{Blend}  &  Blends 2D input tensors $A$, $B$ according to bitmask $C$ \\
    \hline
  \end{tabular}
\caption{\label{tab:ternary}Ternary TPPs}
\end{table}

First, we highlight the ternary \emph{Batch-Reduce GEMM} (BRGEMM) TPP which is the main building block for general tensor contractions in DL kernels~\citep{georganas2020harnessing}. BRGEMM materializes the operation $C = \beta \cdot C + \sum_{i=0}^{n-1} A_i \times B_i$.
In essence, this kernel multiplies the specified blocks $A_i^{M\times K}$ and $B_i^{K\times N}$ and reduces the partial results to a block $C^{M\times N}$. It is noteworthy that tensors $A$ and $B$ can alias and also the blocks $A_i$ and $B_i$ can reside in any position in the input (potentially high-dimensional) tensors $A$ and $B$. Previous work~\citep{georganas2020harnessing} has shown that this single building block is sufficient to express efficiently tensor contractions in the most prevalent DL computational motifs, namely: Convolution Neural Networks (CNN),  Fully-Connected networks (FC), Multi-Layer Perceptrons (MLP), Recurrent Neural Networks (RNN)/Long Short-Term Memory (LSTM) Networks. In Section~\ref{sec:workloads} we exhibit how BRGEMM can be further used to build efficient Attention Cells that comprise the cornerstone of modern Natural Language Processing (NLP) workloads. BRGEMM can be specialized to one of the following three variants that may enable more efficient implementations on various platforms: (i) \emph{address-based BRGEMM}, where the addresses of the blocks $A_i$ and $B_i$ are explicitly provided by the user, (ii) \emph{offset-based BRGEMM}, where the addresses of $A_i$ and $B_i$ can be computed as $address\_A_i = address\_A + o\textit{ff}set_{A_i}$ and $address\_B_i = address\_B + o\textit{ff}set_{B_i}$, and (iii) \emph{stride-based BRGEMM}, where the addresses of $A_i$ and $B_i$ are: $address\_A_i = address\_A_{i-1} + stride\_A$ and $address\_B_i = address\_B_{i-1} + stride\_B$. \newtext{In subsection~\ref{subsec:brgemm_structure} we present the implementation of the BRGEMM TPP in more depth for various ISAs and platforms.}

Table~\ref{tab:unary} presents the unary TPPs that accept one 2D tensor as input. Since most of these TPPs map directly to the equivalent math function, we further elaborate only on the ones which are more complex. The \emph{Identity} TPP essentially copies the input to the output. Since the input and output are fully specified in terms of their precision, this TPP can be also used to perform datatype conversions between tensors.

The \emph{Quantize \& Dequantize} TPPs are used to quantize/dequantize the input tensor whereas the exact algorithm employed is specified by a TPP flag.

The \emph{Transform} TPP uses a flag to determine the exact transformation applied on the input 2D tensor. The \emph{Transpose} transformation is the usual mathematical matrix transpose. The rest two types of transformation, namely \emph{VNNI formatting}, and \emph{VNNI to VNNI-transpose} are DL specific. More specifically, modern hardware (e.g.\ Intel's Cooper Lake) requires tensors to be in specific format called \emph{VNNI} in order to employ hardware acceleration for specific operations, e.g. dot-products (see section~\ref{subsec:brgemm_mixed} for more details). This format represents a logical 2D tensor $[D_1][D_0]$ as a 3D tensor $[D_1/\alpha][D_0][\alpha]$ where essentially the dimension $D_1$ is blocked in chunks of size $\alpha$, which in turn are set as the inner-most tensor dimension. The \emph{VNNI formatting} TPP performs this exact transformation: $[D_1][D_0] \rightarrow [D_1/\alpha][D_0][\alpha]$ and the \emph{VNNI to VNNI-transpose} transposes a tensor which is already laid out in VNNI format, i.e.\ performs  $[D_1/\alpha_1][D_0][\alpha_1]\rightarrow [D_0/\alpha_0][D_1][\alpha_0]$. \newtext{In subsection~\ref{subsec:transform} we outline how the Transform TPPs are implemented via Shuffle Networks.}

The last four entries of Table~\ref{tab:unary} correspond to DL-specific operations. They correspond to activation functions typically encountered in DL workloads. All these activation functions have a counterpart which is required during the back-propagation pass of training DL networks. These DL specific TPPs could be built on top of other TPPs, however since they are prevalent in DL workloads we opt to define them as self-contained TPPs for ease of usage. \newtext{ In subsection~\ref{subsec:approx} we describe the TPP implementation of non-linear approximations for several activation functions on various ISAs.}

Table~\ref{tab:binary} and Table~\ref{tab:ternary} present the binary/ternary TPPs that accept two/three 2D tensor as inputs respectively. 

\section{TPP Implementation}
\label{sec:implementation}
In this Section we briefly describe our Proof-Of-Concept (POC) implementation of the TPP specification. Our implementation targets multiple CPU architectures from various vendors that support different ISAs, but could be readily extended to support even GPU ISAs. We build upon and extend the open source LIBXSMM~\citep{libxsmm} library which leverages JIT techniques. Such JIT techniques have been successfully used for optimal code generation on CPUs by taking advantage of the known (at runtime) tensor shapes/dimensions in HPC and DL applications~\citep{libxsmm,sc18,georganas2020harnessing}. Nevertheless, the TPP specification is platform-agnostic and does not preclude any TPP back-end implementation.  In our POC implementation, the usage of TPPs is governed by two APIs: i) A dispatch API with which the user can request the code generation of a specific TPP, and such a  dispatch call JITs a function implementing the requested operation, ii) an API to call the JITed TPP kernel. \newtext{First, in Subsection~\ref{subsec:tpp_structure} we provide a generic blueprint of our TPP implementation. Then, in subsection~\ref{subsec:brgemm_structure} we describe in more detail the BRGEMM TPP implementation which comprises the main tensor contraction tool in the TPP abstractions. Subsection~\ref{subsec:transform} details the implementation of the unary transform TPPs via shuffle networks since their efficient implementation diverts from the generic TPP blueprint. Finally, subsection~\ref{subsec:approx} outlines the approximation techniques we leverage in our TPP implementation of non-linear activation functions; such approximations are essential in achieving high-performance, while at the same time their accuracy is sufficient for the purposes of training DL workloads.}

\subsection{Generic TPP Implementation Blueprint}
\label{subsec:tpp_structure}
\begin{algorithm}[t]
\begin{algorithmic}[1]
\LState \textbf{Inputs}: $\textbf{X}^{M\times N}$, ($\textbf{Y}^{M\times N}$, $\textbf{Z}^{M\times N}$ if binary/ternary)
\LState \textbf{Output}: $\textbf{O}^{M\times N}$
\For{$i_n=0 \dots N-1\ \textbf{with\ step\ }\mathbf{n_b}$}
\For{$i_m=0 \dots M-1\ \textbf{with\ step\ }\mathbf{m_b}$}
\LineComment{Generic loads, may have broadcast/gather semantics,}
\LineComment{and may perform datatype conversions}
\State $X_b$ $\leftarrow$  load\_generic $m_b \times n_b$ $X$-$\text{subblock}_{i_m,i_n}$
\If{(unary TPP)}
\State $X_b$ $\leftarrow$ $\textbf{Unary\_op}$$(X_b)$
\EndIf
\If{(binary TPP)}
\State $Y_b$ $\leftarrow$ load\_generic $m_b \times n_b$ $Y$-$\text{subblock}_{i_m,i_n}$
\State $X_b$ $\leftarrow$ $\textbf{Binary\_op}$$(X_b, Y_b)$
\EndIf
\If{(ternary TPP)}
\State $Y_b$ $\leftarrow$ load\_generic $m_b \times n_b$ $Y$-$\text{subblock}_{i_m,i_n}$
\State$Z_b$ $\leftarrow$ load\_generic $m_b \times n_b$ $Z$-$\text{subblock}_{i_m,i_n}$ 
\State $X_b$ $\leftarrow$ $\textbf{Ternary\_op}$$(X_b, Y_b, Z_b)$
\EndIf
\LineComment{Generic store, may have scatter semantics, and may}
\LineComment{perform datatype conversion}
\State  $O$-$\text{subblock}_{i_m,i_n}\ \  \overleftarrow{\text{store\_generic}} \ \   $ $X_b$
\EndFor
\EndFor
\end{algorithmic}
\caption{The generic unary/binary/ternary TPP algorithm}
\label{alg:generic_tpp}
\end{algorithm}

Algorithm~\ref{alg:generic_tpp} exhibits at a high-level the pseudocode that is used to implement the Unary/Binary/Ternary TPPs in a unified fashion. The inputs of the TPPs are tensors $X$, $Y$ (in case of binary/ternary TPPs) and $Z$ (in case of ternary TPP), and an output tensor $O$. For the purposes of this simplified presentation we assume all tensors are of size $M\times N$, however, depending on the operation these might have different sizes. For example, if the unary OP is a reduction-by-columns and the input is $M\times N$, then the output is an $M\times 1$ vector. First, we show that the $M$/$N$ loops are blocked with factors $m_b$/$n_b$ such that the working sets of each microkernels fits on the available register file. The latter is architecture specific, e.g.\ AVX2-enabled ISAs expose 16 256-bit vector registers, AVX512-enabled ISAs expose 32 512-bit vector registers and Aarch64 features 32 128-bit (NEON)/512-bit (SVE) vector registers. The ``load\_generic" function used in Algorithm~\ref{alg:generic_tpp} denotes the loading of a sub-tensor to a register block; this load may imply row/column/scalar broadcast semantics if the user specified the TPP in that way, or even may have strided-load/gather semantics if the TPP involves a strided-load/gather operation. Also, for simplicity we do not show here the handling of ``secondary" fields of the tensors that may be required (e.g.\ indices array for the gather operation, bitmasks arrays). Additionally, the generic load also handles datatype conversion, for instance provided the input is in bfloat16 (BF16)~\citep{bfloat16_tf} whereas the compute is going to happen in FP32 precision. Once all the required sub-tensors are loaded, then the corresponding Unary/Binary/Ternary operator is applied. This operator may be directly mapped to an available instruction (e.g. a vector add in case of binary addition), or to a sequence of instructions for more complicated operators (e.g.\ reductions, random number generation via xorshift algorithm~\citep{marsaglia2003xorshift}, approximation algorithms for transcendental functions~\citep{banerjee2019optimizing}). Last but not least, the optimal sequence generation depends on the available instructions and this is handled by the TPP back-end/JITer. For example, some ISAs may have masking/predicate support (e.g.\ AVX512 \& SVE) that enable efficient handling of loop remainders, the selected unrolling degree heavily depends on the instructions in use, their latency and the number of available architectural registers. Once the result is computed, the resulting register block is stored back to the corresponding output sub-tensor position. Similarly to the generic load, the ``generic" store may induce strided accesses or may be even a scatter operation. Additionally, the generic store also handles potential datatype conversions.

\subsection{The BRGEMM TPP Implementation}
\label{subsec:brgemm_structure}
\subsubsection{The BRGEMM kernel structure}
\begin{algorithm}[t]
\begin{algorithmic}[1]
\LState \textbf{Inputs}: $A_i^{M\times K},B_i^{K\times N}\ \text{for} \ i = 0,...,n\text{-}1$,\ $C^{M\times N}$, $\beta \in {\rm I\!R}$
\LState \textbf{Output}:$\ C = \beta \cdot C + \sum_{i=0}^{n-1} A_i \times B_i$
\For{$i_n=0 \dots N-1\ \textbf{with\ step\ }\mathbf{n_b}$}
\For{$i_m=0 \dots M-1\ \textbf{with\ step\ }\mathbf{m_b}$}
\State acc\_regs $\leftarrow$  load\_generic $m_b \times n_b$ $C$-$\text{subblock}_{i_m,i_n}$ 
\For{$i=0 \dots n-1\ \textbf{with\ step\ } \mathbf{1}$}
\For{$i_k=0 \dots K-1\ \textbf{with\ step\ } \mathbf{k_b}$}
\LineComment{Outer product GEMM microkernel}
\State acc\_regs $\pluseq$  $A_i\ \text{sub-panel}_{i_m,i_k}\times B_i\ \text{sub-panel}_{i_k,i_n} $ 
\EndFor
\EndFor
\State  $C$-$\text{subblock}_{i_m,i_n}  \overleftarrow{\text{store\_generic}}\ $ acc\_regs
\EndFor
\EndFor
\end{algorithmic}
\caption{The batch-reduce GEMM TPP}
\label{alg:br_kernel}
\end{algorithm}
\begin{figure}[t!]
\centering
\includegraphics[width=\columnwidth]{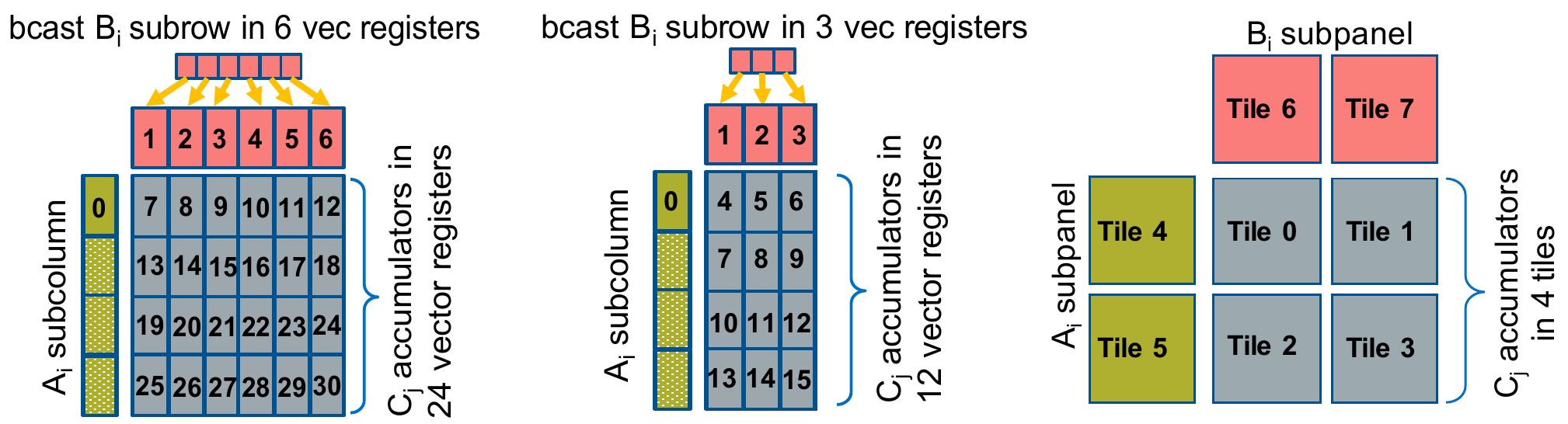}
\caption{Outer product GEMM microkernels, \emph{Left}: On a platform with 32 vector registers, \emph{Middle}: On a platform with 16 vector registers, \emph{Right}: On a platform with 8 2D registers (tiles).}
\label{fig:brgemm_micro}
\end{figure}
We present in more detail the BRGEMM TPP because it comprises the tensor contraction tool in the TPP abstraction, and is ubiquitous in the DL kernels and workloads described in Section~\ref{sec:workloads}.  Algorithm~\ref{alg:br_kernel} exhibits the high-level algorithm implementing: $\ C = \beta \cdot C + \sum_{i=0}^{n-1} A_i \times B_i$ . Lines 1-2 block the computation of the result $C$ in $m_b \times n_b$ tensor sub-blocks. Once such a subblock is loaded into the accumulation registers (line 3), we loop over all pairs $A_i,\ B_i$ (line 4) and we accumulate into the loaded registers the products of the corresponding $m_b\times K$ subblocks of $A_i$ with the relevant $K\times n_b$ subblocks of $B_i$ (lines 5-7). In order to calculate a partial product of an $m_b\times k_b$ sub-panel of $A_i$ with a $k_b\times n_b$ sub-panel of $B_i$, we follow an outer product formulation. The loading of $A_i$ and $B_i$ sub-panels, and the outer-product formulation is heavily dependent on the target platform. We provide BRGEMM implementations for multiple x86 ISAs: SSE, AVX,  AVX2,  AVX512, including the recently introduced Intel AMX (Advanced Matrix Extensions) ISA~\citep{intelisa}. Additionally, we have implemented the BRGEMM TPP for AArch64 and ARMv8 SVE ISAs. Depending on the targeted platform, the ``register" can be either a typical vector register with varying width (e.g.128-512bit vector length), or in the case of AMX-enabled target the ``register" is a 2D tile-register. Similarly, the outer-product formulation may employ the available Fused-Multiply-Add (FMA) instructions, or even 2D tile-multiplication instructions. In all these cases, the TPP implementation emits the appropriate load/store/prefetch/FMA instructions, and takes into account the available architectural registers/unrolling factors/instruction mix in order to achieve close to peak performance. Last but not least, the BRGEMM supports multiple datatypes (FP64, FP32, BF16, INT8), and whenever possible employs hardware acceleration, e.g.\ via specialized FMA instructions for INT8/BF16 datatypes. \newtext{In order to highlight the differences of the outer product GEMM microkernels that are heavily dependent on the target platform, we show in Figure~\ref{fig:brgemm_micro} three different implementations.}

\newtext{Figure~\ref{fig:brgemm_micro}-Left shows an exemplary outer product microkernel on a platform with 32 available vector registers, for example an x86 with AVX512 or on ARM AArch64/SVE. In this case vector register v7-v30 constitute the accumulators, vector registers v1-v6 hold a broadcasted subrow of B, and vector register v0 is used to load a partial subcolumn of A. First, we load on v1-v6 a subrow of B via broadcasts, then we load on v0 the first chunk of the A subcolumn and with 6 fused multiply-add (FMA) instructions (v0 with v1-v6) we multiply-and-add the corresponding partial results on the accumulators v7-v12 (first logical row of accumulators). Then, we load on v0 the second chunk of the A subcolumn, and subsequently with yet another 6 FMA instructions (v0 with v1-v6) we multiply-and-add the computed partial results on the accumulators v13-v18 (second logical row of accumulators) etc. The registers v1-v6 are reused 4 times throughout the outer product computation, and v0 is reused 6 times for each loaded A chunk. In other words, the corresponding A subcolumn and B subrow are loaded from memory/cache into the vector registers exactly once and we get to reuse them from the register file. Also, in such a formulation we expose 24 independent accumulation chains which is critical in order to hide the latency of the FMA instruction. Last but not least, the platform (i.e.\ vector register width) and the datatype of the microkernel determine the exact values of the blocking parameters $m_b$, $n_b$, and $k_b$. For example for single precision datatype FP32 and an x86 AVX512 platform, each vector register can hold 16 FP32 values (the vector registers are 512-bit wide). Therefore, this microkernel operates with blocking values $m_b=64$, $n_b=6$, and $k_b=1$ and it calculates a small matrix multiplication $C_{64\times 6} \pluseq A_{64\times1}\times B_{1\times6}$.}

\newtext{Figure~\ref{fig:brgemm_micro}-Middle shows an exemplary outer product microkernel on a platform with 16 vector registers, for example an x86 with up to AVX2 ISA. The microkernel is similar with the previous case; since we have only 16 vector registers available, we dedicate 12 of those as $C$ accumulators, 3 vector register are utilized for holding a partial B subrow, and 1 vector register is used to load a chunk of an A subcolumn. In this case 12 independent accumulation chains are also sufficient to hide the FMA latency. Analogously to the previous case, for single precision datatype FP32 and an x86 AVX2 platform, each vector register can hold now 8 FP32 values (the vector registers are now 256-bit wide). Thus, this microkernel operates with blocking values $m_b=32$, $n_b=3$, and $k_b=1$ and it calculates a small matrix multiplication $C_{32\times 3} \pluseq A_{32\times1}\times B_{1\times3}$.}

\newtext{Figure~\ref{fig:brgemm_micro}-Right shows a small GEMM microkernel on a platform with 8 2D registers (tiles), for example what is available in the recently introduced Intel AMX (Advanced Matrix Extensions) ISA. In this case each 2D tile register has size (up to) 1KB, logically holds (up to) 16 rows of a submatrix, and can be loaded with a proper tile-load instruction. In this particular example, tiles 0-3 comprise the $C$ accumulators, tiles 4-5 are used to hold a subpanel of A and tiles 6-7 are used to hold a subpanel of B. Once we load the subpanels of A and B onto the respective tiles, we can perform 4 tile multiply-and-add instructions: $tile0 \pluseq tile4 \times tile6$, $tile1 \pluseq tile4 \times tile7$, $tile2 \pluseq tile5 \times tile6$ and $tile3 \pluseq tile5 \times tile7$, and we update the $C$ accumulators. In such a microkernel, each A/B tile is reused 2 times. Given each tile may have size up to 1KB and may hold up to 16 rows of a submatrix, by considering BF16 datatype for A/B matrices and FP32 accumulator tiles, such a microkernel operates with blocking values $m_b=32$, $n_b=32$, $k_b=32$, and can compute (up to) a small matrix multiplication $C_{32\times 32} \pluseq A_{32\times32}\times B_{32\times32}$. Each A/B tile represents a logical $16\times 32$ BF16 A/B submatrix, and each C tile represents a $16\times 16$ FP32 accumulator. The AMX instructions will be available within the upcoming Intel Xeon processors code-named Sapphire Rapids, and the corresponding BF16-input/FP32-output tile multiplication instructions can deliver up to $16\times$ more FLOPs/cycle compared to FP32 AVX512 FMA instructions on current Xeon platforms.}

\newtext{These considerably different GEMM microkernel variants highlight yet another aspect of the TPPs: The TPPs specify \emph{what} needs to be done rather than how it is done/implemented. In this case, the user may just specify/employ a BRGEMM TPP in order to perform a tensor contraction, whereas the TPP backend/implementation is responsible for generating the optimal code for each platform at hand. In this methodology, all the architectural nuances are hidden completely by the user, and the same exact user code written in terms of TPPs may be reused across platforms with different characteristic/ISAs without sacrificing performance or portability.}

\subsubsection{Mixed Precision BRGEMM and its emulation}
\label{subsec:brgemm_mixed}

\newtext{While the previous section presents the general structure of mapping matrix multiplication to various physical ISAs, this paragraph is used to demonstrate how the idea of a virtual ISA allows to implement operations efficiently which are not natively supported by a specific physical ISA. The example we are choosing here is our GEMM kernel and its support for bfloat16 and int8 on architectures which don't support these novel ISA SIMD-extension.}

\begin{figure}[t!]
\centering
\includegraphics[width=\columnwidth]{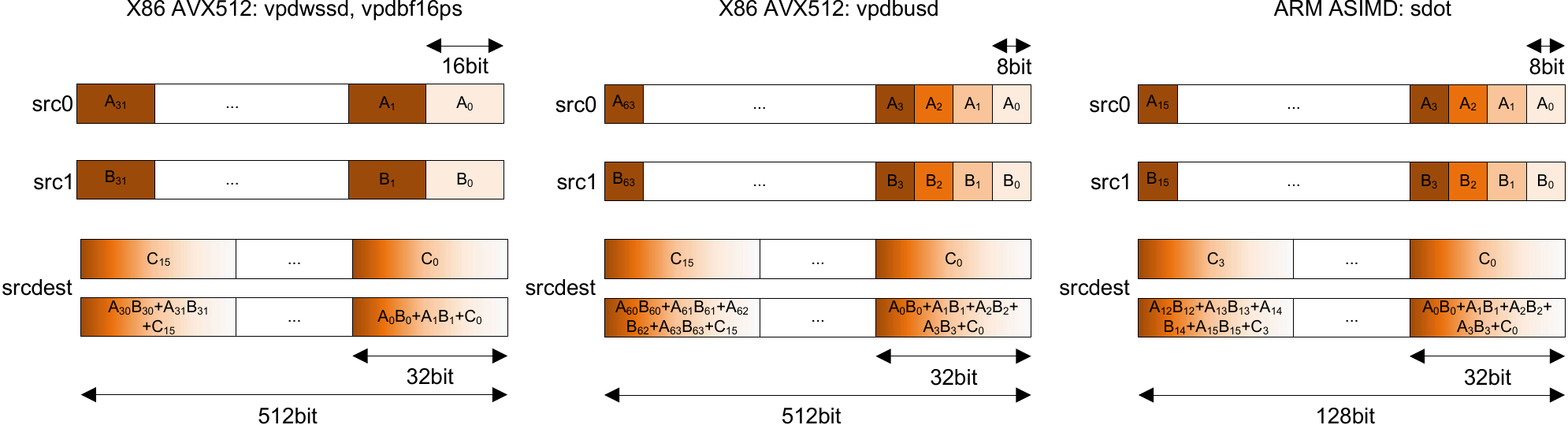}
\caption{Mixed-precision dot-product instructions, \emph{Left}: 16\,bit integer and bfloat16 on Intel AVX512, \emph{Middle}: 8bit integer using Intel AVX512, \emph{Right}: 8\,bit integer using ARM ASIMD.}
\label{fig:vnni_instr}
\end{figure}

\newtext{Before going into the details of the emulation, we first need to introduce special memory layouts which are used by x86 and aarch64  mixed-precision dot-product instructions as shown in Figure~\ref{fig:vnni_instr}. As we can see in all cases (x86/aarch64 and bf16/int8), the overall concept is identical: Although doing mixed-precision and mixed-datatype-length computations, these instructions are functioning from a matrix multiplication point-of-view similar to 32\,bit instructions. This is achieved by having an implicit 2-wide (BF16/int16) and 4-wide (int8) dot-product of $A_i$ and $B_i$  values leading to a horizontal summation per single 32\,bit $C_i$, e.g. $C_0 = A_0 \cdot B_0 + A_1 \cdot B_1 + A_2 \cdot B_2+ A_3 \cdot B_3 + C_0$ as shown for the int8 variant. If we apply blockings with these instructions as discussed in Figure~\ref{fig:brgemm_micro}-Left and Figure~\ref{fig:brgemm_micro}-Middle, then we realize that matrix $B$ is still read via 32-bit broadcast (containing 2 16-bit or 4 8-bit values along the inner-product or common dimension). However, matrix $A$ is in need of reformatting. This is due to the fact that the GEMM kernel in Figure~\ref{fig:brgemm_micro}-Left and Figure~\ref{fig:brgemm_micro}-Middle requires full SIMD-width contiguous loads for optimal performance (which is along $M$ and not $K$). Therefore, we need to reformat $A$ into $[K^o][M][K^i]$ with $K^o \cdot K^i = K$ and $K^i = 2$ for 16-bit and $K^o=4$ for 8-bit inputs. We refer to such a format as \emph{VNNI-format} throughout this paper. After such reformatting of $A$, we can perform full SIMD loads on $A$; combined with the 32-bit broadcast loads on $B$ we have a 32-bit GEMM kernel which has a shorter $K$ dimension, 2$\times$ for 16-bit datatypes and 4$\times$ for 8-bit datatypes.}

\begin{figure}[t!]
\centering
\includegraphics[width=\columnwidth]{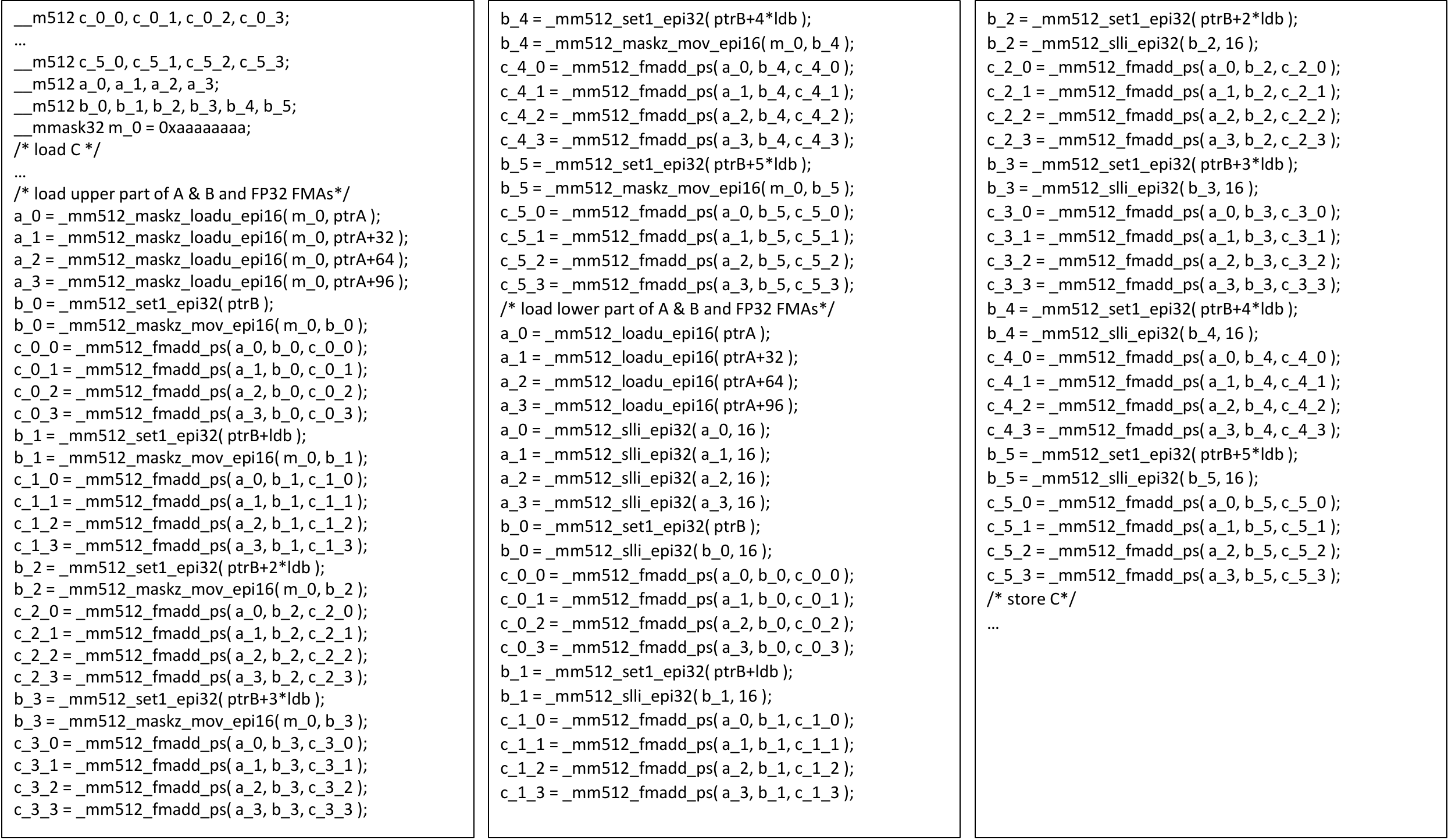}
\caption{Emulation of a bit accurate GEMM kernel using AVX512F instructions matching a GEMM kernel as depicted in Figure~\ref{fig:brgemm_micro} using vdpbf16ps AVX512 instructions. The glossary contains detailed descriptions of the used intrinsic functions.}
\label{fig:emu_bf16_gemm}
\end{figure}

\newtext{In case these novel instructions are not available, especially for bfloat16 as this is a relatively new format, one might think, that an efficient mapping to a classic FP32 SIMD ISA is not possible. This is correct as long as the machine does not offer int16 support. However, with int16 support and SIMD masking we can implement the aforementioned non-trivial mixed-precision dot-product efficiently and even bit-accurately as shown in Figure~\ref{fig:emu_bf16_gemm}. This is done by processing $K^i$ in two rounds in the case of bfloat16 datatype. As shown in Figure~\ref{fig:emu_bf16_gemm} we first process the odd (or upper) bfloat16 number. This is done by exploiting the fact that a bfoat16 number perfectly aliases with an FP32 number in its 16 MSBs. Therefore, on AVX512 we can just execute a full SIMD load as a 16-bit-typed load with masking. As a mask we chose \texttt{0xaaaaaaaa} and as masking-mode we use zero masking. With this trick we automatically turn on-load the upper bfloat16 numbers in $A$ into 16 valid FP32 numbers, and for $B$ we broadcast and then perform an overriding register move. A little bit more work is needed for the lower/even bfloat16 number: In this case we perform an unmasked load and then we use a 32-bit integer shift by 16 to create valid FP32 numbers. A simple inspection of the instruction sequence in Figure~\ref{fig:emu_bf16_gemm} shows that we are mainly executing fused-multiply-add instructions with little overhead compared to a classic FP32 GEMM as illustrated in Figure~\ref{fig:brgemm_micro}-Left and Figure~\ref{fig:brgemm_micro}-Middle. Therefore, we can execute a bfloat16 GEMM with a reformatted matrix $A$ with close to FP32-peak and still benefit from the smaller memory footprint (and therefore a small performance gain, as we will show later in section~\ref{sec:results}). Replacement sequences for int16 and int8 matrix inputs can be carried out in a similar way and their detailed discussion is skipped here.}

\newtext{In addition to the presented emulation of mixed-precision GEMM kernels using SIMD instructions, we have also added support for emulation of Intel AMX instructions bit-accurately on AVX512. This addition enables running numerical accuracy experiments, such as convergence studies, before the release of a chip that supports Intel AMX instructions. A similar path is possible for ARM's SME instruction set and subject to future work. These emulation capabilities further highlight the aspect of TPP as a virtual tensor ISA.} 

\subsection{Examples of Non-Trivial Non-GEMM TPPs}

The previous sections covered most of the TPP implementations: straightforward element-wise unary/binary/ternary operations and various forms of mixed precision GEMMs including their emulation on older hardware. However, there are cases in which we are not operating on the data in an element-wise fashion, e.g. transpose, or the Unary\_op, Binary\_op or Ternary\_op is not an elementary operation. The goal of this section is to shed some light on these cases by presenting the transpose TPP in detail, and sketching fast non-linear approximations on SIMD machines that match the accuracy requirements of deep learning applications.  

\subsubsection{Transform-Transpose TPP via Shuffle Networks}
\label{subsec:transform}

\begin{figure}[t!]
\centering
\includegraphics[width=\columnwidth]{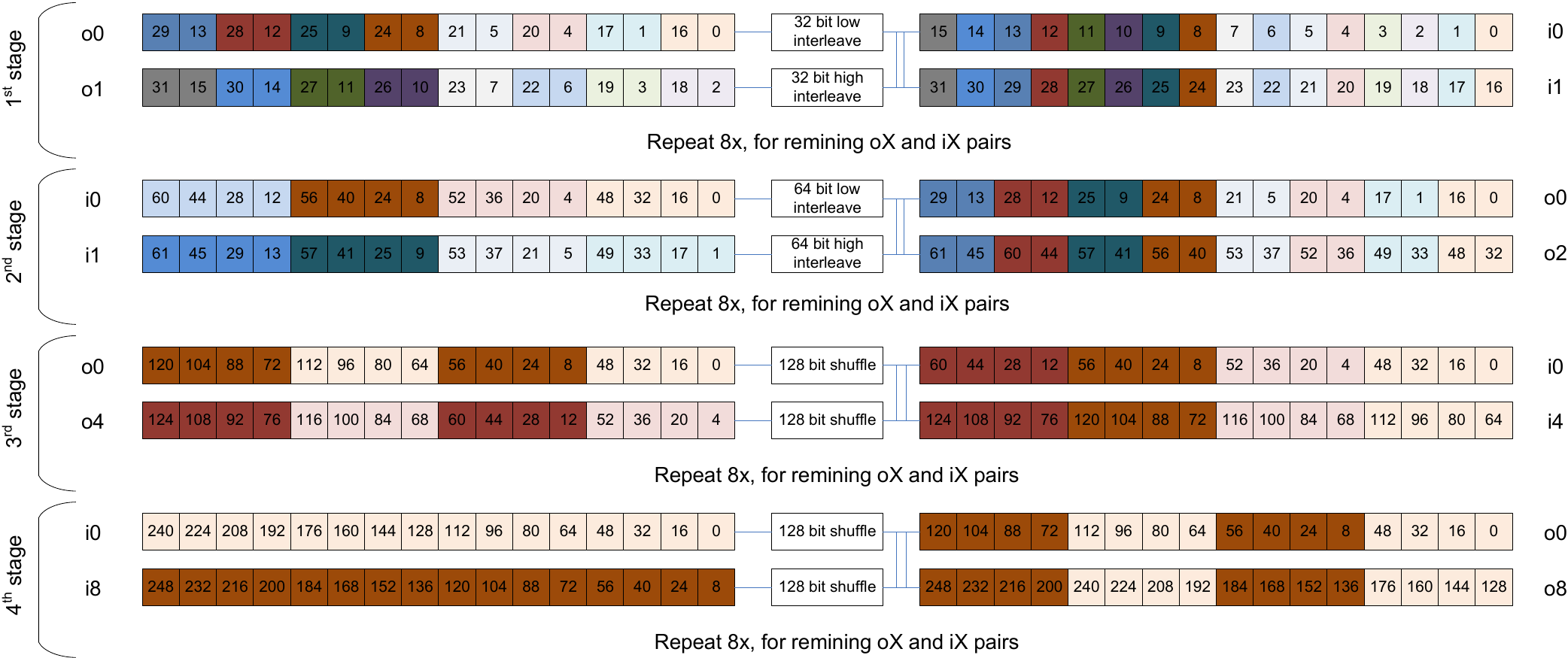}
\caption{Sketch of a shuffle network for a 32-bit transpose of a 16$\times$16 matrix using Intel AVX512 instructions. Via 4 stages (each one having 16 independent shuffles that double in width per stage), the 16$\times$16 matrix (256 elements) can be transposed with only 64 instructions and fully leverages the 32 architectural registers.}
\label{fig:transpose_shuffle_network}
\end{figure}

\begin{figure}[t!]
\centering
\includegraphics[width=\columnwidth]{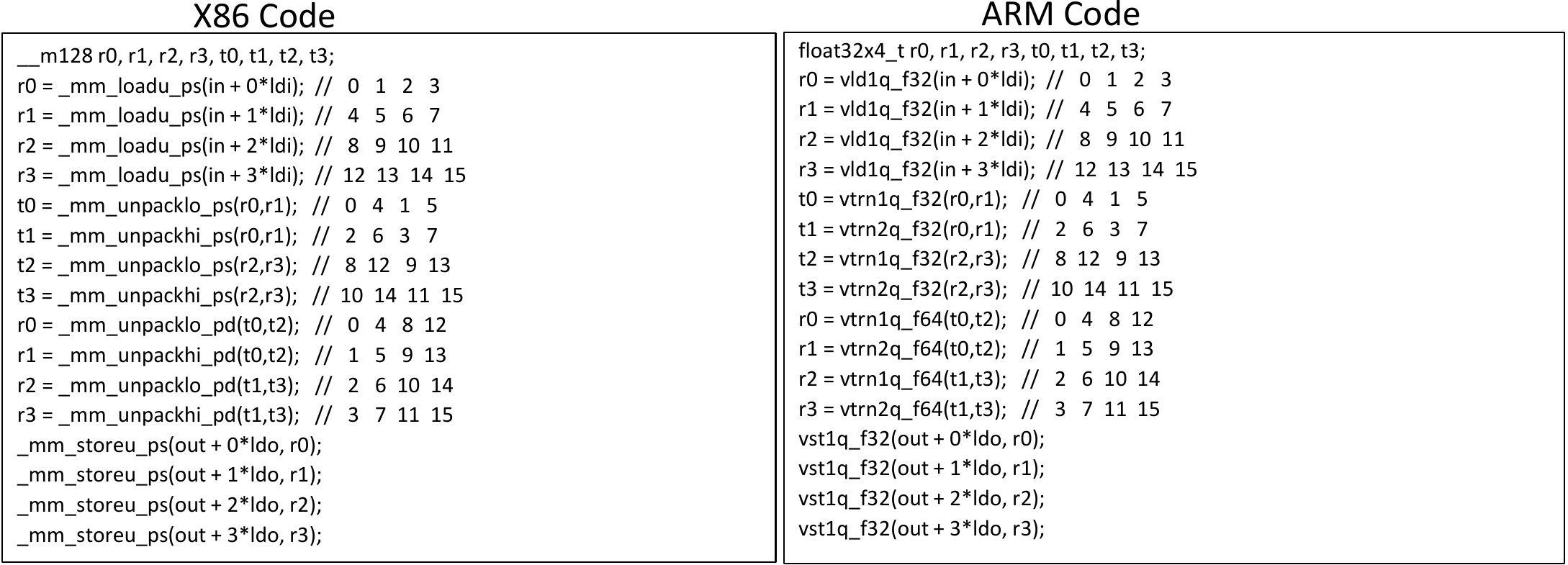}
\caption{Comparison of X86 and ARM code for a simple 4$\times$4 single precision transpose using unpack instructions. The glossary contains detailed descriptions of the used intrinsic functions.}
\label{fig:transpose_code}
\end{figure}

\newtext{When working with matrices, the transpose kernel is ubiquitous. It is needed to access the matrix's elements in various contractions along the mathematically correct dimension. However, a transpose operation is scalar at first sight. In this subsection we exhibit how transpose can be implemented using shuffle networks in a fully vectorized fashion, e.g. Figure~\ref{fig:transpose_shuffle_network} demonstrates how a 16$\times$16 matrix with 256 32-bit elements can be transposed in 64 cycles using AVX512 instructions.}

\newtext{The shuffle-network presented in Figure~\ref{fig:transpose_shuffle_network} is a blueprint for all datatype-lengths and ISAs: in $\log_{2} \text{\it{SIMD-Length}}$ stages we can transpose a matrix held in a set of SIMD registers. In this particular example, we need $\log_{2} 16 = 4$ stages and in each stage we increase the shuffling/interleaving width of logical elements, and also increase the distance at which we access the 32 registers grouped into two sets of 16 registers each. More specifically, we start with registers $i_0$ to $i_{15}$ and interleave elements at the same position in a pair of registers close to each other. This constructs now pairs of 32\,bit values in $o_0$ and $o_1$ which are already containing the transpose's result for 2 out of 16 elements and we repeat this for all other 7 input register pairs. The analogous transformation is now repeated in the second stage with 64-bit values and accessing $o_0$ and $o_2$ as input pair pattern. This constructs a new set output registers $i_0$ and $i_1$ which are holding the transpose's result at 128-bit granularity. After that, stage 3 is shuffling at 128-bit granularity on register pairs which have a distance of ``4" and creates output registers that hold 256-bit of transposed data. Finally, in stage 4, these 256-bit transposed input registers are shuffled once again creating the final set of 16 register holding the transposed 16 $\times$ 16 matrix. For non-square matrices we a) just use masked loads or set registers to zero, b) transpose the zeros as well, and then c) don't store all registers or employ masked stores. This basic kernel is used as a basic building block to create large transpose operators by simply adding outer loops.}

\newtext{This algorithm can be implemented by any SIMD ISA which offers support for picking arbitrary values from a pair of SIMD registers to construct a result register containing values from the two sources, i.e.\ a general shuffler. However, ``structured" shuffle instructions are adequate as shown in Figure~\ref{fig:transpose_code}. Both x86 and aarch64 offer instructions exactly implementing the needs for 32-bit and 64-bit interleaves as needed in the first two stages covered in the previous description. In the case of 128-bit-wide SIMD registers this is enough to carry out the entire transpose of 4 $\times$ 4 matrices as shown in Figure~\ref{fig:transpose_code}.}

\newtext{Finally, we want to note that broadcast loads, as supported by various ISAs, can be used to implement the first stage of the shuffle network. This has the advantage that one stage of the shuffle network can be executed faster and in parallel to the shuffler. The shuffle operations needed in all of these networks are relatively expensive in hardware, therefore modern CPUs often only provide one execution unit for such operations (such ``shuffle-viruses" like transposes are pretty rare in general code). However, broadcasts on the load path are cheap and can run in parallel to the shuffle unit, hence the overall performance of the transpose operation improves. This microkernel variation leads to relatively complex code, and as such we skip its presentation. However our TPP implementation backend employs all these microkernel variations.}

\subsubsection{Approximations for non-linear TPP Activation Functions}
\label{subsec:approx}
\newtext{ Activation functions are used to represent non-linear behavior of neural networks. Popular known activation functions are sigmoid, tanh and GELU.  These activation functions can be  approximated to increase the efficiency of deep learning networks without effecting it's non-linear characteristics. 
In this section we will discuss different approximation techniques based on Pad\'e rational polynomials, piecewise minimax polynomials and Taylor expansions, along with their TPP implementation on different ISAs. For simplicity we present the relevant algorithms in terms of x86 and arm intrinsics (see glossary for the semantics of these intrinsics), however the actual TPP implementation relies on JIT code generation.}

\noindent {\bf Rational Pad\'e  polynomials}

\begin{figure}
\centering
\includegraphics[width=\columnwidth]{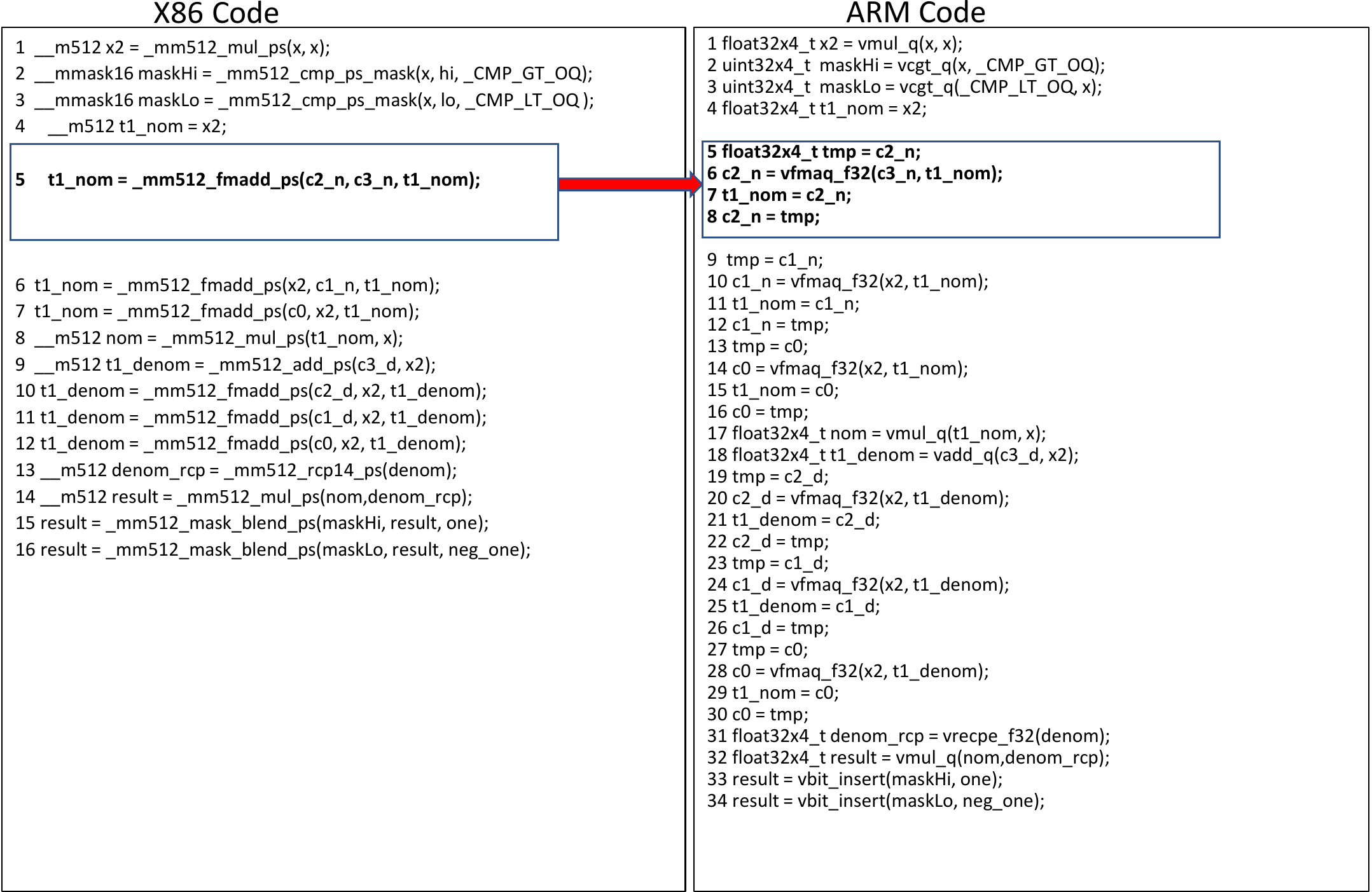}
\caption{\newtext{Rational Pad\'e 7/8 tanh approximation pseudocode with equivalent intrinsics on x86 and Arm/AArch64. We highlight here how the FMADD instruction on x86 ISAs has an equivalent instruction sequence on AArch64.}}
\label{fig:pade 7/8}
\end{figure}

\newtext{The Pad\'e approximation of a function $f$ is the ratio of two polynomials with degrees p and q:}
\begin{equation*}
\newtext{Pad\acute{e}_{[p/q]f}(x)=\frac{\sum_{i=0}^{p}a_ix^i}{\sum_{i=0}^{q}b_ix^i}}
\end{equation*}
\newtext{The coefficients $a_i$ and $b_i$ can be calculated by considering the first $p+q$ derivatives of $f$ at zero and  solving the corresponding system of equations:
\begin{align*}
f(0) =& Pad\acute{e}_{[p/q]f}(0) \\
f^\prime(0) = & Pad\acute{e}^\prime_{[p/q]f}(0) \\
\newtext{\vdots} \\
f^{(p+q)}(0) =& Pad\acute{e}^{(p+q)}_{[p/q]f}(0)
\end{align*}}

\newtext{As an example we consider the approximation of the tanh function which has two asymptotes, hence approximating it with a Taylor expansion of lower degree polynomials may not yield good results. The implementation of the $Pad\acute{e}_{[7/8]}(x)$ tanh approximation is shown in Figure \ref{fig:pade 7/8}. FMA operations are used to compute the numerators and denominators via Horner's rule. The reciprocal of the denominator is multiplied by the numerator to get the final result. The accuracy of reciprocal instruction is different among different CPU's. This difference in accuracy does not affect the non-linear region of the tanh function, keeping the TPP behavior same across different CPU's. The sigmoid activation function can be approximated via tanh by leveraging the following identity:}
\newtext{
\begin{equation*}
sigmoid(x) = (\tanh(x/2)+1)/2
\end{equation*}
}

\noindent {\bf Piecewise minimax polynomial approximations}

\begin{figure}
\centering
\includegraphics[width=\columnwidth]{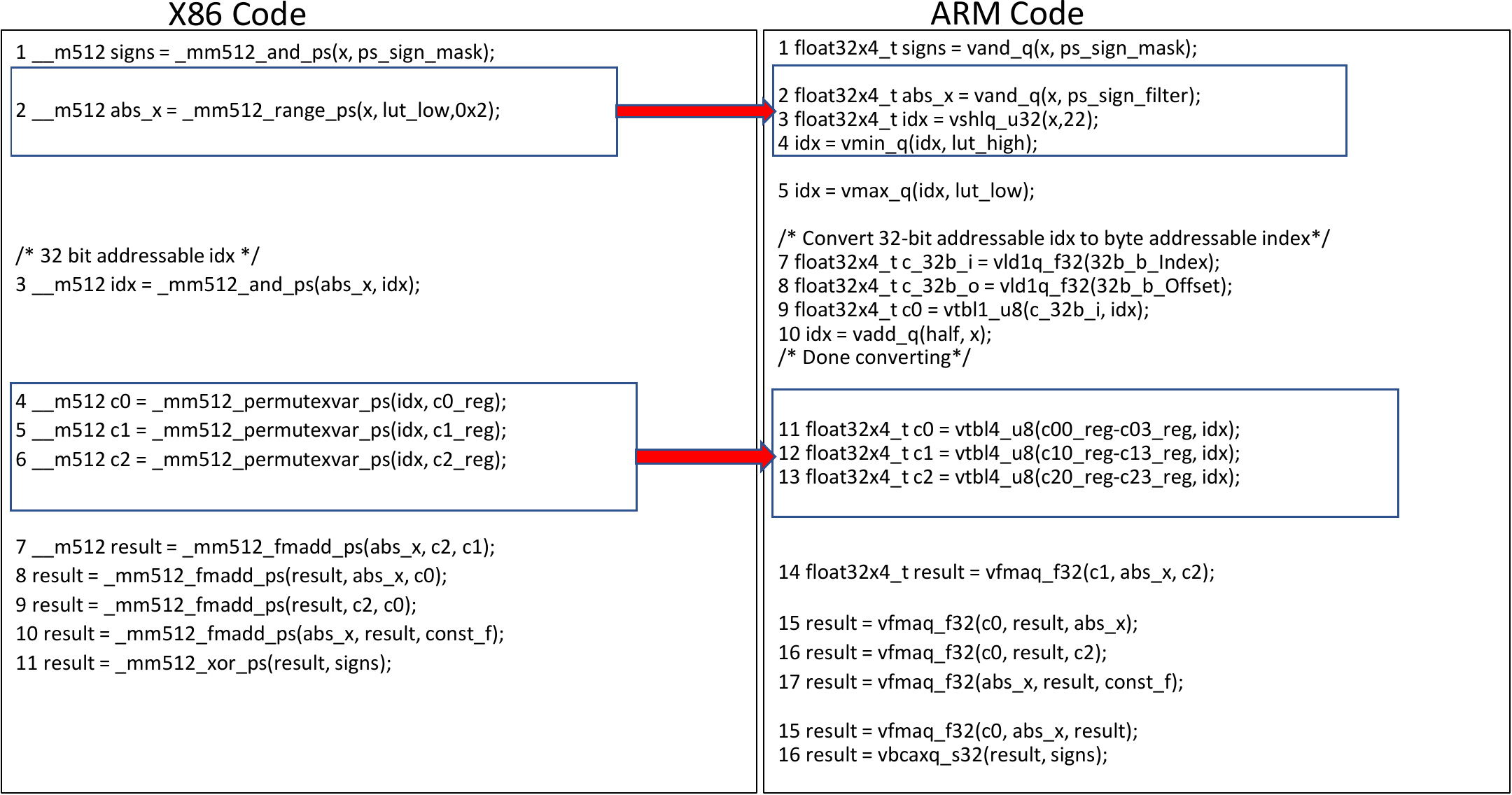}
\caption{Tanh minimax approximation pseudocode with equivalent intrinsics on x86 and Arm/AArch64. We highlight here how the \_mm512\_range\_ps instruction on x86 ISAs has an equivalent instruction sequence on AArch64. Also the permutes on x86 have equivalent Table lookup instructions on AArch64.}
\label{fig:Minmax}
\end{figure}

\newtext{In this section we discuss the minimax polynomials approach~\citep{powell1981approximation} with the truncated Chebyshev series~\citep{chb} for approximations of activation functions. In this approach, the input range of a function $f(x)$ is divided into intervals and for each interval $[a,b]$  we find  a polynomial $p$ of degree max $n$ to minimize:}
\begin{equation*}
\newtext{\max_{a\leq x \leq b}|f(x)-p(x)|}
\end{equation*}

\begin{figure}
\centering
\includegraphics[width=\columnwidth]{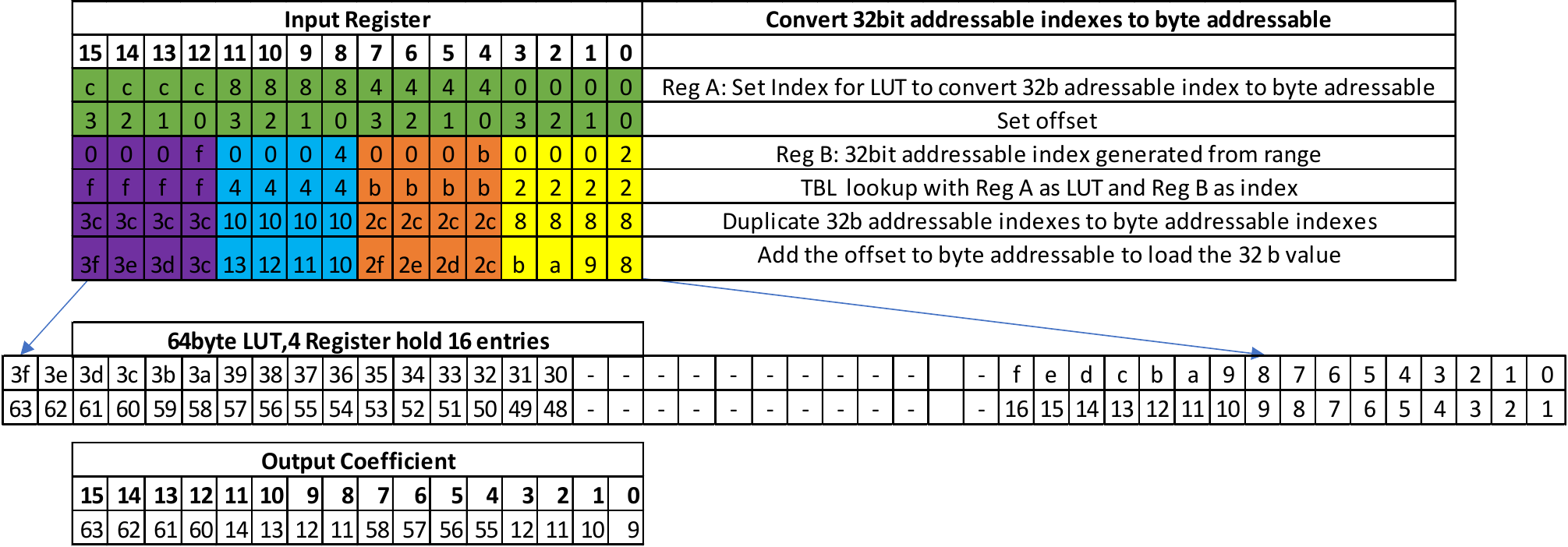}
\caption{Byte addressable Table look up setup in ARM/AArch64. We highlight the conversion of 32bit indexes to byte indexes and the use of byte indexes to get the coefficients in 16 FP32 intervals.}
\label{fig:armlut}
\end{figure}

\begin{figure}
\centering
\includegraphics[width=\columnwidth]{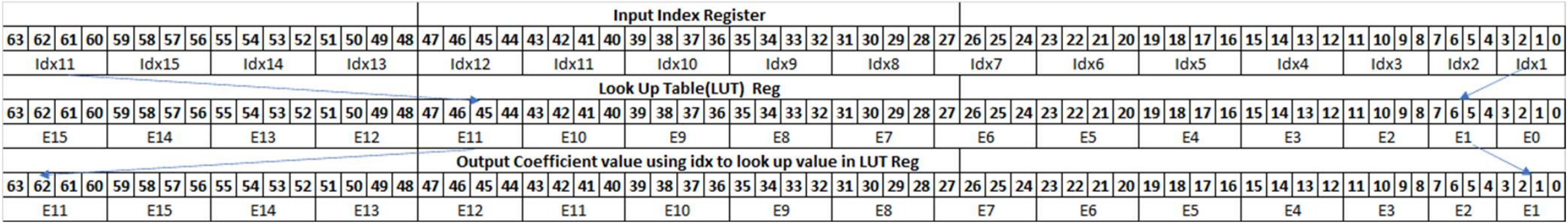}
\caption{32Bit Addressable Table look up setup on x86 AVX512 platforms.}
\label{fig:x86lut}
\end{figure}

\newtext{We approximate tanh and GELU activation functions using this approach in our TPP implementation. The input range is divided into 16 intervals and for each interval we investigate a polynomial $p$ of $3^{rd}$ degree (i.e.\ we find appropriate $p$'s coefficients c0, c1, c2 based on the minimized absolute maximum difference of $f$ and $p$). Figure~\ref{fig:Minmax} shows the x86 and arm implementation of evaluating such minimax polynomials. The register index (idx) is calculated using the exponent and MSB of the respective input values, and represents the 16 intervals where the input values are located.  The range intrinsic \_mm512\_range\_ps(A,B) is used to generate the register index (idx) on AVX512 platforms (Figure~\ref{fig:Minmax}-Left, line 2). In ARM, the range functionality is emulated with equivalent and, shlq, min and max instructions as shown in Figure~\ref{fig:Minmax}-Right, lines 2-4. To evaluate the $3^{rd}$ degree polynomial we need to locate 3 coefficients (c0,c1,c2) based on the values at the register index (idx), which holds 16 entries. We use 3 look up operations to find the three coefficients, each involving 16 FP32 entries. The 512-bit register length in AVX512 is sufficient to hold 16 coefficients required for each look up, resulting in using 3 registers for 3 look up operations (see Figure~\ref{fig:Minmax}-Left, lines 4-6). Each ARM 128-bit wide vector register can only hold 4 FP32 entries, subsequently we are using 12 vector registers to hold the 16 entries for all 3 coefficients of the polynomial. The in-register look-up table is performed using \_mm512\_permutexvar\_ps(A,B) instructions in x86 AVX512 as shown in Figure~\ref{fig:x86lut}. In ARM we have byte addressable table look up instructions which are analogous to 32-bit addressable permutes instructions in x86. Hence, we need to convert the 32-bit addressable (0-16) register indexes to byte addressable (0-64 bytes) indexes. In order to do that, we use a constant register A with a table look up instruction to duplicate the register index (idx) to each byte in the 32-bit entry. A constant offset (0,1,2,3) is added to the duplicated byte index to get the byte addressable index for each FP32 entry in 16 FP32 entries (Figure~\ref{fig:Minmax}-Right, lines 7-9). The table look up instruction in ARM provides the 64 byte look up capability, which is sufficient enough to search into 4 registers holding the 16 entries of each coefficient; we are using the generated byte indexes as shown in Figure~\ref{fig:armlut}. Finally, 4 FMA operations are used to evaluate the polynomial using Horner's rule. The FMA instruction in x86 provides the user the flexibility to decide among the sources to destroy and the ones to preserve. ARM requires mov instructions to save intermediate results in order to avoid the data overwriting during FMA operations}.

\noindent {\bf Approximation with Taylor series}

\begin{figure}
\centering
\includegraphics[width=\textwidth,height=\textheight,keepaspectratio]{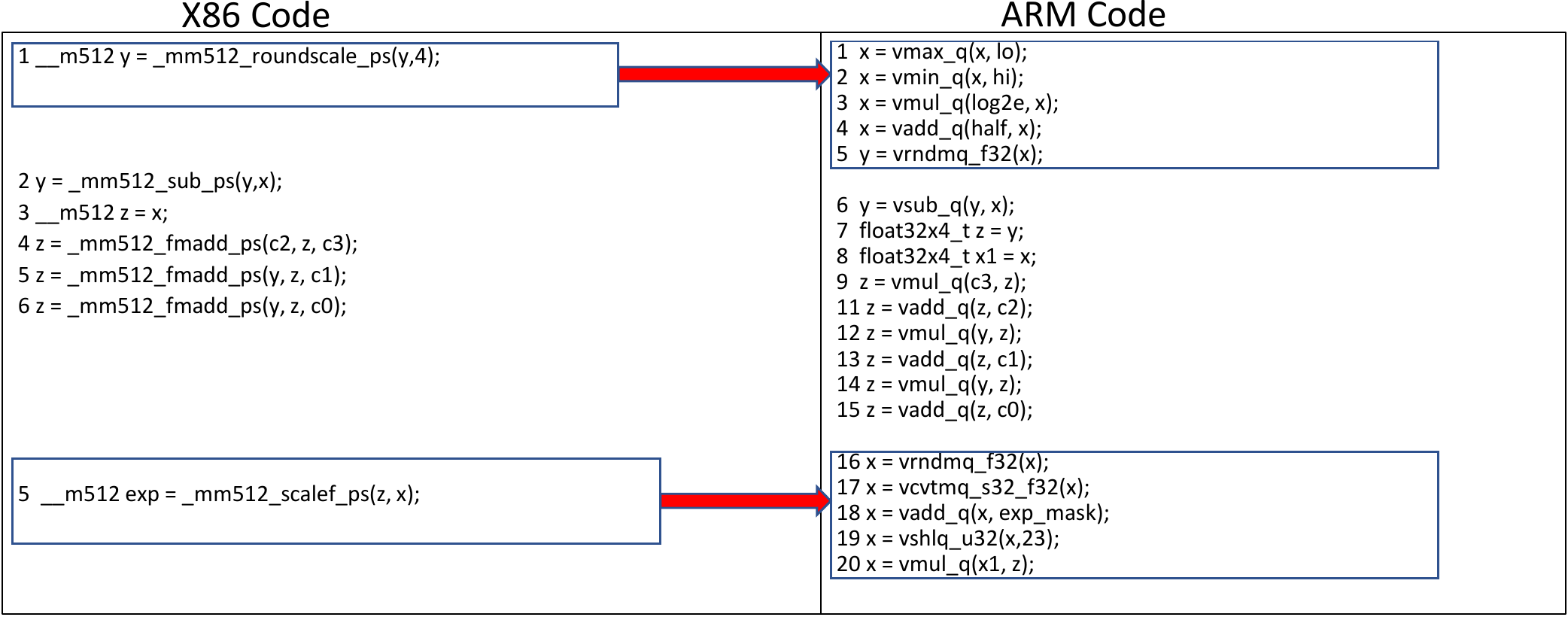}
\caption{Pseudocode for $e^x$  approximation with Taylor series on AVX512 x86 and ARM.}
\label{fig:exp}
\end{figure}

\newtext{As an example of approximation with Taylor series we illustrate here the exp() activation function. The $e^x$ is approximated using the identity $e^x = 2^{x\log_2e}=2^{n+y}=2^n\cdot2^y$ with $n=round(x\log_2e)$ and $y=x\log_2e -n$.   We need to calculate $2^n$ with $n$ being an integer and the term $2^y$ with $|y| \in [0,1)$. A Taylor polynomial of third degree is used to calculate the term $2^y$ with 3 FMA instructions (see Figure~\ref{fig:exp}-Left, lines 4-6). Once $2^y$ is calculated, we leverage the instruction \texttt{\_mm512\_scalef\_ps(A,B)} which returns a vector register holding $a_i\cdot 2^{floor(b_i)}$ for each $a_i\in A$ and $b_i\in B$. This scale instruction  concludes the exp() approximation on x86 with AVX512. On ARM we calculate $2^n$ and $2^y$ with equivalent replacement instructions as shown in Figure~\ref{fig:exp}}.

\section{TPP Matrix Equations }
\label{sec:tpp_eqs}
One of the main design principles of TPPs (as described in Section~\ref{subsec:principles}) is that they can be composed in a producer-consumer fashion to form complex operations. For example consider the scenario where a user wants to implement the composite operation $C = Tanh(A+B)$. One way to express this via TPPs would be to allocate an intermediate tensor $tmp$ with same shape as $A$ and $B$, and perform first $tmp=Add(A,B)$ via the binary Add TPP. Then the user can compute the final result by leveraging the Tanh Unary TPP: $C = Tanh(tmp)$. Even though this approach is functionally correct, it requires the explicit management of intermediate tensors/buffers by the user and also may result in low performance since there are redundant loads/stores to the $tmp$ tensor.

In order to increase the productivity, efficiency and expressiveness pertaining to composite operators, we implemented an embedded Domain Specific Language (eDSL) in LIBXSMM~\citep{libxsmm}. Our Proof-Of-Concept implementations allows the user to express the desired composite operator as a Matrix Equation. More specifically, the user can express the composite operator as an equation tree, where the head and internal nodes are the available TPPs, whereas the leaves of the tree are the input 2D tensors of the composite operation. \newtext{In the next subsections we describe in detail the methodology we employ for JITing matrix equations of TPPs.}

\subsection{\newtext{Definitions and notations for TPP Matrix Equations}}
\begin{figure}[t!]
\centering
\includegraphics[width=\columnwidth]{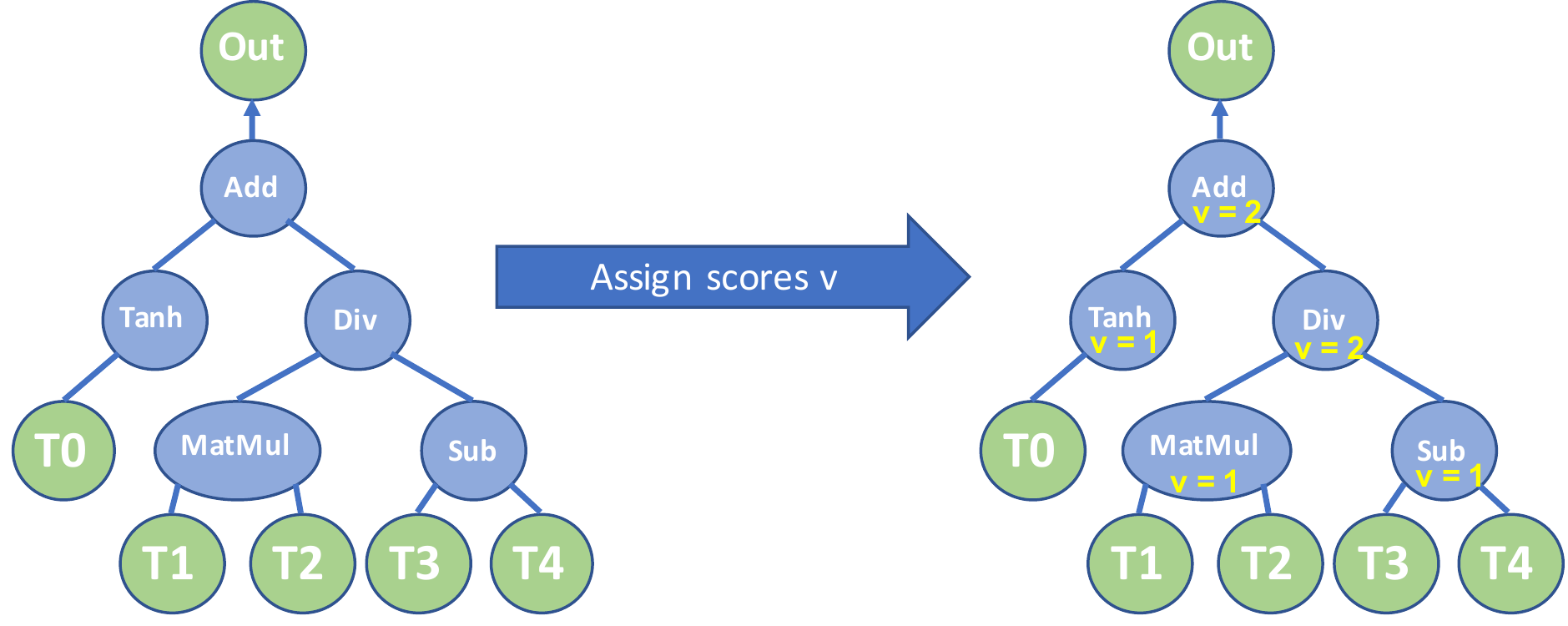}
\caption{Left: TPP Equation tree for $Out = Tanh(T_0) + (T_1 \times T_2) / (T_3 - T_4)$. Right: Assigned register scores $v$ on the equation TPP nodes after running Algorithm~\ref{alg:register_score}.}
\label{fig:example_eq}
\end{figure}
\newtext{A TPP matrix equation is represented as a tree with unary/binary/ternary TPP operations as internal nodes and the equation's input tensors are the leaves of the tree. The inputs of a TPP tree node are essentially its children in the equation tree. The output of an internal TPP node can be represented as a \emph{temporary} intermediate tensor which in turn can be fed as input to the parent TPP node in the tree. Depending on the TPP node type (unary/binary/ternary), each internal node requires a number of inputs (one/two/three) to be computed/ready before performing the corresponding TPP operation. Let's consider for example the TPP equation tree in Figure~\ref{fig:example_eq}-Left that is used to express the following operator:
\begin{equation}
Out = Tanh(T_0) + (T_1 \times T_2) / (T_3 - T_4)
\label{eq:example_eq}
\end{equation}
We will illustrate with this example how our eDSL for TPP Matrix Equations works.}

\subsection{\newtext{Optimized Execution plan for TPP Matrix Equations}}
\newtext{The equation tree in Figure~\ref{fig:example_eq}-Left can be naively evaluated by assigning to each intermediate node a temporary tensor to hold the corresponding TPP output, and performing e.g. 1) the Tanh operation, 2) the Matrix Multiplication, 3) the Subtract operation, 4) the Div operation , and finally 5) the Add TPP. In such an evaluation schedule we would need 4 intermediate tensors to hold the corresponding intermediate results. In this subsection we illustrate how we can construct optimized execution plans for TPP Matrix Equations that minimize the number of intermediate tensors.}

\newtext{For each TPP node $r$ we can assign a \emph{register score} value $v_r$ that essentially dictates how many temporary/intermediate tensors are required to calculate the subtree in the equation where node $r$ is root. We extend the methodology of ~\cite{flajolet1979number} and we generate the register score values using the recursive Algorithm~\ref{alg:register_score}. This algorithm calculates recursively the register scores of the children for a given node $r$, and in this way we know how many temporary tensors are required for the evaluation for each child. Now, if all of its children have the same register score, the node $r$ get an increased register score value, otherwise the node gets as register score the maximum of its children's register score values. Intuitively this means that we can first evaluate a child $c$ and its subtree with whatever intermediate tensor requirements it has, e.g.\ $v_c$ temporary tensors, and eventually we need only one temporary tensor to hold $c$'s output. We can do the same afterwards for all other siblings of $c$, however we can reuse/recycle the rest $v_c-1$ temporary tensors that were required by $c$ since $c$ and its subtree have been already computed.}

\newtext{This algorithm optimizes the number of temporary tensors/storage that are required for the equation evaluation, and it reuses the temporary storage as much as possible. For instance, for the equation in Figure~\ref{fig:example_eq}-Left, after executing Algorithm~\ref{alg:register_score} on the TPP equation tree, we see that the root's register score value is 2 (see Figure~\ref{fig:example_eq}-Right), meaning that only 2 intermediate tensors are required to evaluate the entire TPP tree rather than naively assigning one temporary tensor to each internal TPP node which would result in 4 intermediate tensors.} 

\begin{algorithm}[H]
\begin{algorithmic}[1]
\LState \textbf{Input}:\ TPP equation tree with root node $r$
\LState \textbf{Output}:\ TPP equation tree with assigned register score values on its nodes
\If {is\_Leaf( $r$ )}
\State $v_r$ $\leftarrow$  0 
\EndIf
\If {$r$ is \textbf{unary TPP}}
\State Assign\_Register\_Score( Left\_Child( $r$ ) )
\LineComment{If child is leaf, then we assign current register score of 1, else we assign the child's register score}
\If { is\_Leaf( Left\_Child( $r$ ) )}
\State $v_r$ $\leftarrow$  1
\Else
\State $v_r$ $\leftarrow$  Register\_Score(Left\_Child( $r$ ))
\EndIf
\EndIf
\If {$r$ is \textbf{binary TPP}}
\State Assign\_Register\_Score( Left\_Child( $r$ ) )
\State Assign\_Register\_Score( Right\_Child( $r$ ) )
\LineComment{If the register scores of children are equal, then we get the children's register score increased by one, otherwise we get the max value of the children's register score}
\If { Register\_Score(Left\_Child( $r$ )) \textbf{equals} Register\_Score(Right\_Child( $r$ ))}
    \State $v_r$ $\leftarrow$  Register\_Score(Left\_Child( $r$ )) + 1
\Else
    \State $v_L$ $\leftarrow$  Register\_Score(Left\_Child( $r$ ))
    \State $v_R$ $\leftarrow$  Register\_Score(Right\_Child( $r$ ))
    \State $v_r$ $\leftarrow$  MAX($v_L$ , $v_R$)
\EndIf
\EndIf
\If {$r$ is \textbf{ternary TPP}}
\State Assign\_Register\_Score( Left\_Child( $r$ ) )
\State Assign\_Register\_Score( Middle\_Child( $r$ ) )
\State Assign\_Register\_Score( Right\_Child( $r$ ) )
\LineComment{If all children are leaves, then we assign current register score of 1, otherwise we get the max value of the children's register score}
\If { is\_Leaf( Left\_Child( $r$ ) ) \textbf{AND} is\_Leaf( Middle\_Child( $r$ ) ) \textbf{AND} is\_Leaf( Right\_Child( $r$ ) )}
\State $v_r$ $\leftarrow$  1
\Else
\State $v_L$ $\leftarrow$  Register\_Score(Left\_Child( $r$ ))
\State $v_M$ $\leftarrow$  Register\_Score(Middle\_Child( $r$ ))
\State $v_R$ $\leftarrow$  Register\_Score(Right\_Child( $r$ ))
\State $v_r$ $\leftarrow$ MAX(3, $v_L$, $v_M$, $v_R$)
\EndIf
\EndIf
\end{algorithmic}
\caption{Assign\_Register\_Score( $r$ )}
\label{alg:register_score}
\end{algorithm}

\begin{figure}[t!]
\centering
\includegraphics[width=\columnwidth]{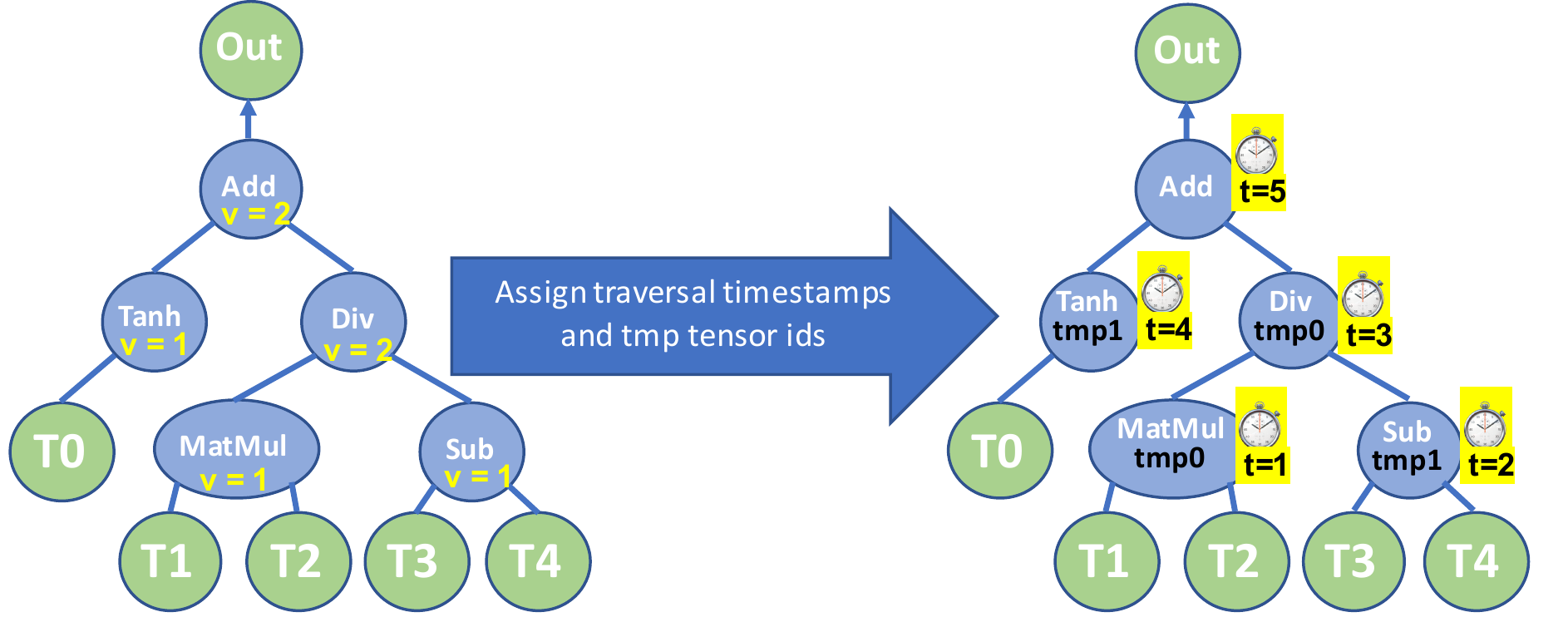}
\caption{Left: TPP equation tree with assigned register scores $v$ on the nodes. Right: TPP equation tree with assigned traversal timestamps $t$ and temporary tensor ids $tmp$ after executing Algorithm~\ref{alg:execution_plan}.}
\label{fig:example_eq_2}
\end{figure}

\newtext{Now that we have assigned the register scores for each node we can devise an execution plan for the TPP equation tree that minimizes the number of required intermediate tensors. Algorithm~\ref{alg:execution_plan} recursively creates such an optimal execution plan and essentially it calculates: 1) the order/traversal timestamps $t$ with which the TPP equation nodes have to be evaluated, and also 2) assigns to each intermediate node $r$ a temporary tensor id $tmp_r$ that holds the intermediate resulting tensor of that TPP node. Figure~\ref{fig:example_eq_2}-Right shows the optimized execution plan by applying Algorithm~\ref{alg:execution_plan} on our example equation. This algorithm recursively visits/evaluates the children of a node $r$ in order of decreasing register score value. This means that the child/subtree with the maximum register score value is evaluated first, one of the temporary tensors is dedicated to hold that child's intermediate output, whereas the remaining temporary tensors can be reused for the evaluation of the siblings/subtrees, which per definition/order of traversal, require less or equal number of intermediate tensors. Such a strategy guarantees that the temporary tensors are optimally reused/recycled, and as a result we can leverage the minimum required temporary tensors for the evaluation of the entire equation TPP tree. For simplicity in our description, we assumed that all intermediate temporary tensors have the same size, however our implementation considers the actual sizes of the intermediate output tensors and takes the maximum one as representative size for all temporary tensors.}

\subsection{\newtext{Implementation of Optimized Execution plan for TPP Matrix Equations}}
\newtext{By employing Algorithm~\ref{alg:execution_plan} we can devise an optimal execution plan for the TPP Matrix equation, and here we describe the implementation of such a plan. We consider three implementation strategies:
\begin{itemize}
\item \emph{Strategy 1}: Using stack-allocated buffers as intermediate temporary tensors
\item \emph{Strategy 2}: Using vector-register blocks as intermediate temporary tensors
\item \emph{Strategy 3}: Hybrid implementation where some intermediate temporary tensors are stack-allocated buffers and some are vector-register blocks
\end{itemize}
So far in our description we have used the abstract notation ``temporary tensor" without specifying how such a temporary tensor is instantiated in the implementation. The exact instantiation of a temporary/intermediate tensor is the differentiation factor among the 3 implementation strategies for the TPP matrix equations.}

\newtext{Strategy 1 considers each intermediate tensor as a physical buffer, and our TPP equation implementation allocates on the stack some space/buffer for each temporary tensor. Then, by following the timestamp order of the optimal execution plan (e.g.\ see Figure~\ref{fig:example_eq_2}-Right), we emit/JIT the corresponding TPP code (e.g.\ see Algorithms~\ref{alg:generic_tpp} and~\ref{alg:br_kernel}) where the input tensors might be either the equation's input buffers provided by the user, or one of the stack allocated buffers representing an intermediate result. The fact that we have minimized the number of intermediate temporary buffers/tensors is critical for performance since these stack-allocated buffers may remain in some level of cache. Such a strategy is generic and can be leveraged to implement arbitrary equations. However, Strategy 1 may suffer from store-to-load forwarding inefficiencies on modern processors. Additionally, some of the intermediate tensors may spill from cache (e.g.\ when the intermediate outputs exceed the corresponding cache capacity) which would make the communication of temporary tensors among TPP nodes via loads/stores from/to stack allocated buffers quite expensive.}

\newtext{Strategy 2 considers each intermediate tensor as an $r_m\times r_n$ vector-register block. For example, on an AVX512 platform with 32 512-bit wide registers we have available 2 KBytes of register file that may be used for intermediate tensors. Each one of such 512-bit wide vector registers can hold 16 single-precision values and by stacking e.g.\ 4 of these we can form a logical 16$\times$4 intermediate tensor and in total we have available $32/4=8$ of such intermediate tensors that could be used by the equation. In Strategy 2 we block the computation of the equation's output in blocks with size $r_m\times r_n$, and we can calculate the corresponding $r_m\times r_n$ output by following the timestamp order of the optimal execution plan. We emit/JIT the corresponding TPP code for sub-tensors with size $r_m\times r_n$ where each intermediate output tensor is the assigned temporary vector-register block. Essentially this strategy performs vertical register fusion within the equation TPP nodes and incurs \emph{no} communication via loads/stores from/to stack allocated buffers. However, such a methodology is limited by the number of available vector registers on each platform.}

\newtext{Strategy 3 combines the strengths of Strategies 1 and 2 by considering some intermediate tensors as stack-allocated buffers and some intermediate tensors as vector-register blocks. As such, in Strategy 3 the TPP operations/subtrees which exhibit \emph{both} high register pressure and reuse (e.g.\  transposes, GEMM/BRGEMM, transcendental approximations), propagate the intermediate results towards the rest of the TPPs in the tree via stack-allocated temporal tensors. On the other hand, TPP subtrees without large register pressure are implemented using Strategy 2 that employs vertical register fusion and avoids loads/stores from/to stack-allocated buffers. }

\newtext{In addition to the aforementioned 3 strategies, in the TPP equation back-end we identify idioms/motifs of combined TPPs (e.g.\ a gather TPP followed by a reduce TPP) and we JIT an instruction sequence which is optimal for the composite access pattern. In subsection~\ref{subsubsec:emb} we show an example of such a combined TPP motif that is optimized by the TPP backend.}

Even though we developed a rudimentary method/POC of combining the TPPs via Matrix Equation Trees, we have found that it is sufficient to express all the complex operators we encountered in a wide-range of workloads discussed further in Section~\ref{sec:workloads}. Nevertheless, we envision that when/if TPPs are widely adopted within Tensor Compiler frameworks (e.g.\  as an MLIR dialect) then more complicated Graphs (instead of simple trees) and more sophisticated analyses/optimization passes can be leveraged during the composition of TPPs. The key-ingredient that makes the composition of TPPs amenable to optimization opportunities is the TPP specification itself: TPPs comprise a small, well-defined compact set of tensor operators with declarative semantics as shown in Section~\ref{sec:specification}.

We would like also to highlight one use-case of Matrix Equations that can be beneficial for specialized DL accelerators. The BRGEMM TPP described in Section~\ref{subsec:brgemm_structure} corresponds to an output-stationary flow that is suitable for CPUs and GPUs. Given an accelerator that favors e.g. $A$-stationary GEMM formulations, one could express the following Matrix Equation: internal nodes $G_i$ would be GEMM ternary TPPs, for each GEMM node $G_i$ we would have the same input leaf $A$ and a varying input $B_i$, and the output of each node would be a result $C_i$. Essentially this formulation dictates an $A$-stationary flow, and the back-end could optimize accordingly for the specific accelerator.

\begin{algorithm}[H]
\begin{algorithmic}[1]
\LState \textbf{Input}:\ TPP equation tree with root node $r$ and assigned register score values on its nodes
\LState \textbf{Output}:\ TPP equation tree with assigned traversal timestamps $t$ and temporary tensor ids $tmp$
\If {is\_Leaf( $r$ )}
\State return 
\EndIf
\If {$r$ is \textbf{unary TPP}}
\State Create\_Execution\_Plan( Left\_Child( $r$ ) )
\State $t_r$ $\leftarrow$  global\_timesteamp++
\LineComment{If child is leaf, reserve a new tmp, else re-use tmp from child}
\If { is\_Leaf( Left\_Child( $r$ ) )}
\State $tmp_r$ $\leftarrow$  Reserve\_Tmp() 
\Else
\State $tmp_r$ $\leftarrow$  tmp\_Left\_Child( $r$ ) 
\EndIf
\EndIf
\If {$r$ is \textbf{binary TPP}}
\LineComment{Recursively visit children in order of decreasing register score}
\State Create\_Execution\_Plan( Child\_Max\_Register\_Score( $r$ ) )
\State Create\_Execution\_Plan( Child\_Min\_Register\_Score( $r$ ) )
\State $t_r$ $\leftarrow$  global\_timesteamp++
\LineComment{If all children are leaves, reserve a new tmp, else re-use the tmp from a non-leaf child and recycle the tmp of the other non-leaf child}
\If { is\_Leaf( Left\_Child( $r$ ) \textbf{AND} is\_Leaf( Right\_Child( $r$ ) ) )}
\State $tmp_r$ $\leftarrow$  Reserve\_Tmp() 
\Else
\If { not\_Leaf( Left\_Child( $r$ ) }
\State $tmp_r$ $\leftarrow$  tmp\_Left\_Child( $r$ )
\State Recycle\_Tmp(  tmp\_Right\_Child( $r$ ) )
\Else
\State $tmp_r$ $\leftarrow$  tmp\_Right\_Child( $r$ ) 
\State Recycle\_Tmp(  tmp\_Left\_Child( $r$ ) )
\EndIf
\EndIf
\EndIf
\If {$r$ is \textbf{ternary TPP}}
\LineComment{Recursively visit children in order of decreasing register score}
\State Create\_Execution\_Plan( Child\_Max\_Register\_Score( $r$ ) )
\State Create\_Execution\_Plan( Child\_Mid\_Register\_Score( $r$ ) )
\State Create\_Execution\_Plan( Child\_Min\_Register\_Score( $r$ ) )
\State $t_r$ $\leftarrow$  global\_timesteamp++
\LineComment{If all children are leaves, reserve a new tmp, else re-use the tmp from a non-leaf child and recycle the tmps of the other non-leaf children}
\If { is\_Leaf( Left\_Child( $r$ ) ) \textbf{AND} is\_Leaf( Middle\_Child( $r$ ) ) \textbf{AND} is\_Leaf( Right\_Child( $r$ ) )}
\State $tmp_r$ $\leftarrow$  Reserve\_Tmp() 
\Else
\If { not\_Leaf( Left\_Child( $r$ ) }
\State $tmp_r$ $\leftarrow$  tmp\_Left\_Child( $r$ )
\State Recycle\_Tmp(  tmp\_Middle\_Child( $r$ ) ) , Recycle\_Tmp(  tmp\_Right\_Child( $r$ ) )
\Else
\If { not\_Leaf( Right\_Child( $r$ ) }
\State $tmp_r$ $\leftarrow$  tmp\_Right\_Child( $r$ )
\State Recycle\_Tmp(  tmp\_Middle\_Child( $r$ ) ) , Recycle\_Tmp(  tmp\_Left\_Child( $r$ ) )
\Else
\State $tmp_r$ $\leftarrow$  tmp\_Middle\_Child( $r$ ) 
\State Recycle\_Tmp(  tmp\_Left\_Child( $r$ ) ) , Recycle\_Tmp(  tmp\_Right\_Child( $r$ ) )
\EndIf
\EndIf
\EndIf
\EndIf
\end{algorithmic}
\caption{Create\_Execution\_Plan( $r$ )}
\label{alg:execution_plan}
\vspace{-0.2em}
\end{algorithm}

\section{TPP-based Kernels \& Workloads}
\label{sec:workloads}
This section covers how DL kernels and workloads (image processing, recommendation systems, natural language processing, graph processing and applications in science) can leverage TPPs to achieve high performance. Although this paper's work is targeting CPUs, we cover the entire training pipeline and not only inference. The main purpose of this is to demonstrate the versatility of TPPs which is valuable in the more complicated backward pass kernels, and to handle training's implications to the forward pass. 

\subsection{TPP-based Kernels}
\label{subsec:standalone_kernels}
\subsubsection{Softmax Kernel}
\label{subsec:smax_kernel}
\label{subsec:equations_structure}
\begin{figure}[t!]
\centering
\includegraphics[width=0.8\columnwidth]{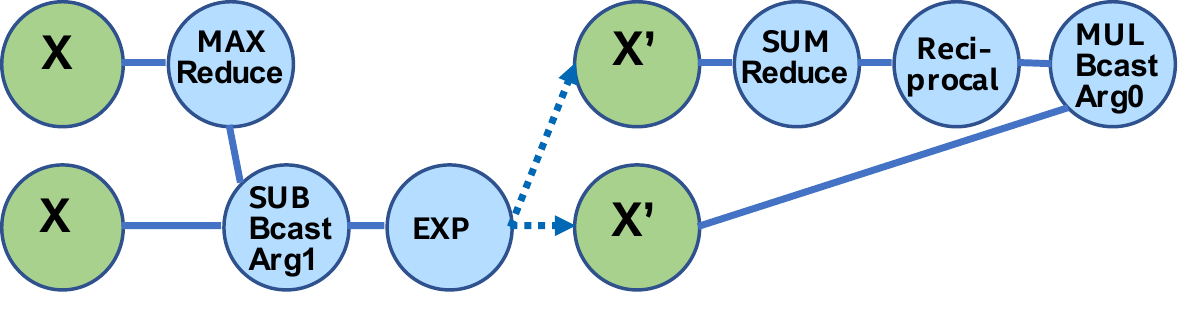}
\caption{Softmax operator by combining TPPs.}
\label{fig:smax_eq}
\end{figure}
 Figure~\ref{fig:smax_eq} illustrates two Matrix Equation trees that are used to express the softmax operator~\citep{gibbs2014elementary}:
\begin{equation}
Y=\text{softmax}(X)\ \text{with}\ y_{ij} = \frac{e^{\left({x_{ij}-\max_{x_{ij}\in X}x_{ij}}\right)}}{\sum\limits_{x_{ij}\in X}e^{\left({x_{ij}-\max_{x_{ij}\in X}x_{ij}}\right)}}
\label{eq:smax}
\end{equation}

Equation~\ref{eq:smax} shows the formula for the softmax operator~\citep{gibbs2014elementary}, which is often used as the last activation function of a neural network, aiming to normalize its output to a probability distribution. We can represent this operator via two TPP equation trees illustrated in Figure~\ref{fig:smax_eq}. The left tree computes the nominator of Equation~\ref{eq:smax}: first the maximum value of the input tensor $X$ is found (via the max-reduce TPP), then we subtract this max value from each entry of $X$ (note the broadcast semantics in the second argument of the subtraction TPP), and a new tensor $X'$ is computed by calculating the element-wise exponent on the earlier subtraction's outcome. Finally, in the right TPP tree each entry of the tensor $X'$  is normalized by the sum of all values in $X'$ to obtain the softmax output, a tensor $Y$. This example illustrates the expressiveness of the TPP abstractions, since the components of the mathematical formula map to TPPs in a straightforward way. At the same time, this example highlights the separation of concerns: the user does not need to worry about the efficient implementation of this equation on each different platform, since the TPP back-end is responsible for optimized code generation which is target-specific (contrary to the TPP expression itself which is platform-agnostic).

\subsubsection{Normalization Kernels}
\newtext{\textbf{Batch normalization (batchnorm)} is a technique ~\citep{ioffe2015batch} that normalizes neuron layer input tensors to improve the overall training process. Batchnorm removes the need for careful parameter initialization and reduces the required training steps~\citep{ioffe2015batch} in the neural networks. 
The batchnorm computations can be divided in two stages: i) First the mean and variance of the input tensor are computed across the ``batch" dimension: 
$\mu_j=\sum_{i=0}^{n-1}x_{ij}$, $\sigma_j^2=\frac{1}{n}\sum_{i=0}^{n-1}(x_{ij}-\mu_i)^2$ where $i$ is the ``batch" dimension and $j$ is the ``feature" dimension, ii) then the tensor entries $x_{ij}$ are normalized based on  $\mu$ and $\sigma$: $x_{ij}^{\prime}=(x_{ij}-\mu_j)/(\sqrt{\sigma_{j}^{2}+\epsilon})$.} 

\begin{figure}[t!]
\centering
\includegraphics[width=0.7\columnwidth]{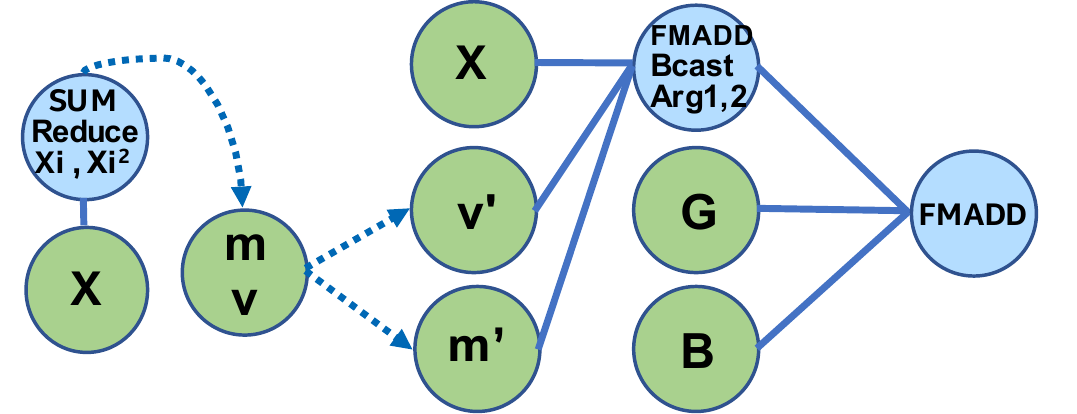}
\caption{Layernorm via TPPs.}
\label{fig:lnorm_eq}
\end{figure}

\newtext{Depending upon the workload, different TPPs and TPP equations can be employed to implement the batchnorm. Here, we take an example of batchnorm on a ResNet50~\citep{he2016deep} convolution layer tensor $X$. The input tensor $X$ has a four-dimensional shape of \{N, C, H, W\} with dimensions of batch ($N$), feature ($C$), height ($H$), and width ($W$). We first use sum-reduce TPPs on $H$ and $W$ dimensions to compute the sum ($m[N, C]$) and the sum of squared elements ($v[N, C]$) matrices. Subsequently, we use binary add TPPs across the batch dimension of $m[N, C]$ and $v[N, C]$ matrices for eventual computation of mean ($\mu[C]$) and variance ($\sigma^2[C]$) vectors. In the next step, we use a scaling equation to normalize each element of the input tensor. The scaling equation $Y = (m^\prime* X + v^\prime)*G + B$ converts the input tensor $X$ into a normalized tensor $Y$. Here, $G[C]$ and $B[C]$ are scaling vector inputs to batchnorm, and $m^\prime[C]$ and $v^\prime[C]$ are intermediate vectors that are computed from mean and variance vectors. We implement the scaling equation by a single TPP equation containing two FMADD ternary TPPs. The second equation tree of Figure~\ref{fig:lnorm_eq} shows an analogous scaling equation implementation. However, for this particular implementation, we broadcast $m^\prime, v^\prime, G, B$ vectors into $H$, $W$, and $N$ dimensions inside the TPP equation tree. An efficient implementation of batchnorm uses blocking on the $C$, $H$, and $W$ dimensions along with multi-threading on the $N$ and feature block dimension. We do not show the details of this implementation for sake of simplicity.}

\label{subsec:layernorm_kernel}
\textbf{Layer normalization (layernorm)}~\citep{ba2016layer} is a technique that normalizes the neurons \emph{within} a layer, and was motivated by the limitations of Batch Normalization~\citep{ioffe2015batch} in Recurrent Neural Networks.
The layernorm computations can be divided in two stages: i) First the mean and variance of the input tensor are computed across the ``feature" dimension: 
$\mu_i=\sum_{j=0}^{m-1}x_{ij}$, $\sigma_i^2=\frac{1}{m}\sum_{j=0}^{m-1}(x_{ij}-\mu_i)^2$ where $i$ is the batch dimension and $j$ is the ``feature" dimension, ii) then the tensor entries $x_{ij}$ are normalized based on  $\mu$ and $\sigma$: $x_{ij}^{\prime}=(x_{ij}-\mu_i)/(\sqrt{\sigma_{i}^{2}+\epsilon})$. Depending on the workload (e.g.\ attention cell in BERT), the scaled tensor may be further scaled with two other tensors $\gamma$ and $\beta$. Figure~\ref{fig:lnorm_eq} illustrates two TPP equation trees that implement this composite layernorm operator. The left equation is using the sum-reduce TPP to compute the sum and sum of squared elements of the input tensor, namely $m$ and $v$. These two scalars are combined (not shown in the equation for simplicity), and are fed as inputs to the right TPP tree, where the FMADD ternary TPP is used to scale the input tensor $X$. Finally, a cascading FMADD ternary TPP computes the final result via the scaling tensors $G$ and $B$. We illustrate this layernorm via means of TPPs since all DL norming layers essentially exhibit similar computational motif, and this specific norm is used in the BERT workload described in subsection~\ref{subsec:bert}.

\newtext{\textbf{Group normalization (groupnorm)}~\citep{wu2018group} is a technique that normalizes the neurons within a group of features. Groupnorm was proposed as an alternative to batchnorm~\citep{ioffe2015batch} to reduce normalization error for smaller batch sizes. In groupnorm, features are divided into groups, and mean and variance are computed within each group for normalization. Groupnorm is also a generalization of the layer normalization~\citep{ba2016layer} and instance normalization~\citep{ulyanov2016instance} approach. Layernorm is groupnorm with a single group, and instance norm is groupnorm with group size equal to one. Groupnorm can be implemented with the same set of TPPs and TPP equations that were used in the batchnorm kernel. We again take the example of ResNet50~\citep{he2016deep} convolution layer tensor $X$ and apply groupnorm on it with $g$ number of groups. We can ignore the batch dimension ($N$) for this discussion as groupnorm works independently upon each batch. Therefore, the input tensor $X$ now has a three-dimensional shape of \{C, H, W\} with dimensions of feature ($C$), height ($H$), and width ($W$). We first use sum-reduce TPPs on $H$ and $W$ dimensions to compute the sum ($m[C]$) and the sum of squared elements ($v[C]$) vectors. Subsequently, we add $m[C]$ and $v[C]$ values within a feature group for eventual computation of group mean ($\mu[g]$) and group variance ($\sigma^2[g]$) vectors. Similar to batchnorm, we use a scaling equation to normalize each element of the input tensor. The scaling equation $Y = (m^\prime * X + v^\prime)*G + B$ converts input tensor $X$ into a normalized tensor $Y$. Here, $G[C]$ and $B[C]$ are scaling vector inputs to groupnorm, and $m^\prime[C]$ and $v^\prime[C]$ are intermediate vectors that are computed from group mean and group variance vectors. The second equation tree of Figure~\ref{fig:lnorm_eq} shows an analogous scaling equation implementation. However, for this particular implementation, we broadcast $m^\prime, v^\prime, G, B$ vectors into $H$ and $W$ dimensions inside the TPP equation tree. We can also apply the same scaling equation to a single group or set of groups with few parameter changes. An efficient implementation of groupnorm uses blocking on the $C$, $H$, and $W$ dimensions. We do not show the details of this implementation for sake of simplicity.}

\subsubsection{BF16 Split-SGD Kernel}
\label{subsec:splitsgd_kernel}
\begin{figure}[t!]
\centering
\includegraphics[width=0.6\columnwidth]{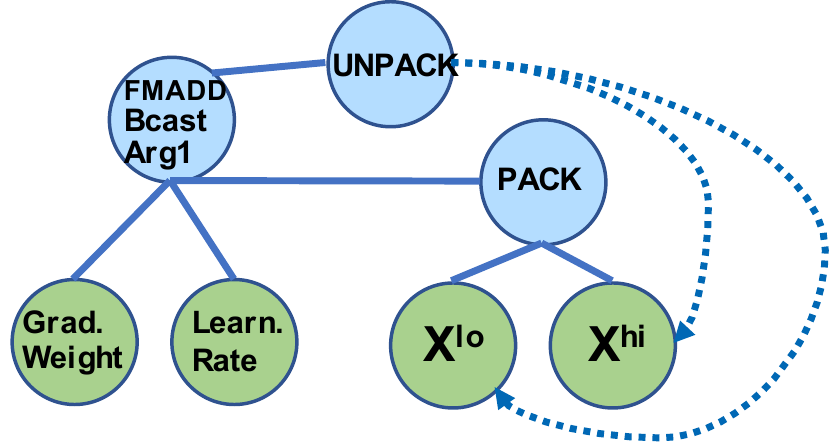}
\caption{BF16 Split-SGD operator by combining TPPs.}
\label{fig:sgd_eq}
\end{figure}
Unlike the previous kernels which are well-established in DL workloads, and as such potentially optimized in DL libraries, we present here an example of a novel operator, which per definition is not existent in DL libraries. BF16 split-SGD was recently introduced in the context of DLRM training with BF16 datatype~\citep{kalamkar2020optimizing}. The Split-SGD-BF16 solver aims at efficiently exploiting the aliasing of BF16 and FP32 (i.e.\ the 16 Most Significant Bits (MSB) on both are identical) in order to save bandwidth during the SGD-solver in training. 
The employed trick is that the weights are not stored as FP32 values in a single tensor. Instead, the FP32 tensors are split into their high and low 16bit-wide parts: the 16 MSBs of the FP32 values, and the 16 LSBs of the same values are stored as two separate tensors $X^{hi}$ and $X^{lo}$ respectively. The 16 MSBs represent a valid BF16 number and constitute the model/weight tensors during training. These BF16 weights are used exclusively in the forward and backward passes, whereas the lower 16 bits are only required in optimizer. More specifically, the $X^{hi}$ and $X^{lo}$ tensors are packed together to form an FP32 tensor, resulting in a fully FP32-accurate optimizer. Figure~\ref{fig:sgd_eq} illustrates the BF16 Split-SGD operator written entirely via TPPs. First the $X^{hi}$ and $X^{lo}$ are packed, and the formed FP32 tensor is used in a cascading FMADD TPP that performs the SGD scaling with the corresponding Gradient Weight tensor and learning rate. Finally, the resulting FP32 tensor is unpacked to the $X^{hi}$ and $X^{lo}$ tensors for further use in the training process.

\subsubsection{Convolutional Neural Network (CNN) kernel}
\label{subsec:cnn}
Convolutional Neural Networks (CNN) consist of layers with multiple neurons connected by weights, and they have been applied with success in image recognition, semantic segmentation, autonomous driving, medical imaging and in an increasing number of scientific applications. Previous work~\citep{sc18,georganas2020harnessing} has shown that CNNs, despite their seemingly complicated loop structure due to the involved high-dimensional tensors, can be mapped efficiently onto small 2D GEMMs and BRGEMMs. In this work, we adopt the same strategy to implement CNNs via the BRGEMM TPP. Unlike the previous work which presents only the address-based BRGEMM formulation, here we leverage the CNN kernels with stride-based BRGEMM for 1$\times$1 convolutions and offset-based BRGEMM for 3$\times$3 convolutions to get even more performant implementations (see Section~\ref{subsec:tpppresentation} for a brief description of the BRGEMM variants).

\subsubsection{Sparse Embedding Kernel}
\label{subsubsec:emb}
\begin{algorithm}[t]
\begin{algorithmic}[1]
\LState \textbf{Inputs}: $\alpha^T = [0,\ldots,a_{p_{1}},\ldots,a_{p_{k}},\ldots,0]$ with entries $a_p = 1$ for $p \in \{p_1, \ldots, p_k\}$ and $0$ elsewhere, $W^{M\times E}$ 
\LState \textbf{Output}: $o^T=a^T\times W$
\For{$j=0 \dots E\ \textbf{with step}\ vlen\cdot U$}
\LineComment{Initializing accumulator registers to 0}
\For{$u=0 \dots U-1$}
\State $vec\_out_u \leftarrow 0$
\EndFor
\LineComment{Iterating over non-zero entries/indices in $\alpha^T$}
\For{$i\ \textbf{in}\ 1, 2,\ldots,\ k $}
\State $idx = p_{i}$
\State $next\_idx = p_{i+1}$
\LineComment{Unroll innermost kernel $U$ times: load indexed vector, prefetch next indexed vector, accumulate loaded vector to accumulator register}
\For{$u=0 \dots U-1$}
\State $vec\_W \leftarrow$ load\_vector($W[idx][j+u\cdot vlen:j+(u+1)\cdot vlen]$)
\State prefetch($W[next\_idx][j+u\cdot vlen:j+(u+1)\cdot vlen]$)
\State $vec\_out_u \pluseq vec\_W$
\EndFor
\EndFor
\LineComment{Store accumulator registers to $o^T$}
\For{$u=0 \dots U-1$}
\State $o^T[j+u\cdot vlen:j+(u+1)\cdot vlen] \leftarrow vec\_out_u $
\EndFor
\EndFor
\end{algorithmic}
\caption{Sparse Gather-Reduce operation}
\label{alg:embedding}
\end{algorithm}

\begin{figure}[t!]
\centering
\includegraphics[width=0.5\columnwidth]{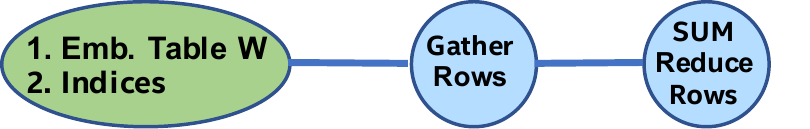}
\caption{Sparse Embedding Lookups via TPPs}
\label{fig:emb_bag}
\end{figure}

The sparse embedding kernel is comprised of multi-hot encoded lookups into an embedding table $W^{M\times E}$ with $M$ being the number of rows and $E$ the length of each row, whereas the multi-hot weight-vector is denoted as $\alpha^T = [0,\ldots,a_{p_{1}},\ldots,a_{p_{k}},\ldots,0]$ with entries $a_p = 1$ for $p \in \{p_1, \ldots, p_k\}$ and $0$ elsewhere ($p$ being the index for the corresponding lookup items). Mathematically, the embedding lookup output vector $o^T$ can be obtained via $o^T=a^T\times W$. This operation (assuming row-major storage for $W$) is equivalent to gathering the rows of $W$ based on the non-zero indices $a_p$, and then adding them up to get the output vector $o^T$. Figure~\ref{fig:emb_bag} illustrates the TPP tree that is used to express the Sparse Embedding lookup kernel.

\newtext{We note that the TPP backend optimizes this sequence of TPPs, and performs register fusion across the gather and the reduce TPP components. More specifically, given a non-zero index $a_p$, the corresponding row of $W$ is loaded in vector registers, and is added to a set of running accumulators/vector registers that hold the output $o^T$. Algorithm~\ref{alg:embedding} illustrates the optimized JITed implementation in our TPP backend. The $E$ dimension is vectorized in an SIMD-fashion with vector length $vlen$. Note that in line 13 we expose multiple independent accumulation chains in order to hide the latency of the vector-add SIMD instructions. Since we JIT this sub-procedure, we know the exact value of $E$ at runtime. As such, we can pick appropriate unrolling factor $U$ as well as the remainder handling can be performed optimally via masking in case $E$ is not perfectly divisible by the vector length $vlen$. Last but not least, the JITed aggregation procedure employs prefetching of the subsequent indexed vectors in $W$ (line 12) in order to hide the latency of these irregular accesses.}

\subsubsection{Multi-Layer Perceptron (MLP) kernel}
\label{subsec:mlp}
\begin{algorithm}[t]
\begin{algorithmic}[1]
\LState \textbf{Inputs}:\ $A^{M_b\times K_b\times b_k\times b_m}$,\ $B^{N_b\times K_b\times b_n\times b_k}$
\LState \textbf{Output}: $C^{N_b\times M_b\times b_n\times b_m}$
\State Based on $thread\_id$ calculate $M_b\_start$, $M_b\_end$, $N_b\_start$ and $N_b\_end$ to assign output work items
\For{$ib_n=N_b\_start \dots N_b\_end$}
\For{$ib_m=M_b\_start \dots M_b\_end$}
\State $Out =  \&C[ib_n][ib_m][0][0]$
\LineComment{Stride-based BRGEMM, stride\_A=$b_k\cdot b_m$, stride\_B=$ b_n\cdot b_k$}
\State $\mathbf{BRGEMM}(\&A[ib_m][0][0][0], \&B[ib_n][0][0][0], Out, K_b)$ 
\State $C[ib_n][ib_m][0][0] \leftarrow \mathbf{UNARY}(C[ib_n][ib_m][0][0])$
\EndFor
\EndFor
\end{algorithmic}
\caption{Fully-Connected Layer with Unary Activation Function}
\label{alg:fc}
\end{algorithm}

Multilayer perceptrons (MLP) form a class of feed-forward artificial neural networks. An MLP consists of (at least three) \emph{fully connected} layers of neurons. Each neuron in the topology may be using a non-linear activation function. In this section we present the implementation of the \emph{Fully Connected} layers since they constitute the cornerstone of MLP. Even though we illustrate the forward pass of Fully Connected layers, we also implement via TPPs the kernels of the back-propagation training in an analogous fashion. Algorithm~\ref{alg:fc} shows the fully connected layer implementation which is mapped to TPPs. First we note that the input tensors are conceptually 2D matrices $A^{M\times K}$ and $B^{K\times N}$ that need to be multiplied. We follow the approach of previous work~\citep{georganas2020harnessing} and we block the dimensions $M$, $K$, and $N$ by factors $b_m$, $b_k$, and $b_n$ respectively. Such a blocked layout is exposing better locality and avoids large, strided sub-tensor accesses which are known to cause TLB misses and cache conflict misses in case the leading dimensions are large powers of 2~\citep{georganas2020harnessing}. We leverage the BRGEMM TPP in order to perform the tensor contraction with $A$ and $B$ across their dimensions $K_b$ and $b_k$ (which constitute the $K$/inner-product dimension of the original 2D matrices). We employ the stride-based BRGEMM because the sub-blocks ``$A_i$" and ``$B_i$" that have to be multiplied and reduced are apart by constant strides $stride\_A=b_k\cdot b_m$ and $stride\_B=b_n\cdot b_k$ respectively. Finally, we apply (optionally) a unary TPP corresponding to the requested activation function (e.g.\ RELU) onto the just-computed output block of $C$.

\subsection{TPP-based Workloads}
\subsubsection{1D Dilated Convolutions \& Computational Biology}
\label{subsec:dilated}
\begin{algorithm}[t]
  \caption{1D Dilated convolution forward pass using TPPs}\label{Algo:Forward Pass TPP}
  \begin{flushleft}
        \textbf{Inputs:} $I^{C \times W}$, $W^{K \times C \times S}$, $d \in \mathbb{R}$\\
        \textbf{Output:} $O^{K \times Q}$
  \end{flushleft}
  \begin{algorithmic}[1]
      \State \texttt{$W^T \leftarrow \mathbf{TRANSPOSE}(W)$}
      \For{$pos=0 \dots Q-1\ \textbf{with\ step\ }\mathbf{b_q}$}
       \LineComment{Address-based BRGEMM, prepare arguments $A_{ptrs},\ B_{ptrs}$}
      \For{$s=0 \dots S-1\ \textbf{with\ step\ }\mathbf{1}$}
            \State \texttt{$A_{ptrs}[s] = \&W^T[s, 0, 0]$}
            \State \texttt{$B_{ptrs}[s] = \&I[0, (pos + s\cdot d)]$}
        \EndFor
        \State \texttt{${\mathbf{BRGEMM}}(A_{ptrs},B_{ptrs},\&O[0,pos],S)$}
      \EndFor
  \end{algorithmic}
\end{algorithm}
In this subsection, we show the implementation of a special type of convolution via TPPs in their entirety, namely one-dimensional (1D) dilated convolution layer of a 1D CNN named ATACworks~\citep{Lal829481}. ATACworks is used for de-noising and peak calling from ATAC-Seq genomic sequencing data~\citep{Lal829481}. The 1D dilated convolution layer in ATACworks takes more than 90\% of the training time, and it has input tensor width $W$, output tensor width $Q$, $C$ input channels, $K$ output channels, filter size of $S$, and dilation $d$. We employ the transpose TPPs, copy TPPs, and BRGEMM TPPs to optimize the forward pass and the backward pass of the PyTorch-based 1D convolution layer. Algorithm~\ref{Algo:Forward Pass TPP} shows an example of the forward pass procedure with an input tensor $I$, a weight tensor $W$, and an output tensor $O$.

\subsubsection{Deep Learning Recommendation Model}
\label{subsec:dlrm}
Facebook recently proposed a deep learning recommendation model (DLRM)~\citep{naumov2019deep}. Its purpose is to assist the systematic hardware-software co-design for deep learning systems. DLRM is comprised of the following major components: (a) a sparse embedding (see subsection~\ref{subsubsec:emb}) involving tables (databases) of varying sizes, (b) a small dense Multi-Layer Perceptron (see subsection~\ref{subsec:mlp}), and (c) a larger and deeper MLP which is fed by the interaction among (a) and (b). All three parts can be configured (number of features, mini-batch sizes and table sizes) to stress  different aspects of the system. We also note that in the case of training with BF16 datatype we leverage the BF16 split-SGD optimizer (see subsection~\ref{subsec:splitsgd_kernel}). For more details on the workload and CPU-oriented optimizations we refer to prior work~\citep{kalamkar2020optimizing}.

\subsubsection{Natural Language Processing - BERT}
\label{subsec:bert}
The BERT model is a bidirectional transformer pre-trained via a combination of masked language modeling objective, and next-sentence prediction~\citep{devlin2018bert}. The heart of the BERT model is
comprised by sequence of BERT layers which are built using smaller building blocks. For ease of use and implementation, we followed modular building blocks from Hugging Face transformers library \citep{wolf-etal-2020-transformers} and implemented four fused 
layers using TPP building blocks, namely \emph{Bert-Embeddings}, \emph{Bert-SelfAttention}, \emph{Bert-Output}/\emph{Bert-SelfOutput} and \emph{Bert-Intermediate} layers.

The \emph{SelfAttention} layer in turn can be formulated as a bunch of Matrix / batch Matrix-Multiplications mixed with element-wise scale, add, dropout and softmax operators. We formulate these Matrix-Multiplications as tensor contractions on blocked-tensors via the stride-based BRGEMM TPP (similarly to Algorithm~\ref{alg:fc}). We opt to use blocked tensor layouts for the same reasons briefly described in Section~\ref{subsec:mlp}. Furthermore, by working on one small sub-tensor at a time we naturally follow a ``dataflow" computation, which has been shown to maximize the out-of-cache-reuse of tensors among cascading operators~\citep{banerjee2019optimizing,zhang2018deepcpu}. The softmax operator is also formulated entirely by TPPs as described in Section~\ref{subsec:smax_kernel}. We note that the sequence of Matrix-Multiplications in the attention layer requires sub-tensors to be transposed (and VNNI transformed in case of BF16 implementation), and for this task we leverage the transpose/transform TPPs. \emph{Bert-Output} and \emph{Bert-SelfOutput} layers perform GEMM over blocked layout, and fuse bias addition, dropout, residual addition and layernorm  using TPPs. The \emph{Bert-Embeddings} layer also performs layernorm and dropout after embedding lookups that are also implemented using TPPs. Finally, \emph{Bert-Intermediate} layer performs blocked GEMM followed by bias addition and GELU activation function which we implement using the GELU TPP.

\subsubsection{Emerging AI - Graph Neural Networks}
\label{subsec:gnn}
\begin{figure}[t!]
\centering
\includegraphics[width=0.7\columnwidth]{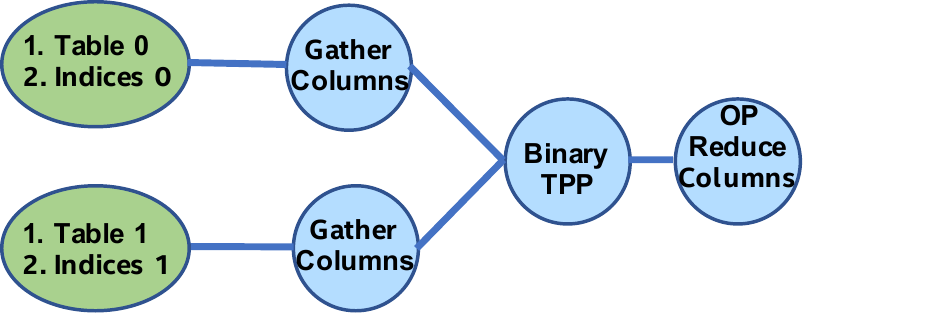}
\caption{Binary-Reduce aggregation kernel via TPPs}
\label{fig:aggregation}
\end{figure}

Graph Neural Networks (GNN)~\citep{hamilton2017inductive} form an emerging class of Neural Networks for learning the structure of large, population-scale graphs. Depending on the specific algorithm and task that a GNN is designed for (e.g.\ node classification, link prediction), feature-vector aggregation precedes or succeeds a shallow neural network. Such a shallow neural network typically materializes one or more linear transformations, followed by a classification or regression mechanism~\citep{avancha2020deep}, and the relevant TPP-based implementation is essentially the one we present in Algorithm~\ref{alg:fc}.

We focus here on the TPP-based implementation of the feature-vector aggregation. This aggregation motif can be seen as a sequence of linear algebraic expressions involving node/edge features, along with the relevant operators. Prior work~\citep{avancha2020deep} has focused on the following two algebraic sequences: Copy-Reduce and Binary-Reduce. We elaborate here on the latter sequence Binary-Reduce (as the first is even simpler). The feature-vectors (either pertaining to vertices or edges) are represented via dense 2D matrices/tables. At the same time, the adjacency information in the graphs can be eventually found via arrays of indices. Therefore, by providing a set of indices and the appropriate Tables of feature-vectors (assuming column-major storage), one can extract selectively the desired feature-vectors via Gather-columns operations. Then, the extracted feature-vectors are fed into a binary operator, and the outcome of the binary operations are finally reduced (the reduce operation could be sum/max/min etc).

Figure~\ref{fig:aggregation} illustrates a TPP tree that is used to express the Binary-Reduce aggregation kernel. The TPP back-end optimizes this sequence of TPPs and performs horizontal register fusion across them. \newtext{More precisely, two feature-vectors namely $v_0$ and $v_1$ are extracted at a time from  Table 0 and Table 1 respectively by using the relevant indices arrays, and they are combined via the proper binary op to get an intermediate vector $v_i$. Subsequently, $v_i$ is reduced with a running reduce-vector $v_o$ that holds the output of this composite operator. Once the running reduction has been completed (i.e.\ all indexed columns from Table 0 and Table 1 have been accessed, processed and reduced), the output vector $v_o$ is stored in the corresponding output subtensor.}

\section{Experimental Results of DL kernels \& Workloads}
\label{sec:results}
We use a variety of platforms that span different ISAs, different vendors and micro-architectures. More specifically, our tested platforms include: i) a 22-core Intel Xeon E5-2699 v4 (BDX) supporting up to AVX2 ISA, ii) a 28-core Intel Xeon 8280 (CLX) supporting up to AVX512 ISA, iii) a recently announced 40-core Intel Xeon 8380 (ICX) supporting also up to AVX512 ISA,  iv) a 28-core Intel Xeon 8380H (CPX) supporting up to AVX512 ISA, which also offers BF16 FMA acceleration, v) a 64-core AMD EPYC 7742 (ROME) with AVX2 ISA, vi) an AWS Graviton2 instance with 64 cores at fixed 2.5 GHz and AArch64 ISA, vii) a 48-core Fujitsu A64FX at fixed 1.8 GHz with ARMv8 SVE ISA, and viii) a 4-core client Intel i7-6700 CPU (i7) supporting up to AVX2 ISA. All Intel and AMD chips are operating in Turbo mode. For the cluster experiments we used a 32 node CLX installation with a dual-rail Intel Omnipath 100 pruned 2:1 fat-tree topology.

\subsection{Performance of standalone DL kernels}
\label{subsubsection:kernel_perf}
\begin{figure}[t!]
\centering
\includegraphics[width=\columnwidth]{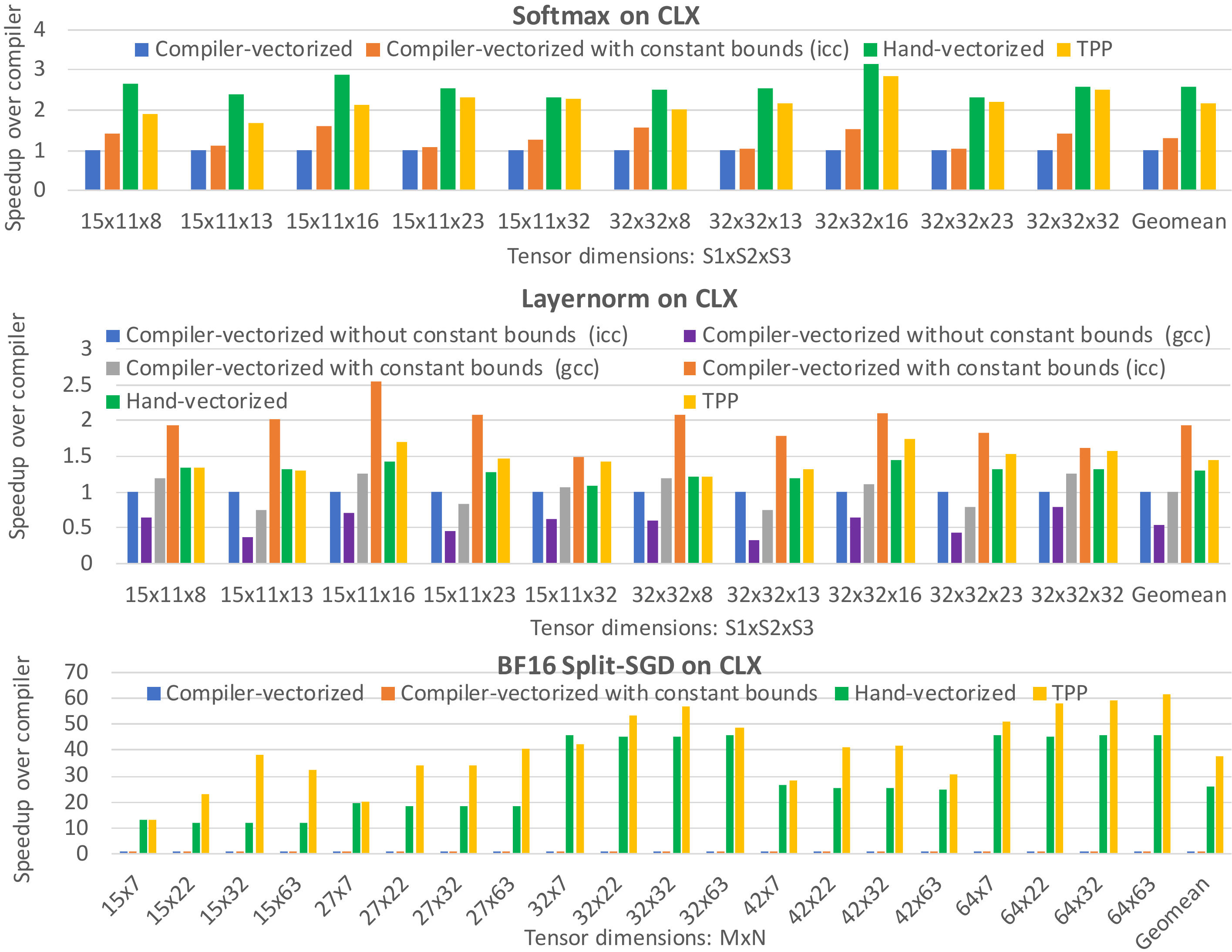}
\caption{TPP kernels on CLX}
\label{fig:standalone_clx}
\end{figure}
\begin{figure}[t!]
\centering
\includegraphics[width=\columnwidth]{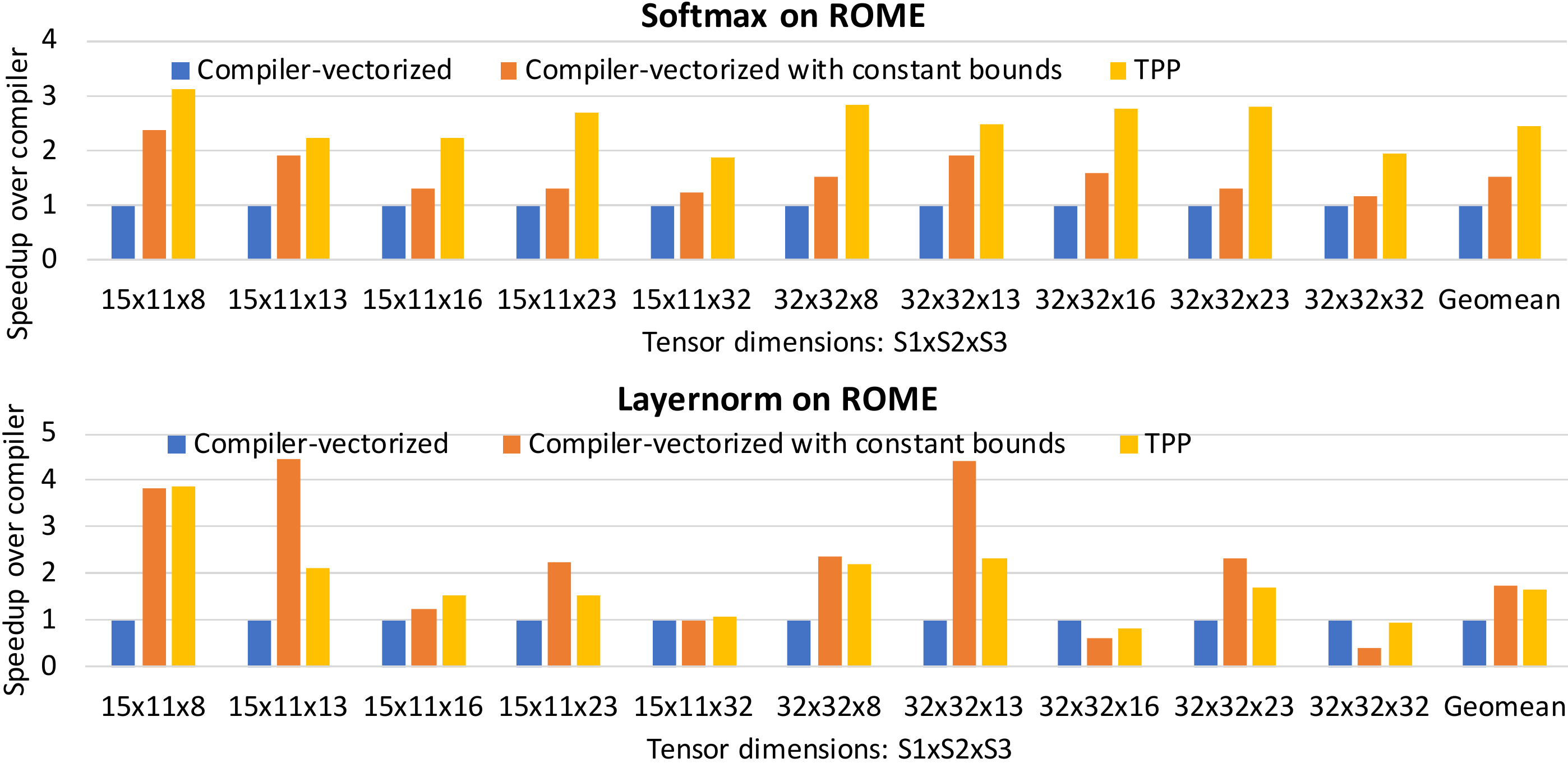}
\caption{TPP kernels on ROME}
\label{fig:standalone_rome}
\end{figure}

We start the performance evaluation with standalone TPP kernels presented in subsection~\ref{subsec:standalone_kernels}. First, we want to highlight the productivity/efficiency provided by TPPs:  the high-level code expressed via TPPs/trees of TPPs can match or outperform code by compilers, and hand-vectorized (thus non-portable code) written by performance experts. Second, we want to show the portability aspect of TPPs, since exactly the same high-level code yields high-performance across different ISAs and micro-architectures.

Figure~\ref{fig:standalone_clx}-Top shows the performance of the Softmax operator of blocked 3D tensors with size $S1\times S2\times S3$, on the CLX platform (i.e.\ targeting AVX512 ISA). Here we perform $S2$ softmax operations over blocked $S1\times S3$ dimensions. The sizes are chosen such that some of the dimensions do not match perfectly with the vector length. The baseline is the icc generated code with \texttt{-O3} optimization level and \texttt{high-zmm usage} flags. The second variant is also icc-generated code, but we propagate the tensor sizes/loop bounds via compile-time constants in order to assist the auto-vectorization/optimize remainder handling via masking. The third code variant is the AVX512 hand-vectorized by an expert, where the $exp$ function uses fast Taylor approximation. Last, we evaluated the TPP-based softmax implementation. As we can see, by propagating the tensor sizes we achieve (geo-mean) speedup of 1.3$\times$ over the baseline. The hand-vectorized code is faster by 2.6$\times$ whereas the TPP-based variant shows similar speedups by being 2.2$\times$ faster. The main shortcoming of the hand-vectorized code is that it is platform-dependent and as such non-portable. More specifically, we didn't have to our avail AVX2 hand-optimized code in order to experiment with it on ROME. On the contrary,
Figure~\ref{fig:standalone_rome}-Top shows the softmax performance on AVX2 enabled platform for the compiler-generated code and the TPP based code. The TPP-based softmax exhibits geo-mean speedup of 2.45$\times$ over the baseline on ROME.

Figure~\ref{fig:standalone_clx}-Middle shows the performance of the layernorm operator on the CLX platform. Since the layernorm code is more straightforward (i.e.\ no expensive \emph{exp} function is involved), we see that icc with compile-constant bounds outperforms by 1.9$\times$ the baseline. We inspected the compiler-generated code and identified that the reduction-loops were recognized and were heavily optimized with multiple accumulation chains etc. Similarly, the hand-vectorized code and the TPP based code outperform the baseline by 1.3$\times$ and 1.5$\times$. We also experimented with gcc and the \texttt{fast-math} flag, and it just matched baseline performance. We want to emphasize that propagating the tensor sizes as compile-time constants throughout the operators is not  practical for real use-cases within DL frameworks. 
Figure~\ref{fig:standalone_rome}-Bottom shows similar performance speedups on ROME, where the TPP-based code is 1.6$\times$ faster than the auto-vectorized baseline.

Figure~\ref{fig:standalone_clx}-Bottom shows the performance of the BF16 split-SGD operator on CLX. This use-case represents a novel, mixed-precision operator where the compiler (even with compile-time constant tensor sizes) struggles to yield good performance; the TPP-based code has geo-mean speedup of $38\times$ over the compiler generated code.

\begin{figure*}[t!]
\centering
\includegraphics[width=\textwidth]{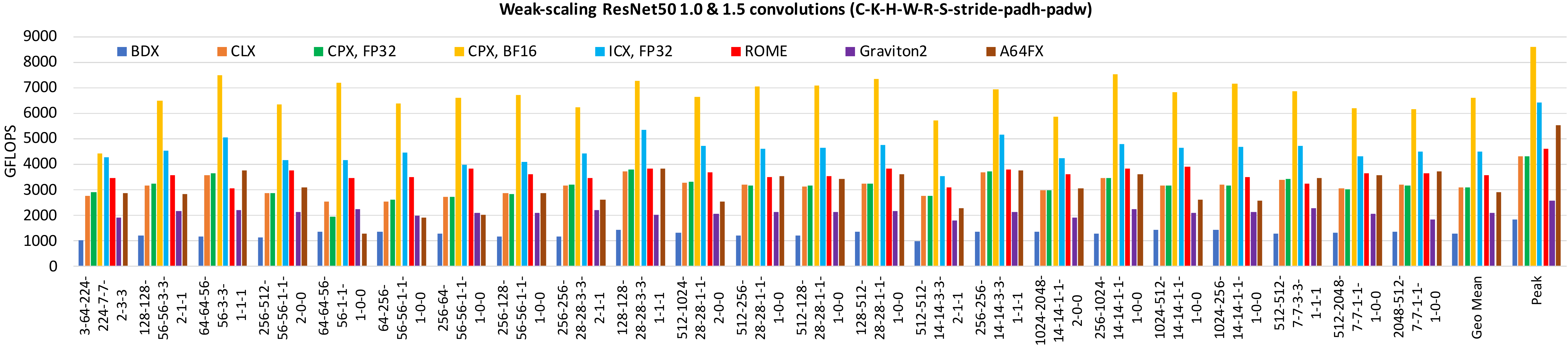}
\caption{Convolutions via BRGEMM TPP}
\label{fig:convolutions}
\end{figure*}
Figure~\ref{fig:convolutions} illustrates the TPP-based implementation of various ResNet50~\citep{he2016deep} Convolution layers across all available platforms. The minibatch size used on each platform equals to the number of the corresponding cores. It is noteworthy that the TPP-user code is identical for all targets, hence truly portable; it is merely that the TPP backend optimizes the code generation (BRGEMM in this case) in a platform/ISA-aware fashion. The geomean efficiencies of these convolutions are: $69\%$ for BDX, $72\%$ for CLX, $72\%$ for CPX, $77\%$ for CPX with BF16 datatype, $70\%$ for ICX, $78\%$ for ROME, $81\%$ for Graviton2 and $52\%$ for A64FX. 
Previous work~\citep{georganas2020harnessing} also showed on an x86 TPP-predecessor that BRGEMM-based convolutions matched or outperformed Intel's oneDNN library~\citep{onednn}. Fujitsu recently contributed an A64FX back-end to oneDNN~\citep{onednn-fujitsu} and our TPP implementation outperforms this by 22\% on the geomean. We observe that our TPP convolutions not only run on all of these different platforms without a single line of code change, but they run at very similar hardware utilization.

\subsection{Performance of end-to-end DL workloads}
\subsubsection{1D Dilated Convolutions and their application to Computational Biology}
\begin{figure}[t!]
\centering
\includegraphics[width=\columnwidth]{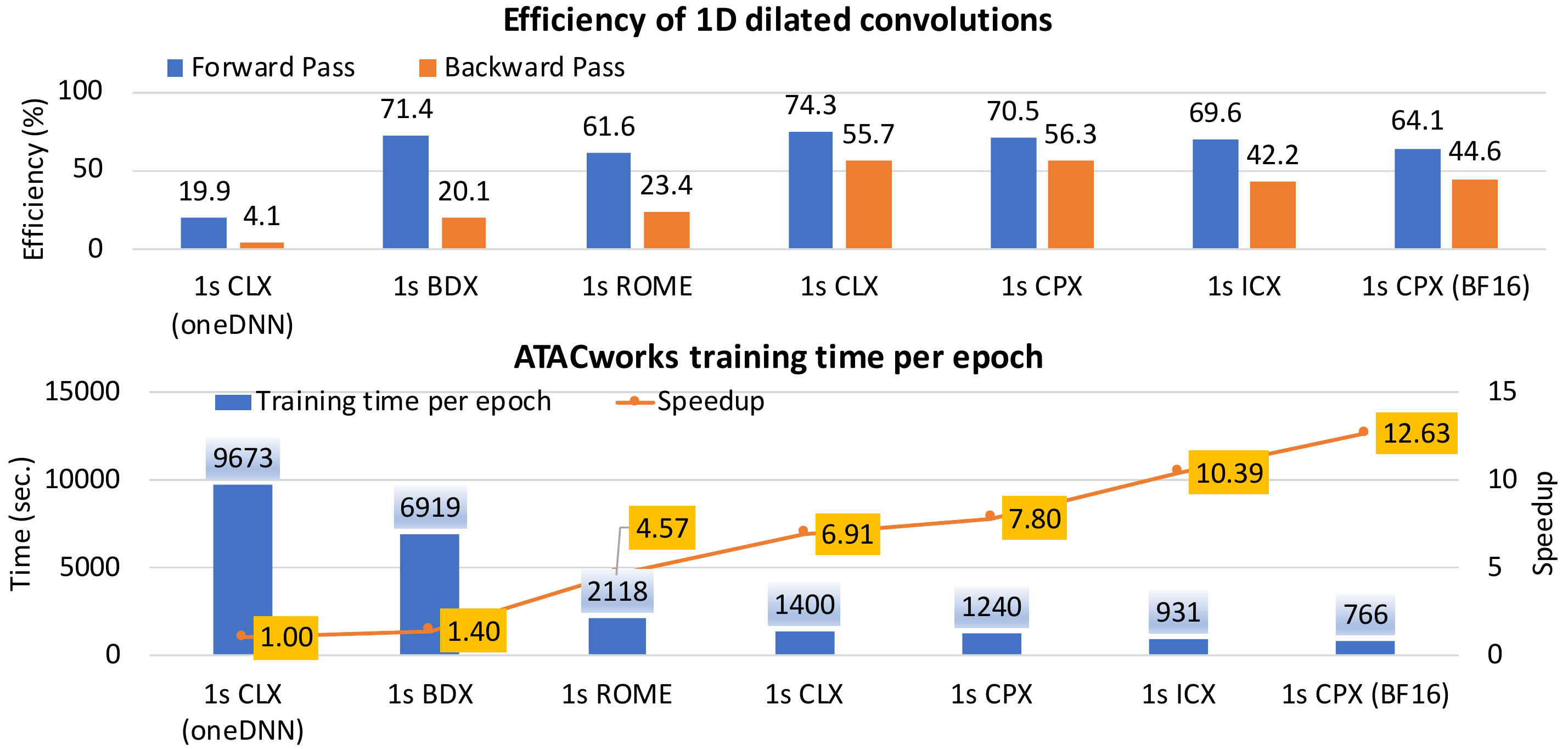}
\caption{1D Dilated Convolutions}
\label{fig:dilated_convs}
\end{figure}
Here we evaluate the oneDNN~\citep{onednn} and TPP-based 1D dilated convolution layer of ATACworks \citep{Lal829481} which takes takes more than 90\% of the training time, and it has input tensor width ($W$) of 60400, output tensor width ($Q$) of 60000, 15 input channels ($C$), 15 filters ($K$), filter size ($S$) of 51, and dilation ($d$) of 8. Figure~\ref{fig:dilated_convs}-Top shows the computational efficiency results of the 1D convolution layer. oneDNN is not reaching peak performance for these specialized convolutions, exhibiting 19.9\% efficiency for the forward pass and only 4.1\% for the backward pass on CLX. Our TPP-based implementation shows 74.3\% and 55.7\% efficiency for the corresponding training passes. We also highlight the performance portability of our TPP-based approach across all tested platforms. Finally, we show training time per epoch results for ATACworks in Figure~\ref{fig:dilated_convs}-Bottom. The TPP-based kernels provide training time speedup of 6.91$\times$ on CLX when comparing to the oneDNN based implementation. We also show that by leveraging the BF16 FMA acceleration of the CPX platform we can further obtain 1.62$\times$ speedup compared to the FP32 implementation on the same platform. In total BF16 yields 12.6$\times$ speedup over the oneDNN baseline.

\subsubsection{Deep Learning Recommendation - DLRM}
\label{subsubsection:dlrm_perf}
\begin{figure}[t!]
\centering
\includegraphics[width=\columnwidth]{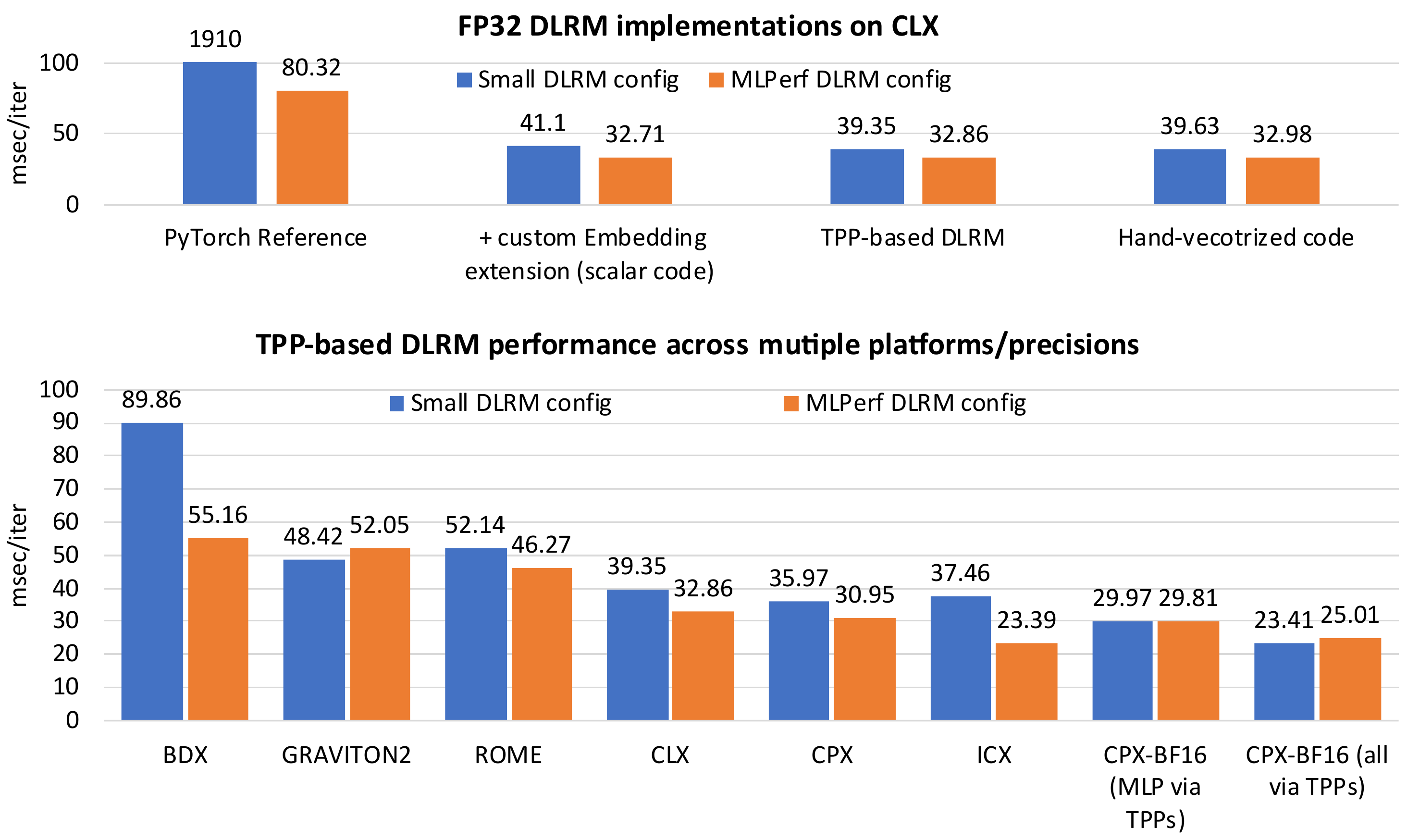}
\vspace{-2.0em}
\caption{DLRM performance on a small config (blue bars) and on the MLPerf config (orange bars)}
\label{fig:dlrm_perf}
\end{figure}
\begin{figure}[t!]
\centering
\includegraphics[width=\columnwidth]{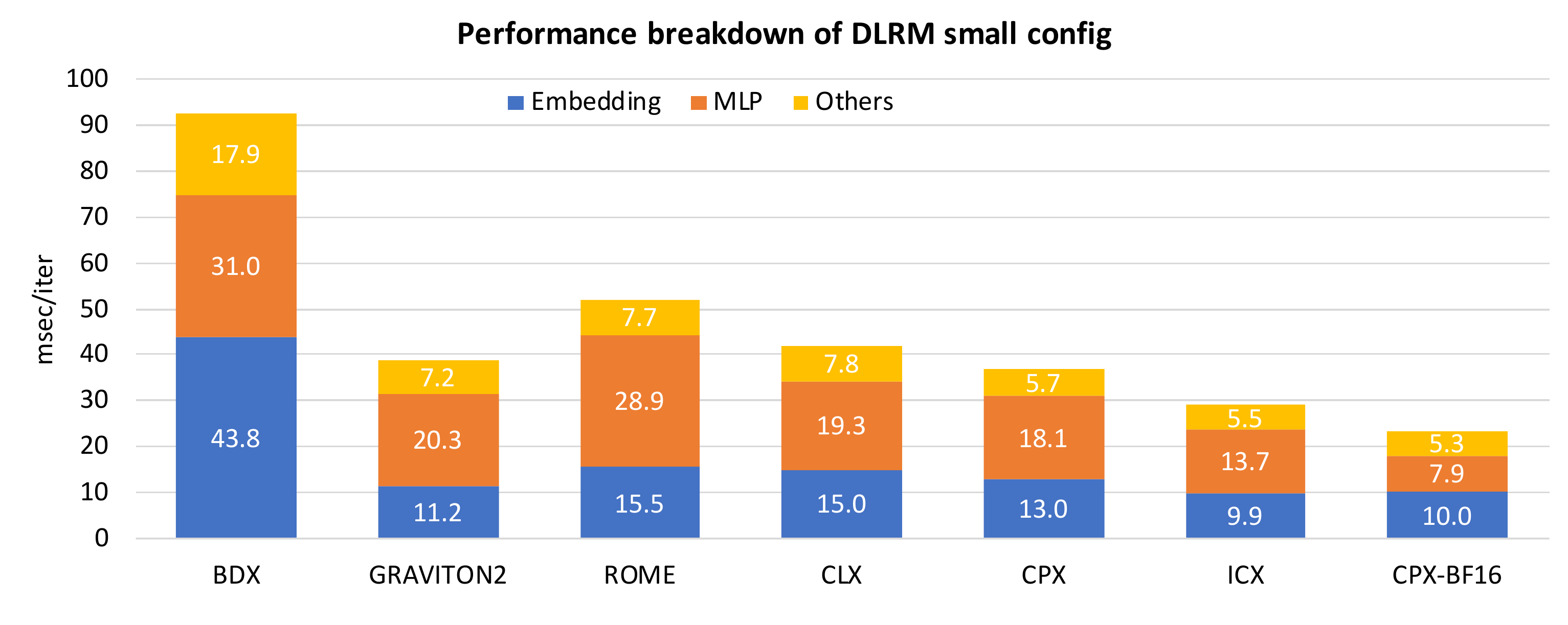}
\caption{DLRM performance breakdown of small config on multiple platforms}
\label{fig:dlrm_breakdown}
\end{figure}
Figure~\ref{fig:dlrm_perf}-Top shows the FP32 DLRM performance on CLX using two different configurations, namely small DLRM (blue bars) and MLPerf DLRM (orange basrs). We refer to previous work for the detailed specification of these configurations~\citep{kalamkar2020optimizing}. We evaluated 4 different implementations of DLRM: i) the PyTorch reference implementation, ii) PyTorch reference + custom Embedding extension auto-vectorized by the compiler, iii) DLRM expressed entirely via TPPs, and iv) hand-vectorized Embedding extension + BRGEMM-TPP based MLPs~\citep{kalamkar2020optimizing}. We conclude that the TPP-based implementation matches the performance of the State-Of-The-Art implementation which is hand-vectorized specifically for AVX512 targets; both of these optimized versions substantially outperform the PyTorch CPU reference implementation by up to 48$\times$. Compared to the version with the custom, auto-vectorized variant the TPP-version is up to 4.4\% faster.  

Figure~\ref{fig:dlrm_perf}-Bottom shows the DLRM performance of our TPP-based implementation across multiple platforms and compute precisions. We want to highlight two aspects: First, we are able to run the same TPP-code without any change across all platforms, something that is not doable with the hand-vectorized SOTA variant (iv) (since it is not able to run on the AVX2-only BDX and ROME platforms, \newtext{or on the Graviton2 platform with AArch64 ISA}). Second, the TPP-based BF16 shows speedup up to 28\% over the variant with auto-vectorized Embedding extension. The culprit here is the mixed precision operations like split-SGD where the compiler struggles to yield efficient code as shown in  Section~\ref{subsubsection:kernel_perf}.

Figure~\ref{fig:dlrm_breakdown} illustrates the performance breakdown of the small config on multiple platforms. The blue portions of the bars correspond to the time spent on the Embedding component, the orange parts reflect the MLP portion, and finally the yellow portions correspond to the remaining components of the DLRM workload. We observe that depending on the platform, the time spent on Embedding varies from 29-37\% of the total execution time, the time spent on MLP is in the range of 33-56\% of the total time, and the rest components account for 15-23\% of the time. We can also observe the correlation of the MLP performance with the compute capabilities of each platform. For example, on CPX which has native BF16 FMA support, the BF16 MLPs are sped up by $\sim$2$\times$ compared to the FP32 MLPs on the same platform. In regard to the time spent on the Embedding kernel which tends to be bandwidth bound, we observe  correlation with the corresponding bandwidth capabilities of the machines. 

\subsubsection{Natural Language Processing - BERT Large}
\begin{figure}[t!]
\centering
\includegraphics[width=\columnwidth]{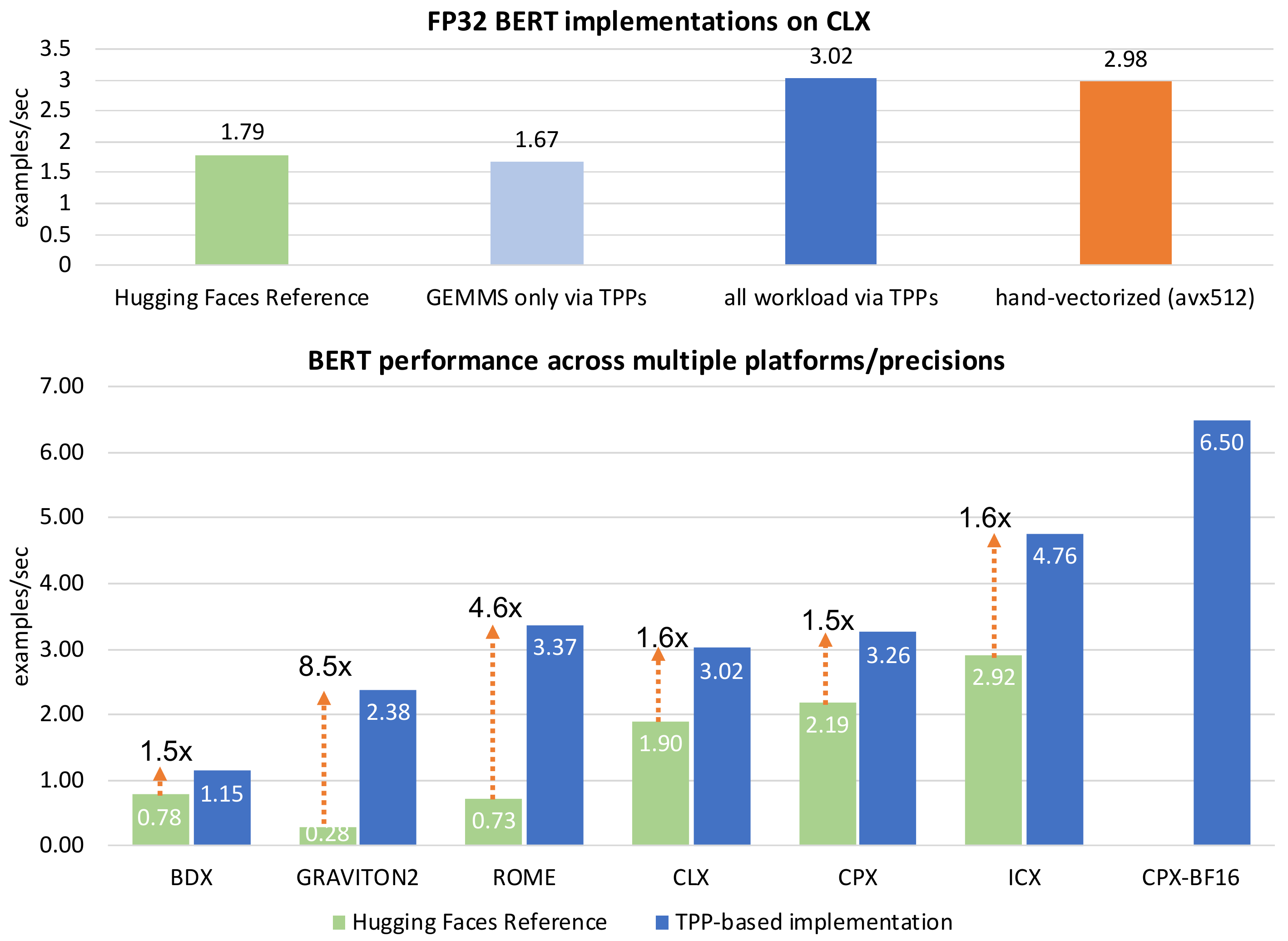}
\caption{BERT Large performance}
\label{fig:bert_perf}
\end{figure}

Figure~\ref{fig:bert_perf}-Top shows end-to-end performance (in examples/second) on CLX for the BERT large SQuAD fine-tuning task in FP32, using a max sequence length of 384 and minibatch of 24. We observe that the TPP-based implementation (blue bar) matches the performance of the AVX512-hand-vectorized code/orange bar. At the same time, our implementation is  1.69$\times$ faster than the Reference Hugging Faces CPU reference code~\citep{huggingfaces} (green bar).

\newtext{Figure~\ref{fig:bert_perf}-Bottom shows the performance of the reference Hugging Faces code (green bars) versus the TPP-based code (blue bars) across multiple platforms (x86 and AArch64/Graviton2) and compute precisions (FP32 for all platforms, and BF16 for the CPX platform). The TPP-based BERT shows speedups ranging from 1.5$\times$ to 8.5$\times$ over the Hugging Faces code. This result highlights the performance portability through the TPP abstractions. In regard to various compute precisions, we note that with minimal changes inside the fused operators to handle the VNNI tensor layout (required for BF16 GEMM/BRGEMM), and a couple of lines changes in the application code to enable BF16 training, we were able to realize 2$\times$ speed up using BF16 training on CPX (compared to FP32 training on CPX) with 28 cores, surpassing 40-core FP32-ICX performance by 37\%.}

\begin{figure}[t!]
\centering
\includegraphics[width=\columnwidth]{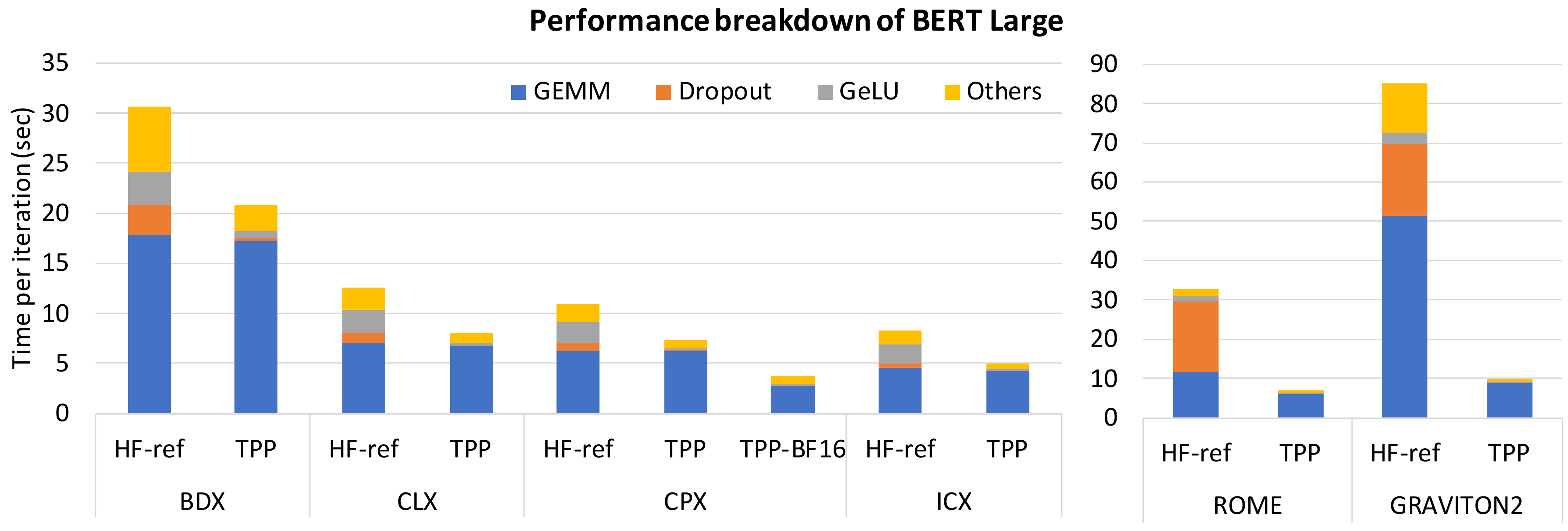}
\caption{BERT Large performance breakdown on multiple platforms.}
\label{fig:bert_breakdown}
\end{figure}

\newtext{In order shed light on where the benefits are coming from, we present in Figure~\ref{fig:bert_breakdown} the performance breakdown of the Hugging Faces reference code and the TPP-based implementation. In particular we focus on 4 components:
\begin{enumerate}
\item \emph{GEMM} which corresponds to the tensor contractions implemented via either the BRGEMM-TPP in the TPP implementation, or it leverages optimized GEMM routines within BLAS libraries in the Hugging Faces implementation (MKL for x86 platforms and OpenBLAS for AArch64/Graviton2).
\item \emph{Dropout} corresponding to the dropout layer in BERT, where the TPP-based implementation employs fast random number generation via xorshift algorithm. 
\item \emph{GeLU} corresponding to the Gaussian Error Linear Unit activation function in BERT, where the TPP-based implementation leverages fast approximations as discussed in section~\ref{subsec:approx}.
\item \emph{Others} capturing the remaining operators: Transpose, Layer-norm, softmax, bias addition, vnni-reformatting (in case of BF16 training), copy, add, scale, zero-kernel, reduce, optimizer. Note that all these operators map to either unary/binary/ternary TPPs (see section~\ref{sec:specification}) or the can be expressed via Matrix Equation TPPs (see section~\ref{sec:workloads}). 
\end{enumerate}}

\begin{figure}[t!]
\centering
\includegraphics[width=\columnwidth]{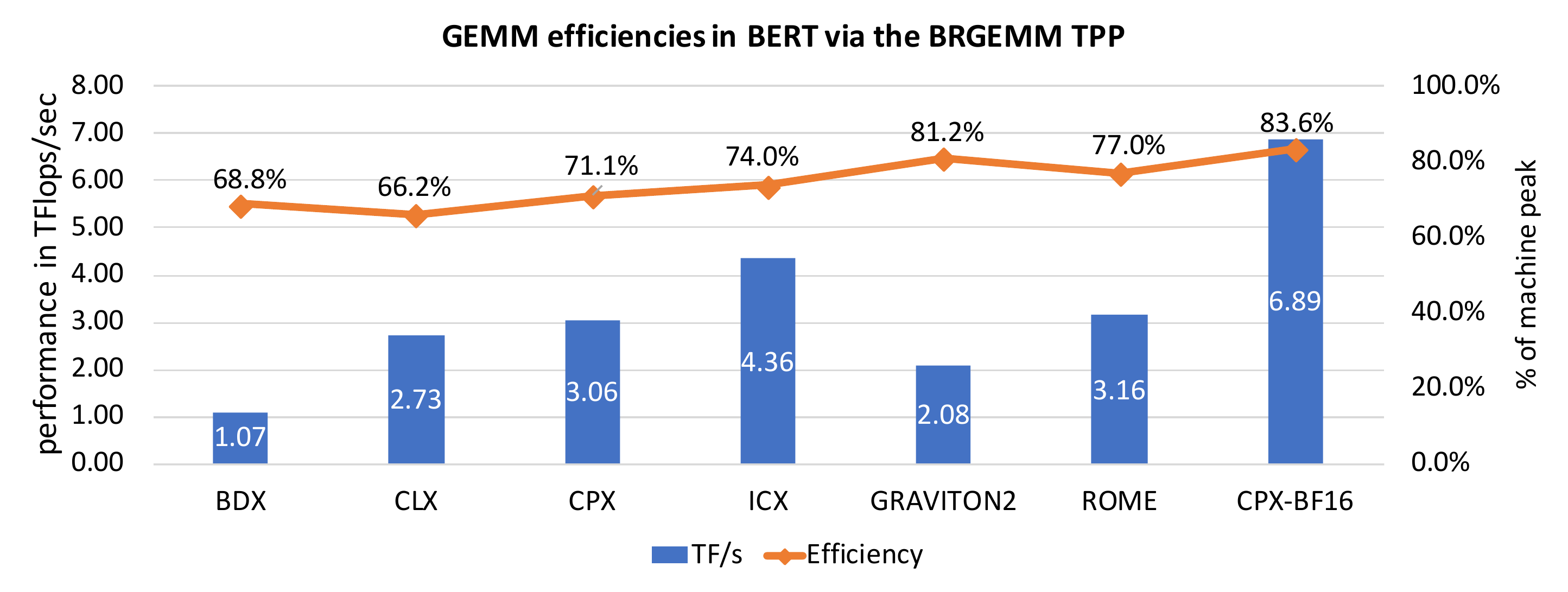}
\caption{BERT GEMM/tensor contraction efficiencies via the BRGEMM-TPP on multiple platforms.}
\label{fig:bert_gemm_eff}
\end{figure}

\newtext{First, we note that for the Intel x86 platforms (left part of the breakdown plot) the tensor contractions show speedups over the highly-optimized MKL GEMM implementation in Hugging Faces in the range of 2-6\%. On the right side of the breakdown plot we observe that the BRGEMM-TPP benefits are even larger on the non-Intel platforms. More specifically, on AMD Rome (AVX2 x86 platform) the tensor contractions are sped up by 1.9$\times$ via the BRGEMM-TPP, and on Graviton2 (Arm AArch64 platform) the tensors contractions are 5.7$\times$ faster via the BRGEMM-TPP compared to the implementation relying on OpenBLAS GEMM calls. To further highlight the performance portability of the tensor contractions via the BRGEMM-TPP across multiple platforms and precisions, Figure~\ref{fig:bert_gemm_eff} shows the achieved GEMM performance (Left axis) on each platform for the entire training process (blue bars), whereas the orange line (Right axis) dictates the \% of machine peak. The conclusion here is that the BRGEMM-TPP delivers high-efficiency for the corresponding tensor contractions in the range of 66-84\% for all tested ISAs and micro-architectures.} 

\newtext{The second conclusion we can draw from the performance breakdown in Figure~\ref{fig:bert_breakdown} is that our fused/dataflow TPP implementation outlined in section~\ref{subsec:bert} makes the dropout and GeLU times shrink substantially, offering speedups in the range of 10-360$\times$. The BERT implementation via the dropout/GeLU TPPs in tandem to the BRGEMM TPPs take advantage of temporal locality, and virtually make the corresponding times disappear from the overall execution time. Last but not least, the remaining components are sped-up in the TPP-based implementation by 2.5-14$\times$ depending on the platform. As a result of these optimizations, the TPP-based BERT implementation spends the majority of the time (75.5-88.8\%) in tensor contractions which are executed at high-efficiency as Figure~\ref{fig:bert_gemm_eff} shows.}

\subsubsection{Emerging AI - Graph Neural Networks}
\begin{figure}[t!]
\centering
\includegraphics[width=\columnwidth]{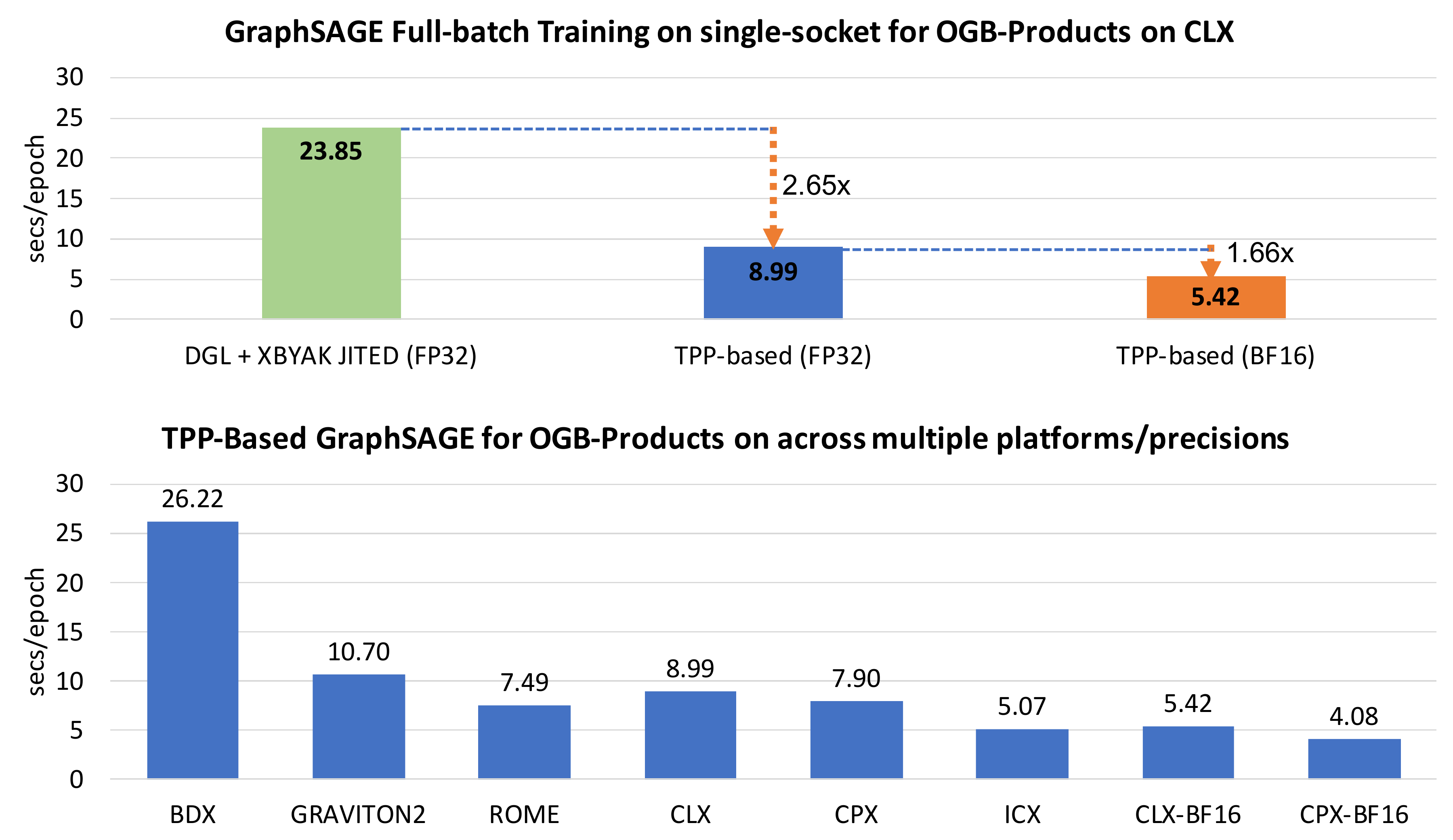}
\caption{GNN performance of GaphSAGE Full-batch training for OGB-Products}
\label{fig:gnn_perf}
\end{figure}
Figure~\ref{fig:gnn_perf}-Top shows end-to-end performance (in seconds/epoch, so lower is better) on CLX for the full-batch training of the GraphSAGE workload on OGB-Products with FP32 and BF16 precision. For the CLX BF16 experiments, since CLX doesn't have native support for BF16 FMAs, we use bit-wise accurate emulated-BF16 BRGEMM TPPs (see section~\ref{subsec:brgemm_mixed}), and we still expect savings due to the bandwidth reduction in the non-GEMM parts, e.g., graph traversal and edge/node aggregation. We observe that the TPP-based implementation outperforms the DGL with Xbyak JIT backend baseline version by 2.65$\times$. The TPP-BF16 version yields another 1.66$\times$ speedup over the TPP-FP32 variant mainly due to reduced bandwidth requirements.

Figure~\ref{fig:gnn_perf}-Bottom shows the performance of the TPP-based code across multiple platforms (x86 and Arm AArch64) and compute precisions (FP32 and BF16). The relative differences in the performance can be justified by the different compute/bandwidth specs of the benchmarked platforms. We highlight that with minimal changes in the MLP portion to handle VNNI layout required for BF16 BRGEMM, and a couple of lines changes in the application code to enable BF16 training, we were able to realize 1.94$\times$ speed up using BF16 training on CPX with 28 cores compared to the FP32 training on the same platform.

\begin{figure}[t!]
\centering
\includegraphics[width=\columnwidth]{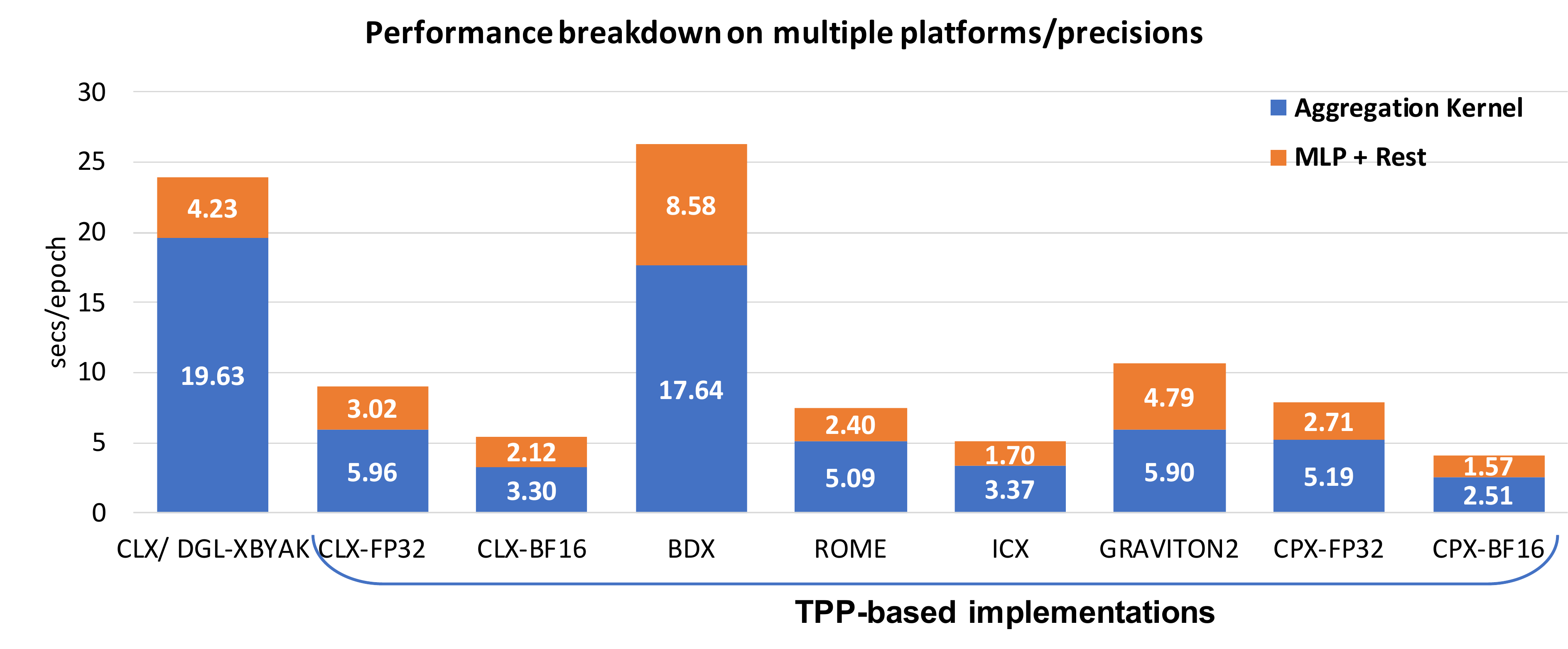}
\caption{GNN performance breakdown of GaphSAGE Full-batch training for OGB-Products}
\label{fig:gnn_perf_breakdown}
\end{figure}

\newtext{In order to further analyze the behavior of the various implementations on multiple platforms, we present on Figure~\ref{fig:gnn_perf_breakdown} the relevant performance breakdown. The very left bar shows the performance breakdown of the FP32 optimized DGL implementation that leverages JITed kernels through Xbyak on the CLX platform. The blue part corresponds to the Aggregation kernel described in subsection~\ref{subsec:gnn} whereas the orange portion represents the time required by the remaining kernels, namely Multilayer-Peceptrons with Activation functions. In the DGL implementation the activation functions are not fused within the MLP's tensor contractions. We observe that in this optimized DGL implementation, 82.3\% is spent on the Aggregation kernel and only 17.7\% is spent on the MLPs. On the second from the left bar (annotated as CLX-FP32) we show the performance of the FP32 TPP-based implementation on the same CLX platform. We conclude that the TPP-based Aggregation kernel exhibits a speedup of 3.29$\times$ compared to the DGL-Xbyak implementation, and the TPP-based MLP kernels (BRGEMM-TPP tensor contractions with \emph{fused} TPP activation functions) exhibit a speedup of 1.4$\times$ compared to the respective DGL-Xbyak implementation. The FP32 TPP-based implementation spends 66.4\%  on the aggregation kernel and 33.6\% on the fused MLP kernels.}

\newtext{The last 8 bars on Figure~\ref{fig:gnn_perf_breakdown} illustrate the performance breakdown of the TPP-based implementation on various platforms (CLX/BDX/ROME/ICX/GRAVITON2/CPX) and various precisions (FP32 and CPX-BF16). We want to emphasize that all these performance numbers are obtained by employing a the same exact TPP-based code (which is \emph{platform-agnostic}); the only modification is pertaining to the BF16 TPP code where we changed the  tensor layouts in the MLP portion in order to deal with the required VNNI format. When comparing the CPX-F32 and the CPX-BF16 performance breakdowns we observe a 2$\times$ speedup on the Aggregation kernel. This kernel is typically bandwidth bound due to its irregular/indexed accesses, and the BF16 TPP code moves half of the data compared to the FP32 TPP code since all the tensors are halved in size (BF16 vs FP32 datatype). The MLP portion of the TPP-based implementation is sped up by 1.73$\times$ by using the BF16 BRGEMM-TPP. The CPX platform supports the BF16 FMA instruction which has effectively 2$\times$ the compute throughput compared to the FP32 FMA on the same platform. The BF16 BRGEMM-TPP internally leverages this BF16 FMA instruction within the GEMM microkernel on CPX (see subsection~\ref{subsec:brgemm_structure}) to speed up the tensor contraction. Finally, we highlight here the speedup of the Aggregation kernel when e.g.\ comparing the CPX and the ICX FP32 TPP-based performance numbers. The ICX platform has STREAM bandwidth of 175 GB/s whereas CPX has 97.7 GB/s, and this trend is reflected also in the performance of the Aggregation kernel (1.54$\times$ faster on ICX than CPX).}

\subsection{Distributed-memory scaling of DL workloads}
\begin{figure}[t!]
\centering
\includegraphics[width=\columnwidth]{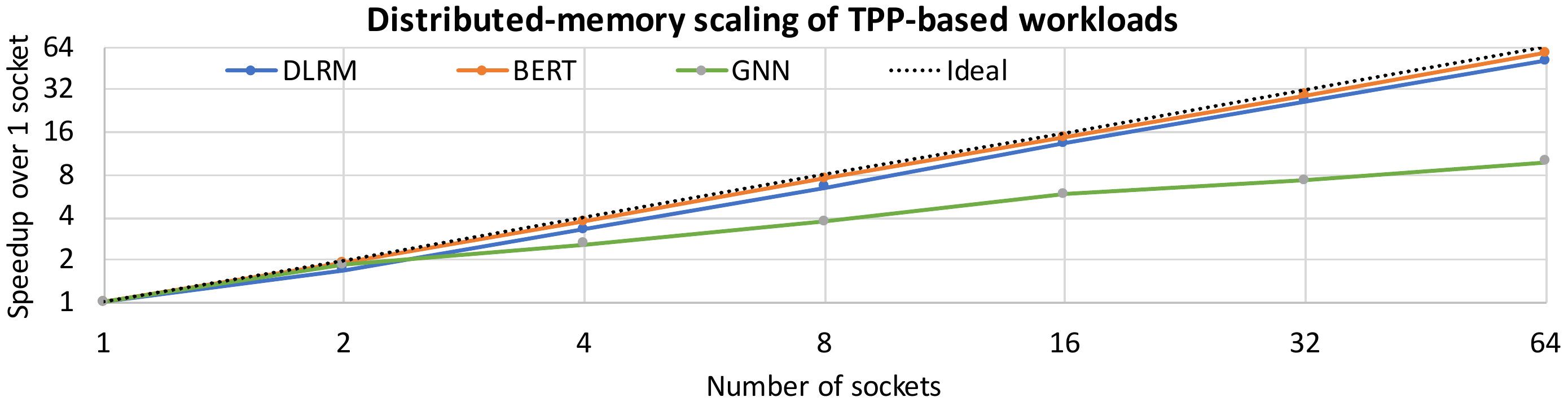}
\caption{Distributed-memory scaling of workloads}
\label{fig:dist_scaling}
\end{figure}
Even though we focused on the evaluation of the TPP-based workloads on a single node, our approach is seamlessly incorporated into the DL frameworks, hence we can scale to multiple nodes in a cluster to accelerate the training process employing the oneCCL library~\citep{oneccl}. Figure~\ref{fig:dist_scaling} shows the distributed-memory scaling of the TPP-based workloads. DLRM and BERT show almost perfect weak-scaling from 1 to 64 sockets of CLX (32 nodes) with speedups 51.7$\times$ and 57.9$\times$ respectively. Regarding the scaling of the GNN workload, the efficiency is directly affected by the quality of the partitions produced by the graph partitioning tools. Using 64 sockets we achieve 10$\times$ speedup compared to single socket, and further scaling improvements constitute future work. We can conclude that TPPs for single node optimizations combined with small-size cluster level execution can accelerate deep learning training on CPUs by up to two orders of magnitude.

\section{TPP within MLIR and a Tensor Compiler}
\begin{figure}[!htb]
\centering
\includegraphics[width=0.65\columnwidth]{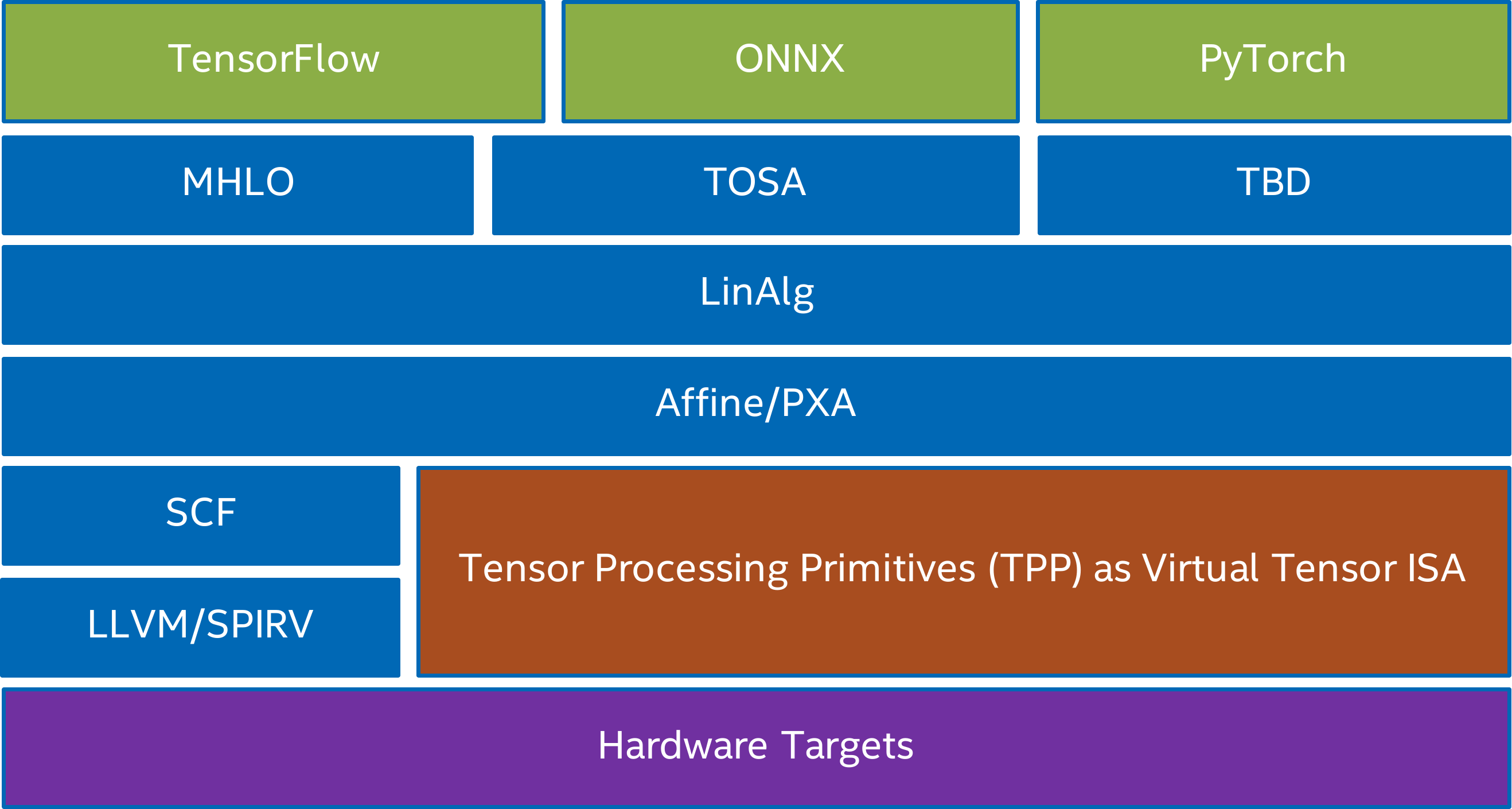}
\caption{\newtext{Example lowering paths within the PlaidML Tensor compiler in order to achieve full network optimization from popular frameworks. The green boxes represent the DL frameworks, the blue boxes correspond to MLIR dialects, the brown box shows the TPP-MLIR dialect within the stack, and the purple box represents the targeted platforms.}}
\label{fig:plaidml-tpp-lowerings}
\end{figure}
In order to illustrate the viability of TPPs as a virtual Tensor ISA within MLIR and Tensor Compilers, we implemented a rudimentary MLIR dialect corresponding to the TPPs. We also implemented lowering passes within the PlaidML~\citep{plaidml} Tensor Compiler that transform intermediate MLIR representations to the TPP-MLIR dialect. The TPP-MLIR dialect is subsequently lowered to the corresponding LIBXSMM TPP calls, therefore such a flow is not relying on LLVM for the code generation of the corresponding tensor operations.

\newtext{The current lowering path through MLIR supports a variety of front-end interfaces with LinAlg or Tile as the lowest level common entry points, i.e.\ the lowest level of abstraction that inbound programs can be specified in such that they will be subject to the full range of optimizations necessary to achieve full performance. Figure~\ref{fig:plaidml-tpp-lowerings} details the lowering paths currently implemented in PlaidML and where key transforms map tensor operations into the TPP dialect. The key transformation is located in the stencil pass of the PXA dialect (Parallel eXtensions for Affine - a staging ground for PlaidML/TPP work that will be proposed upstream to the affine dialect). Operations that cannot be matched to TPP primitives are lowered through standard affine optimization pipelines.}

\label{subsec:plaidml}
\begin{figure}[t!]
\centering
\includegraphics[width=\columnwidth]{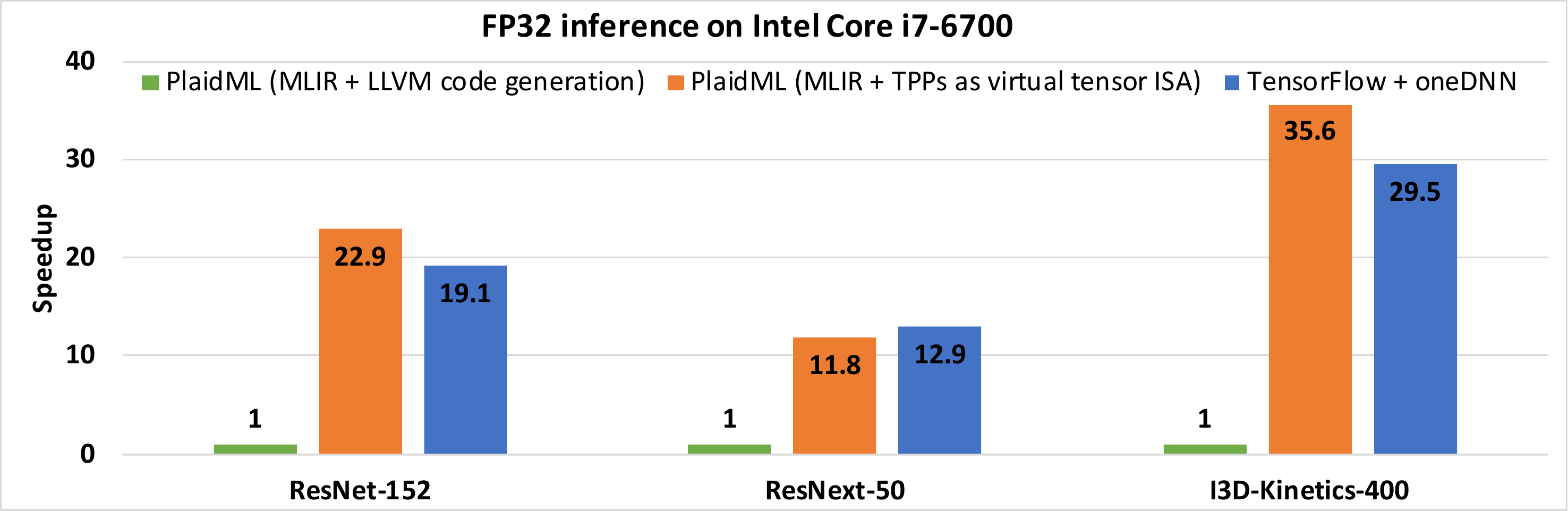}
\caption{\color{black}FP32 inference with PlaidML on various workloads: ResNet-152, ResNext-50, and I3D-Kinetics-400.}
\label{fig:plaidml}
\end{figure}
We experimented with the use-case of FP32 inference on a client CPU (Intel i7-6700) on three different workloads: ResNet-152~\citep{he2016deep}, ResNext-50~\citep{xie2017aggregated}, and I3D-Kinetics-400~\citep{carreira2017quo}. Figure~\ref{fig:plaidml} shows the results of three implementations: i) The green bars show the performance of the code generated by PlaidML with MLIR for intermediate representations, and LLVM for the code generation, ii) The orange bars show the performance of the code generated by PlaidML with MLIR for intermediate representations, and the TPP-MLIR dialect as virtual Tensor ISA for the code generation of the corresponding tensor contractions, and iii) TensorFlow FP32 inference backed-up by the vendor-optimized oneDNN library. We observe that the Tensor Compiler variant which relies on the TPP-MLIR dialect for the tensor contractions outperforms the variant which relies exclusively on LLVM (for loop-tiling and vectorization) up to ~35.6$\times$. At the same time, PlaidML assisted by the TPP-MLIR dialect matches/outperforms the performance of TensorFlow which uses internally oneDNN, a highly-tuned vendor library for this CPU target. These preliminary results highlight the viability of the synergistic Tensor Compiler - TPP paradigm as discussed in Section~\ref{sec:introduction}.

\section{TPP and HPC Applications}
\label{sec:hpc}
\newtext{So far, in this paper the focus was on how the TPP abstraction can be leveraged within the Deep Learning Domain. Tensor computations are ubiquitous, and in particular they constitute the cornerstone of many HPC applications. As such, the TPP abstraction can be readily employed by HPC applications to accelerate tensor computations without sacrificing portability. In the rest of this section we examine how TPPs are used within two HPC applications, namely CP2K and EDGE.}

\subsection{CP2K}
\label{sec:hpc_cp2k}
\newtext{The tensor based formulation originated and became common in physics, and it is well adopted in the field of engineering or applied sciences, and in electronic structure (ES) theory in particular. CP2K is an open source ES- and MD-package (molecular dynamics) for atomistic simulations of solid-state, liquid, molecular, and biological systems \citep{doi:10.1063/5.0007045}. CP2K is striving for good performance on HPC and massively parallel systems. Even though the use of novel algorithms in CP2K is the norm for scientific reasons, implementations have not widely tapped tensors in an explicit fashion. In contrast, Machine Learning emerged with similar, yet not coherent APIs and frameworks around the notions of tensors, layers and image processing.}

\newtext{While ES calculations can be formulated with tensors of ranks two to four, CP2K (and similar packages) largely remain with matrix based formulation. Various libraries for tensor contractions gained some attraction for scientific applications but the level of generality is key, e.g., as sparse representations are desired. CP2K explored an API for sparse tensor contractions and published a proof of concept implementation built into the DBCSR library \citep{DBLP:journals/corr/abs-1910-13555}. Efforts targeting accelerators in CP2K, namely GPUs, are not fully booked hence hardware specifically for Deep Learning (with focus on low and mixed precision arithmetic) is not yet a motivation of tensors as an implementation vehicle (and source of acceleration). Therefore a collection of primitives such as TPP is well-suited for an emerging discussion of a more general API.}

CP2K~3.0 introduced LIBXSMM for Small Matrix Multiplications (SMMs). CP2K and DBCSR (previously part of CP2K's code base) since then additionally introduced element-wise operations (copy and transpose) with "elements" being small matrices based on LIBXSMM. Reformulating existing code to build on (batched) GEMM TPP and element-wise TPP operations is an established pattern for increased performance in CP2K. 

To practically improve performance in CP2K one has to consider:
\begin{itemize}
\item Fusing kernels and increasing arithmetic intensity independent of the target being a CPU or an accelerator (performance bound by memory bandwidth).
\item Specializing code at runtime based on workload/input of the application, e.g., generating code Just-In-Time (JIT) a.k.a. meta-programming.
\end{itemize}
These objectives can be delivered by TPPs as a domain-specific language (DSL), enabling the scientist to write more abstract code, e.g., by the means of meta-programming, and by relying on a specification which delivers versatile primitives deferring low-level optimizations to the TPP backend.

For CP2K's performance evaluation, we refer to BDX, CLX, ICX, and ROME as introduced earlier (section~\ref{sec:results}). To show the portability of our approach, we augmented our results by using the Oracle Cloud Infrastructure, namely the result for \textit{Altra} processor (BM.Standard.A1.160 OCI shape). Table~\ref{tab:cp2k} shows the performance benefit of LIBXSMM's GEMM-TPP in CP2K when compared to Intel's MKL GEMM routines.

\begin{table}[htbp]
  \caption{CP2K performance (Cases/Day) of three workloads fitting into single systems with two processors. Single-socket performance is reported here for consistency within this paper. Intel MKL or OpenBLAS are always used for general BLAS operations including large GEMMs. Either \mbox{(BLAS-)}GEMM or TPP-GEMM was used for batched multiplication of small matrices (SMMs). Workloads utilizing CP2K's DBCSR library for distributed block-sparse matrix multiply benefit from \mbox{(runtime-)}specialized GEMM-TPP kernels where the set of matrix shapes is not known at compile-time of the application or depends on the workload in general.}
  \begin{center}
    \begin{tabular}{l|l|c|c|c}
      \textbf{System\ \ \ \ \ } & \textbf{Workload$^\mathrm{a}$} & \textbf{BLAS-GEMM$^\mathrm{b}$} & \textbf{TPP-GEMM$^\mathrm{c}$} & \textbf{TPP-Speedup} \\
      \hline
      \ BDX   & H2O-256     & $91$  & $101$ & $11\%$ \\
              & H2O-512     & $23$  & $27$  & $17\%$ \\
      \hline
      \ CLX   & H2O-256     & $154$ & $162$ & $5\%$ \\
              & H2O-512     & $39$  & $41$  & $5\%$ \\
              & H2O-DFT-LS4 & $45$  & $47$  & $4\%$ \\
      \hline
      \ ICX   & H2O-256     & $235$ & $249$ & $6\%$ \\
              & H2O-512     & $60$  & $65$  & $8\%$ \\
              & H2O-DFT-LS4 & $67$  & $70$  & $4\%$ \\
      \hline
      \ ROME  & H2O-256     & $225$ & $244$ & $8\%$ \\
              & H2O-512     & $55$  & $57$  & $4\%$ \\
              & H2O-DFT-LS4 & $65$  & $65$  & $0\%$ \\
      \hline
      \ Altra & H2O-256     & $228$ & $236$ & $4\%$ \\
              & H2O-512     & $60$  & $62$  & $3\%$ \\
              & H2O-DFT-LS4 & $60$  & $66$  & $10\%$ \\
      \multicolumn{5}{l}{\footnotesize\rule{0pt}{3ex}$^{\mathrm{a}}$H2O-256 (CP2K bench.), H2O-512 (UEABS Case\,A) and H2O-DFT-LS\,NREP=4 (UEABS Case\,C) from \href{https://prace-ri.eu/training-support/technical-documentation/benchmark-suites/}{PRACE UEABS}~2.1.} \\
      \multicolumn{5}{l}{\footnotesize$^{\mathrm{b}}$Intel MKL (x86-64) or OpenBLAS (otherwise).} \\
      \multicolumn{5}{l}{\footnotesize$^{\mathrm{c}}$LIBXSMM.} \\
    \end{tabular}
    \label{tab:cp2k}
  \end{center}
\end{table}

\subsection{EDGE}
\label{sec:hpc_edge}
\newtext{
The Extreme-Scale Discontinuous Galerkin Environment (EDGE) uses the Arbitrary high-order DERivatives (ADER) Discontinuous Galerkin (DG) finite element method to simulate seismic wave propagation \citep{10.1007/978-3-319-58667-0_3}.
The software uses unstructured tetrahedral meshes which are typically adapted to the used seismic velocity models.
Additionally, modelers may introduce mountain topography.
A sophisticated local time stepping scheme allows the solver to operate efficiently in very large and complex settings.
The software is able to fuse multiple ensemble simulations into one execution of the software.
EDGE uses an orthogonal polynomial expansion basis to discretize each of the considered variables in a tetrahedron of the mesh.
In a typical setting, we use three relaxation mechanisms for the viscoelastic part, resulting in a total of 27 seismic variables.
Additionally using a fifth order method gives us 35 basis functions, resulting in a total of $27 \cdot 35=945$ degrees of freedom per tetrahedral element.
The solver advances the degrees of freedom in time by repeatedly computing a triplet of quadrature-free integrators.
While the actual integrators are part of EDGE, their implementation relies heavily on TPPs.
The GEMM-TPP with small and uncommon matrix sizes is the  most crucial operation required by EDGE.
For example, the surface integrator requires the multiplication of a $9 \times 35$ matrix with a $35\times15$ matrix.
The solver's extension with additional, performance-portable TPPs in all parts of the integrators is work-in-progress.
Especially, EDGE's support for viscoelastic attenuation or local time stepping requires ``simpler`` kernels, e.g., the unary TPPs Identity and Zero, or the binary TPPs Mul, Sub and Add.}

\newtext{
We evaluate EDGE's performance-portability through the use of TPPs by studying the performance of a full setup of the Layer Over Halfspace 3 (LOH3) benchmark with 743,066 tetrahedral elements.
The same setting  was also used in \citep{breuer2021nextgen} to study the performance of the solver on a single processor of the Frontera supercomputer located at the Texas Advanced Computing Center (position ten in the 06/21 TOP500-list).
Following this study, a sophisticated simulation of the 2014 $\text{M}_{w}$ 5.1 La Habra earthquake using a mesh with 237,861,634 tetrahedral elements and EDGE's advanced features yielded a performance of 2.20 FP32-PFLOPS on 1,536 nodes.}

\begin{table}[htbp]
  \caption{Sustained 32-bit floating point performance on the studied systems. The performance is given in TFLOPS. Results are presented for Global Time Stepping (GTS) and Local Time Stepping (LTS) when using single and fused forward simulations.}
  \begin{center}
    \begin{tabular}{c|c|c|c|c}
      \textbf{System}
      &
      \multicolumn{2}{c|}{\textbf{GTS}}
      &
      \multicolumn{2}{c}{\textbf{LTS}} \\
      & \textbf{\textit{single}} & \textbf{\textit{fused}}$^\mathrm{a}$ & \textbf{\textit{single}} & \textbf{\textit{fused}}$^\mathrm{a}$ \\
      \hline
      Cascade Lake & 1.08 & 0.78 & 1.02 & 0.74 \\
      Ice Lake     & 1.29 & 1.01 & 1.23 & 0.96 \\
      Rome         & 1.20 & 1.08 & 1.12 & 1.01 \\
      Milan        & \textbf{1.39} & \textbf{1.16} & 1.29 & \textbf{1.07} \\
      Altra        & 1.27 & 0.73 & \textbf{1.51} & 0.76 \\
      \multicolumn{5}{l}{\footnotesize\rule{0pt}{3ex}$^{\mathrm{a}}$EDGE's fused simulations use sparse matrix kernels.}\\
    \end{tabular}
    \label{tab:flops_loh3}
  \end{center}
\end{table}

\newtext{For the EDGE application, we study the software's raw floating point performance and time-to-solution by extending our LOH3-Frontera-only study \citep{breuer2021nextgen} with diverse processors:
\begin{itemize}
    \item \textit{Cascade Lake} (similar to \textit{CLX} as introduced in section~\ref{sec:results}): 2.7 GHz 28-core Intel Xeon Platinum 8280 processor of the Frontera system at the Texas Advanced Computing Center. We only used a single 28-core processor of Frontera's dual-socketed compute nodes in our tests.
    \item \textit{Ice Lake:} 2.3 GHz 40-core Intel Xeon Platinum 8380 processor on Intel's on-premises cluster. We only used a single 40-core processor of the dual-socket compute nodes in our tests.
    \item \textit{Rome} (similar to \textit{ROME} as introduced in section~\ref{sec:results}): 2.25 GHz AMD EPYC 7742 (BM.Standard.E3.128 OCI shape). We only used a single 64-core processor of the bare metal instance in our tests.
    \item \textit{Milan:} 2.55 GHz AMD EPYC 7J13 (BM.Standard.E4.128 OCI shape). We only used single 64-core processor of the bare metal instance in our tests.
    \item \textit{Altra:} 3.0 GHz Ampere Altra Q80-30 processor (BM.Standard.A1.160 OCI shape). We only used a single 80-Armv8.2-core processor of the bare metal instance in our tests.
\end{itemize}}

\newtext{Table~\ref{tab:flops_loh3} shows the sustained floating point performance of the conducted runs.
All numbers are given in FP32-TFLOPS.
Columns two and three present the performance of Global Time Stepping (GTS), whereas columns four and five show that of Local Time Stepping (LTS).
In general, the LTS configurations have a slightly lower peak utilization when compared to their GTS counterparts.
Note, however, that Table~\ref{tab:flops_loh3} only shows raw floating point performance and does not account for time-to-solution speedups through LTS (theoretically up to 2.67$\times$ in this case).
The performance of GTS and LTS is further split into running a single forward simulation and fusing multiple simulations.
In fused mode, the solver parallelizes over the right-hand-side by concurrently simulating seismic wave propagation for a collection of seismic sources.
One of the fused mode's unique advantages is the opportunity for perfect vectorization of all small matrix multiplications, even when considering sparsity \citep{10.1007/978-3-319-58667-0_3}.
In this work we matched the microarchitectures' SIMD-length by fusing 16 simulations on Cascade Lake and Ice Lake, eight simulations on Rome and Milan, and four simulations on Altra.
Once again, note that Table~\ref{tab:flops_loh3} does not include the respective sparsity-driven $2.49\times$ increase of the floating point operations' value when running fused simulations.
Comparing the performance of the different systems, we observe very high overall performance with architectural efficiency gains originating from decreasing SIMD-lengths.
This is especially noticeable when running single forward simulations.
In this case, the vectorized dimension of the small dense matrix kernels coincides with the number of basis functions, i.e., $M=35$, which is challenging when optimizing for AVX512 (Cascade Lake and Ice Lake) and AVX2 (Rome and Milan).
The short 128-bit ASIMD vector instruction (Altra) reach a very high peak utilization of 33.2\% for GTS and 39.2\% in LTS.
For the fused simulations, the differences in relative peak utilization narrow further.}

\begin{table}[htbp]
  \caption{Time-to-solution speedups of the studied systems when using different configurations of the solver EDGE. The performance of the Cascade Lake system, running EDGE with Global Time Stepping (GTS) and a single forward simulation, is used as baseline. In contrast to Table~\ref{tab:flops_loh3}, the speedups include the higher algorithmic efficiencies of EDGE's support for Local Time Stepping (LTS) and fused forward simulations.}
  \begin{center}
    \begin{tabular}{c|c|c|c|c}
      \textbf{System}
      &
      \multicolumn{2}{c|}{\textbf{GTS}}
      &
      \multicolumn{2}{c}{\textbf{LTS}} \\
      & \textbf{\textit{single}} & \textbf{\textit{fused}} & \textbf{\textit{single}} & \textbf{\textit{fused}}\\
      \hline
      Cascade Lake & 1.00 & 1.80 & 2.50 & 4.52 \\
      Ice Lake     & 1.19 & 2.33 & 3.02 & 5.87 \\
      Rome         & 1.11 & 2.48 &  2.76  & 6.17 \\
      Milan        & \textbf{1.28} & \textbf{2.67} & 3.18 & \textbf{6.55} \\
      Altra        & 1.18 & 1.69 & \textbf{3.71} & 4.64 \\
    \end{tabular}
    \label{tab:speedups_loh3}
  \end{center}
\end{table}

\newtext{Table~\ref{tab:speedups_loh3} describes the obtained performance numbers in terms of time-to-solution.
Here, we use the runtime of the studied LOH3 setting on Cascade Lake for GTS and a single forward simulation as baseline.
All other settings are given relative to this.
Further, for the fused settings, we consider the per-simulation time.
We observe that EDGE's overall performance is driven by the high floating point performance through the use of TPPs and the solver's advanced algorithmic features.
Here, Altra performs best for single forward simulations using LTS, accelerating the baseline by 3.71$\times$.
Milan has the best time-to-solution in all other settings and is able to outperform the baseline by 6.55$\times$ when using LTS and fusing simulations.
This performance lead originates from Milan's high theoretical peak combined with a high peak utilization (see Table~\ref{tab:flops_loh3}).
}

\section{Related Work}
\label{sec:related}

The related work in terms of the development methodology of DL workloads has been referenced in the introduction, so here we mention community efforts that share the same design philosophy with TPPs. Tensor Operator Set Architecture (TOSA) is a recent work, concurrently developed with TPPs, that provides a set of whole-tensor operations commonly employed in DL~\citep{tosa}. TOSA allows users to express directly operators on up to 4D/5D tensors which are not naturally mapped even on contemporary 2D systolic hardware. We believe that staying at the 2D primitive level is expressive and sufficient, as we can build higher-order ops with loops around 2D operators, e.g. see Algorithm~\ref{alg:fc}. Despite the similarities of TPP and TOSA specifications, the TOSA back-end is reference C code and is not showcased in full DL-workloads. CUTLASS~\citep{cutlass} and Triton~\citep{tillet2019triton} strive for high-performance on GPUs, while also offer flexible composition that can be easily applied to solve new problems related in DL and linear algebra, and share many design principles with TPPs. XLA~\citep{xla} is a domain-specific compiler for linear algebra and DL that targets TensorFlow models with potentially no source code changes. JAX~\citep{jax} provides automatic differentiation of Python and NumPy functions, and the compilation of the desired operators happens in a user-transparent way with JIT calls, yielding optimized XLA kernels. XLA and JAX share the same philosophy with TPPs: the user is focusing on the DL kernel/workload development using high-level, platform-agnostic, declarative-style programming, whereas the tensor-aware back-end infrastructure undertakes the efficient and portable code generation.

Tensor Compilers (TC)~\citep{plaidml,chen2018tvm,vasilache2018tensor,zheng2020ansor} attempt to optimize DL operators in a platform-agnostics way, however their applicability is restricted to relatively small code-blocks whereas full workload integration is cumbersome. Also, TC undertake the tasks of efficient parallelization, loop re-ordering, automatic tiling and layout transformations, nevertheless the obtained performance is typically underwhelming~\citep{barham2019machine}. We envision that TPPs can be used as a tool by TC in order to attain efficient platform-specific code generation, therefore TC could focus on optimizing the higher level aspects of the tensor programs (e.g.\ layout transformations). Along these lines, TPPs fit in the MLIR~\citep{mlir} ecosystem/stack as a lowering dialect (see Section~\ref{subsec:plaidml}), and in this way the TPP back-end could be leveraged by multiple TC frameworks.

\section{Conclusions And Future Work}
\label{sec:conclusions}
 In this work we presented the Tensor Processing Primitives (TPP), a compact, yet versatile set of 2D-tensor operators, which subsequently can be utilized as building-blocks to construct efficient, portable complex DL operators on high-dimensional tensors. \newtext{We also show how TPPs can be used within HPC applications in order to accelerate tensor computations.} We demonstrate the efficacy of our approach using standalone kernels and end-to-end training DL-workloads (CNNs, dilated convolutions, DLRM, BERT, GNNs) expressed entirely via TPPs that outperform state-of-the-art implementations on multiple platforms. As future work, we plan to create a full-fledged TPP-based MLIR dialect such that Tensor Compilers can leverage the strengths of TPPs . Also, we plan to further enrich the TPP back-end implementation by supporting more ISAs, including GPUs and POWER architectures. 

\section*{GLOSSARY}
\subsection*{Intel Pseudo Intrinsics}
\begin{enumerate}
    \item \emph{\textbf{\_mm128 }}  Represents a vector of width 128 bits.
    \item \emph{\textbf{\_mm128\_loadu\_ps(addr) }} Loads 16byte of 32\,bit elements.
    \item \emph{\textbf{\_mm128\_storeu\_ps(addr) }} Stores 16byte of 32\,bit elements.
    \item \emph{\textbf{\_mm128\_unpacklo\_ps(A, B) }} Unpacks and interleaves 32\,bit elements from the low half of A and B.
    \item \emph{\textbf{\_mm128\_unpackhi\_ps(A, B) }} Unpacks and interleaves 32\,bit elements from the high half of A and B.
    \item \emph{\textbf{\_mm128\_unpacklo\_pd(A. B) }} Unpacks and interleaves 64\,bit elements from the low half of A and B.
    \item \emph{\textbf{\_mm128\_unpackhi\_pd(A, B) }} Unpacks and interleaves 64\,bit elements from the high half of A and B.
    \item \emph{\textbf{\_mm512 }}  Represents a vector of width 512 bits.
    \item \emph{\textbf{\_mm512\_permutexvar\_ps(A,B) }} Shuffle single precision floating point elements in 512 wide vector length using indexes specified in B.
    \item \emph{\textbf{\_mm512\_roundscale\_ps(A,B) }} Round single precision floating point elements to the rounding mode specified by argument B.
    \item \emph{\textbf{\_mm512\_sub\_ps(A,B) }} Subtract single precision floating point elements in A from  B. 
    \item \emph{\textbf{\_mm512\_scalef\_ps(A,B) }} Scales single precision floating point elements in A using values specified in B.
    \item \emph{\textbf{\_mm512\_range\_ps(A,B, int imm8) }} Calculates the min, max or absolute max for each single precision- floating point elements in  A and B. Lower 2 bits of imm8[1:0]  specifies the operation(min/max/absolute max) to be performed.
    \item \emph{\textbf{\_mm512\_xor\_ps(A,B) }} Performs XOR operation between each single precision floating point elements in A and B vector.
    \item \emph{\textbf{\_mm512\_and\_ps(A,B) }} Performs AND operation between each single precision floating point elements in A and B vector.
    \item \emph{\textbf{\_mm512\_rcp14\_ps(A,B) }}  Calculates approximate reciprocal of each single precision floating point element in  range less then 2\^-14.
    \item \emph{\textbf{\_mm512\_cmp\_ps\_mask(A,B,int C) }} Compare the single precision elements in A and B specified by the comparison mode in C.
    \item \emph{\textbf{\_mm512\_mask\_blend\_ps(mask A,B,C) }} Copies single precision floating point element from vector A in vector C if the corresponding  mask bit is set .
    \item \emph{\textbf{\_mm512\_fmadd\_ps(mask A,B,C) }} Fused-Multiply-Add: Multiplies elements from vector A and B and adds them to elements of vector C.
    \item \emph{\textbf{\_mm512\_maskz\_loadu\_epi16(mask, addr) }} Loads 64byte of 16bit elements under zero masking from address addr.
    \item \emph{\textbf{\_mm512\_set1\_epi32( value ) }} sets a 32\,bit value into all 16 entries of the vector, e.g. broadcast.
    \item \emph{\textbf{\_mm512\_maskz\_mov\_epi16(mask, A) }} Moves 16\,bit-type register A under zero-masking to a different register.
    \item \emph{\textbf{\_mm512\_slli\_epi32(A, imm) }} Shifts all entries in the vector registers (typed as 32\,bit elements) by value imm to the left by shifting 0 in.
\end{enumerate}
\subsection*{Arm Pseudo Intrinsics}
\begin{enumerate}
    \item \emph{\textbf{vld1q\_f32(addr)}} Loads 16byte of 32\,bit elements.
    \item \emph{\textbf{vst1q\_f32(addr)}} Loads 16byte of 32\,bit elements.
    \item \emph{\textbf{vtrn1q\_f32(A, B) }} Unpacks and interleaves 32\,bit elements from the low half of A and B.
    \item \emph{\textbf{vtrn2q\_f32(A, B) }} Unpacks and interleaves 32\,bit elements from the high half of A and B.
    \item \emph{\textbf{vtrn1q\_f64(A. B) }} Unpacks and interleaves 64\,bit elements from the low half of A and B.
    \item \emph{\textbf{vtrn2q\_f64(A, B) }} Unpacks and interleaves 64\,bit elements from the high half of A and B.
    \item \emph{\textbf{vmax\_q(A,B) }} Calculates the maximum between each single precision floating point elements in A and B vector.
    \item \emph{\textbf{vmin\_q(A,B) }} Calculates the minimum between each single precision floating point elements in A and B vector.
    \item \emph{\textbf{vmul\_q(A,B) }} Multiply single precision elements in A and B vector.
    \item \emph{\textbf{vsub\_q(A,B) }} Subtract corresponding single precision elements in B from A.
    \item \emph{\textbf{vadd\_q(A,B) }} Add single precision elements in B and A.
    \item \emph{\textbf{vshlq\_u32(A,B) }} Shift left each single precision elements in A by the value specified in B.
    \item \emph{\textbf{vrndmq\_f32(A) }} Round single precision floating point elements in A using minus infinity rounding mode.
    \item \emph{\textbf{vcvtmq\_s32\_f32(A) }} Converts single precision floating point elements in A to signed integers using  minus infinity rounding mode.
    \item \emph{\textbf{float32x4\_t }} Represents 4 single precision floating point elements in vector width of 128.
    \item \emph{\textbf{vand\_q(A,B) }} Performs bit-wise AND operation between A and B vector.
    \item \emph{\textbf{vfmaq\_f32(A,B,C) }} Multiply single precision elements in A and B.Add the intermediate result to C.  
    \item \emph{\textbf{vld1q\_f32(A) }} Load a single precision element from scalar to all single precision element in  a vector.
    \item \emph{\textbf{vtbl1\_u8(A,B) }} Performs a  byte look up operation in vector A using byte addressable indexes specified in vector B.
    \item \emph{\textbf{vtbl4\_u8(A,B) }} Performs a  64 byte look up operation in vector A, A+1, A+2, A+3 using byte addressable indexes specified in vector B.
    \item \emph{\textbf{vbcaxq\_s32(A,B) }} Performs XOR operation between each single precision floating point elements in A and B vector.
    \item \emph{\textbf{vcgt\_q(A,B) }} Compare corresponding single precision elements in A and B. If B is greater then A the corresponding bits are set in the destination vector.
    \item \emph{\textbf{vrecpe\_f32(A) }}  Calculates approximate reciprocal of each single precision floating point element in vector A.
    \item \emph{\textbf{vbit\_insert(A,B) }} Copies single precision floating point element from vector A in destination vector if the corresponding bits are set in vector B.
    
\end{enumerate}

\FloatBarrier

\bibliographystyle{unsrt}
\clearpage
\bibliography{references}

\scriptsize
\noindent
\newline Optimization Notice: Software and workloads used in
performance tests may have been optimized for performance only on
Intel microprocessors.  Performance tests, such as SYSmark and
MobileMark, are measured using specific computer systems,
components, software, operations and functions.  Any change to any
of those factors may cause the results to vary.  You should
consult other information and performance tests to assist you in
fully evaluating your contemplated purchases, including the
performance of that product when combined with other products.
For more information go to http://www.intel.com/performance.

\noindent Intel, Xeon, and Intel Xeon Phi are trademarks of Intel Corporation in the U.S. and/or other

\normalsize


\end{document}